\documentclass[10pt,twocolumn,letterpaper]{article}

\usepackage{iccv}
\usepackage{times}
\usepackage{epsfig}
\usepackage{graphicx}
\usepackage{amsmath}
\usepackage{amssymb}

\usepackage[breaklinks=True,bookmarks=False]{hyperref}
\hypersetup{
     colorlinks = true,
     linkcolor = red!80!white,
     citecolor = green!70!black
}
\iccvfinalcopy 

\usepackage{Inputs/packages}

\begin{document}
\maxdeadcycles=200
\extrafloats{100}

\title{SMIT: \TITLE}

\author{Andr\'es Romero\\
BCV Lab\\
Universidad de Los Andes\\
{\tt\footnotesize rv.andres10@uniandes.edu.co}
\and
Pablo Arbel\'aez\\
BCV Lab\\
Universidad de Los Andes\\
{\tt\footnotesize pa.arbelaez@uniandes.edu.co}
\and
Luc Van Gool\\
ETH Z\"urich\\
KU Leuven\\
{\tt\footnotesize vangool@ethz.ch}
\and
Radu Timofte\\
CV Lab\\
ETH Z\"urich\\
{\tt\footnotesize timofter@ethz.ch}
}

\maketitle
\begin{abstract}
Cross-domain mapping has been a very active topic in recent years. Given one image, its main purpose is to translate it to the desired target domain, or multiple domains in the case of multiple labels. This problem is highly challenging due to three main reasons: \textit{(i)} unpaired datasets, \textit{(ii)} multiple attributes, and \textit{(iii)} the multimodality (\eg style) associated with the translation. Most of the existing state-of-the-art has focused only on two reasons \ie, either on \textit{(i)} and \textit{(ii)}, or \textit{(i)} and \textit{(iii)}. In this work, we propose a joint framework (\textit{i, ii, iii}) of diversity and multi-mapping image-to-image translations, using a single generator to conditionally produce countless and unique fake images that hold the underlying characteristics of the source image. Our system does not use style regularization, instead, it uses an embedding representation that we call domain embedding for both domain and style. Extensive experiments over different datasets demonstrate the effectiveness of our proposed approach in comparison with the state-of-the-art in both multi-label and multimodal problems. Additionally, our method is able to generalize under different scenarios: continuous style interpolation, continuous label interpolation, and fine-grained mapping. Code and pretrained models are available at \url{https://github.com/BCV-Uniandes/SMIT}.
\end{abstract}

\vspace{-0.4cm}
\section{Introduction}
The ability of humans to easily imagine how a black haired person would look like if they were blond, or with a different type of eyeglasses, or to imagine a winter scene as summer is formulated as the image-to-image (I2I) translation problem in the computer vision community. Since the recent introduction of Generative Adversarial Networks (GANs)~\cite{goodfellow2014generative}, a plethora of problems such as video analysis~\cite{tulyakov2017mocogan,bansal2018recycle}, super resolution~\cite{ledig2017superresolution,blau2018PIRM}, semantic synthesis~\cite{isola2017pix2pix,chen2017photographic}, photo enhancement~\cite{ignatov2017dslr,ignatov2017wespe}, photo editing~\cite{shu2017neural,dolhansky2018eye}, and most recently domain adaptation~\cite{hoffman2017cycada,murez2018image} have been addressed as I2I translation problems. 

\begin{figure}[t]
\begin{center}
   \includegraphics[width=0.9\linewidth]{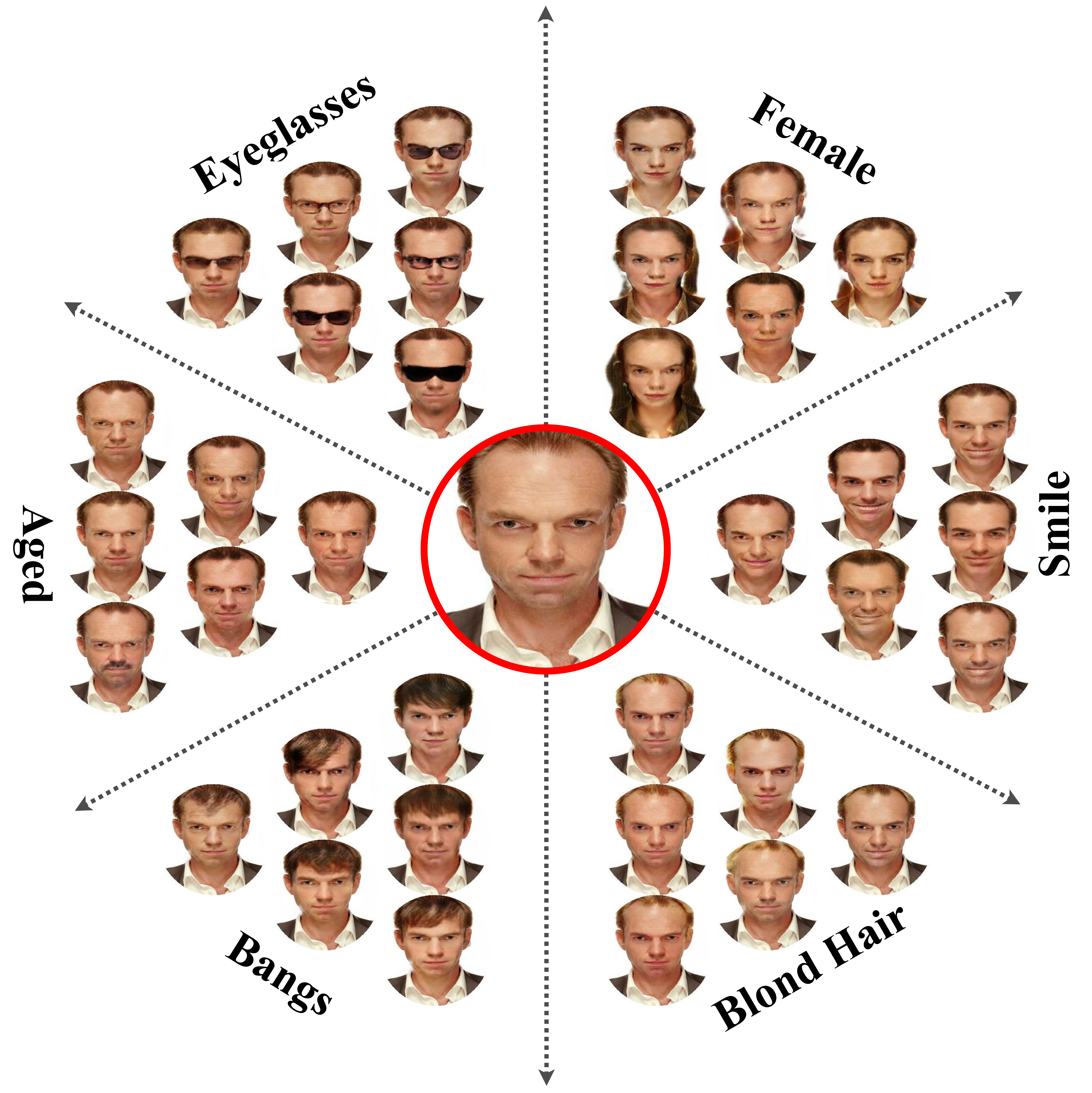}
\end{center}
\vspace{-0.3cm}
   \caption{\textbf{\TITLE~(SMIT)}. Our model learns a full diverse representation for multiple attributes using a single generator.}
\label{fig:teaser}
\vspace{-0.4cm}
\end{figure}

\begin{figure*}[t]
\begin{center}
   \includegraphics[width=0.9\linewidth]{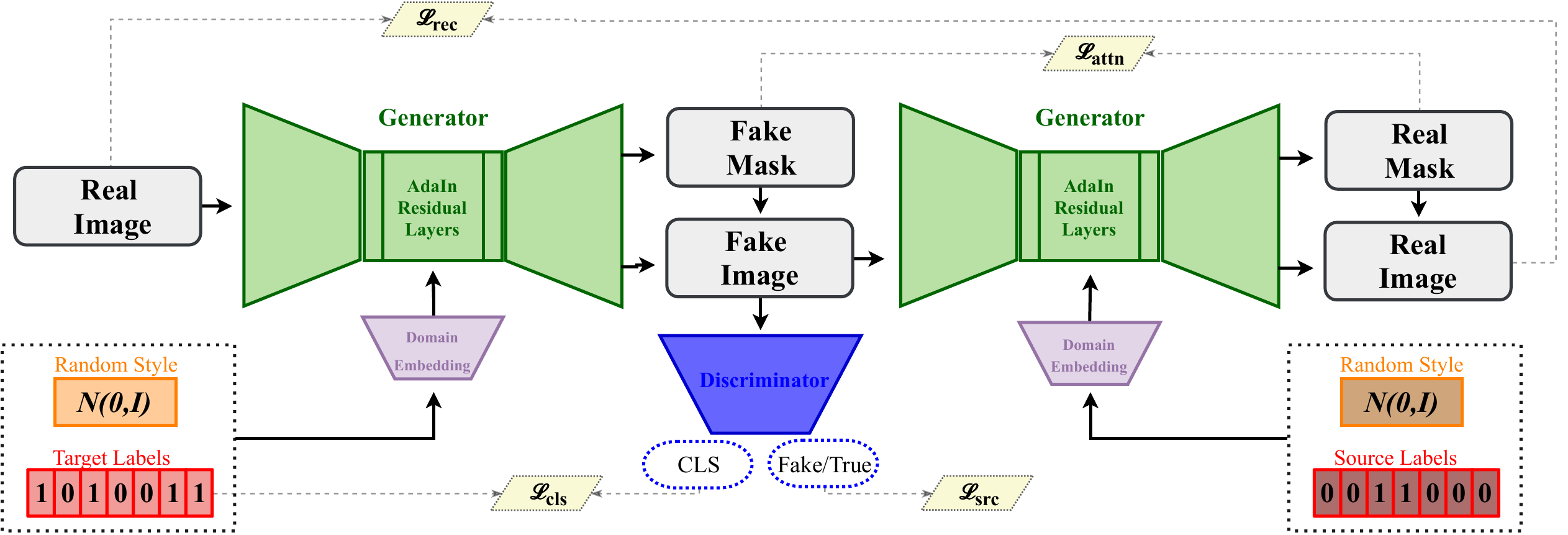}
\end{center}
\vspace{-0.3cm}
   \caption{\textbf{Overview of SMIT.} We translate an image by jointly taking as input a random style and target attributes into the generator. The Domain Embedding is a map projection that uses random and fixed parameters for the embedding. The discriminator aims at classifying only the source and the attributes, \ie no style regularization. We use the original source attributes and a different style to recover the real image.}
\label{fig:overview}
\vspace{-0.3cm}
\end{figure*}

Initially, translating from one domain into another required paired datasets that exactly matched both domains~\cite{isola2017pix2pix} \eg, edges$\leftrightarrow$shoes or edges$\leftrightarrow$handbags datasets. However, this approach is unpractical because the full representation of the cross-domain mapping is, in most cases, intractable. Existing techniques try to perform deterministic I2I translation with unpaired images to map from one domain into another (one-to-one)~\cite{zhu2017cyclegan,anoosheh2018combogan,liu2017unit,ignatov2017wespe}, or into multiple domains (one-to-many)~\cite{choi2017stargan,pumarola2018ganimation,he2017attrgan_v1}. Nevertheless, many problems are fundamentally stochastic as there are countless mappings from one domain to another \eg, a day$\leftrightarrow$night or cat$\leftrightarrow$dog translation. 

Recent techniques~\cite{lee2018drit,huang2018munit,ma2018exemplarmunit2} have successfully addressed the multimodal representation for one-to-one domain translation. These methods are based on the idea developed on traditional I2I approaches~\cite{zhu2017cyclegan,zhu2017bicycle}, in which the generator tends to overlook a noise injection. As a consequence, these techniques studied the problem of disentangling representation as style transfer, including a shared content space representation and a style encoder network.

In this paper, we propose \TITLE~(SMIT), a novel and robust framework that includes multiple labels and diversity, and does not require either style or content regularization. Moreover, we build our entire approach using a single generator that does not ignore the noise perturbation, \ie for different level of noise our method produces different styles with the underlying characteristics and structure of the target domain\footnote{Hereafter, we refer to domains as the number of labels per dataset, and style as the diversity induced by noise.}. As illustrated in \fref{fig:teaser}, SMIT learns a full distribution for each attribute, so it can perform diverse translation for different fine-grained or broader attributes. It is important to remark that in contrast to ~\cite{choi2017stargan,pumarola2018ganimation,lample2017fader} the trainable parameters in the SMIT generator are not label-dependent, that is there is a negligible difference either on computational time or on memory consumption when learning as many as 40 attributes instead of just 2 labels. \fref{fig:overview} presents an overview of our model. We radically depart from mainstream approaches~\cite{choi2017stargan,pumarola2018ganimation,lample2017fader}, where the target domain is inserted through the spatial concatenation, instead we indirectly inject the style and the target labels through Adaptive Instance Normalization (AdaIN)~\cite{huang2017arbitraryadain} layers in the generator, and the discriminator aims at recovering only the labels, \ie we remark the importance of no style regularization.

We perform a comprehensive quantitative evaluation of SMIT either for disentanglement or multiple domain I2I problems, demonstrating the advantages of our method in comparison with existing state-of-the-art models. We also show qualitative results on several datasets that validate the effectiveness of our approach under varied and challenging settings. 

More precisely, our main contribution is to propose a single and end-to-end system with an agnostic-domain generator capable of performing style transformation, multi-label mapping, style interpolation, and continuous label interpolation with no need of style regularization. For reproducibility, we plan to release our source code and trained models. 


\section{Related Work}

\begin{table*}[t]
\vspace{-0.3cm}
\begin{center}
\resizebox{\linewidth}{!}{
\begin{tabular}{rccccccc}
\hline
              & CycleGAN & BiCycleGAN & StarGAN & MUNIT\&alike & DRIT & GANimation &  \textbf{SMIT} \\
              & \cite{zhu2017cyclegan} & \cite{zhu2017bicycle} & \cite{choi2017stargan} &  \cite{huang2018munit,almahairi2018augmentedcyclegan,ma2018exemplarmunit2} & \cite{lee2018drit} &
              \cite{pumarola2018ganimation} &
              \textbf{(ours)} \\     
\hline
\hline
Unpaired Training              & \cmark &        & \cmark & \cmark & \cmark & \cmark & \color{green!80!black}{\cmark} \\
\rowcolor[HTML]{\colortable} 
Multimodal Generation          &        & \cmark &        & \cmark & \cmark &        & \color{green!80!black}{\cmark} \\
Multiple Attributes            &        &        & \cmark &        &        & \cmark & \color{green!80!black}{\cmark} \\
\rowcolor[HTML]{\colortable} 
One Single Generator           &        &        & \cmark &        &        & \cmark & \color{green!80!black}{\cmark} \\
Fine-grained Transformation    &        &        & \cmark &        &        & \cmark & \color{green!80!black}{\cmark} \\
\rowcolor[HTML]{\colortable} 
Continuous Label Interpolation &        &        &        &        &        & \cmark & \color{green!80!black}{\cmark} \\
Style Transformation           &        &        &        & \cmark & \cmark &        & \color{green!80!black}{\cmark} \\
\rowcolor[HTML]{\colortable} 
Style Interpolation            &        &        &        & \cmark & \cmark &        & \color{green!80!black}{\cmark} \\
Attention Mechanism            &        &        &        &        &        & \cmark & \color{green!80!black}{\cmark} \\
\hline
\end{tabular}
}
\end{center}
\caption{\textbf{Feature comparison with state-of-the-art approaches in I2I translation.} SMIT uses a single generator trained with unpaired data to produce disentangled representations of a multi-targeted domain.}
\label{table:related_work}
\vspace{-0.4cm}
\end{table*}

Generative Adversarial Networks (GANs)~\cite{goodfellow2014generative} have proven to be a powerful approach to learn statistical data distributions. GANs rely on game theory where there are two networks (discriminator and generator) optimizing a Minimax function, a training scheme also known as adversarial training. The discriminator learns to distinguish real images from fake ones produced by the generator, and the generator learns to fool the discriminator by producing realistic fake images. Since their introduction, GANs have provided remarkable results in several computer vision problems, such as image generation~\cite{radford2015unsupervised,chen2016infogan,karras2017progressive}, image translation~\cite{isola2017pix2pix,zhu2017cyclegan,almahairi2018augmentedcyclegan,liu2017unit}, video translation~\cite{tulyakov2017mocogan,bansal2018recycle} and resolution enhancement~\cite{bai2018finding,ledig2017superresolution,agustsson2018extreme}. As our approach lies in the domain of image-to-image translation, it is the focus of our related work review. 

\vspace{-0.2cm}
\paragraph{Conditional GANs (cGANs)}
In vanilla GANs~\cite{goodfellow2014generative}, the information regarding the domain is unknown. Conversely, on conditional GANs (cGANs)~\cite{odena2016conditional}, the discriminator not only distinguishes between real and fake, but it also trains an auxiliary classifier for the conditional data distribution. cGANs have been applied in image-to-image translation problems for semantic layouts~\cite{isola2017pix2pix,chen2017photographic}, super resolution~\cite{ledig2017superresolution}, photo editing~\cite{shu2017neural}, and for multi-target domains~\cite{choi2017stargan,lample2017fader,pumarola2018ganimation}. While traditional cGANs exploit the underlying conditional distribution of the data, they are constrained to produce deterministic outputs, \ie given an input and a target label, the output is always the same. In comparison, our approach introduces a style randomness in the generation process.

\vspace{-0.2cm}
\paragraph{Image-to-Image Translation (I2I)}
Isola\ETAL{isola2017pix2pix} introduced a framework in which they trained cGANs using paired datasets. This work led to a new set of previously unexplored I2I problems. Based on these findings, Zhu\ETAL{zhu2017cyclegan} extended the framework by introducing the cycle-consistency loss, which allowed to perform cross-domain mapping using unpaired datasets. Although CycleGAN~\cite{zhu2017cyclegan} is currently one of the most common backbones for I2I models and frameworks, it is constrained to one-to-one domain translation, hence it needs one generator per domain. In contrast, our method uses a single generator regardless of the number of domains.   

Other works~\cite{choi2017stargan,pumarola2018ganimation} extended the cycle-consistency insight in order to cope with multiple domains, by using a single generator. These methods take the label as independent features to the first layer of the generator, hence constraining the generator weights to restricted applications. Similarly, additional methods~\cite{lample2017fader,he2017attrgan_v1} tackled the multilabel mapping problem from a VAE-GAN~\cite{larsen2015vaegan} perspective. Our approach neither uses a variational autoencoder representation nor does it depend on label weights, since the generator has always the same number of parameters regardless of the application.

\vspace{-0.3cm}
\paragraph{Disentangled Representations}
A recurrent limitation in traditional I2I methods is their deterministic output. In image generation problems~\cite{radford2015unsupervised,chen2016infogan,kaneko2017infocgan}, disentangled representations are achieved by injecting random noise in the generator. Nevertheless, this idea cannot be used on the seminal CycleGAN, as this framework learns to ignore the noise vector due to the lack of regularization~\cite{zhu2017cyclegan}.

Recently, there have been efforts~\cite{chen2017photographic,zhu2017bicycle,bansal2017pixelnn} to produce diverse representations from a single input. For instance,
BiCycleGAN~\cite{zhu2017bicycle} bypassed the regularization issues of CycleGAN and it included a random noise vector in the training scheme, thus generating images of higher quality than CycleGAN. However, this approach requires paired data to train, which makes it unfeasible to scale in real-world scenarios. 

Furthermore, generating multimodal images can also be studied as a problem of style transfer~\cite{gatys2016image,gatys2017controlling} between two images. Inspired by the work of Gatys\ETAL{gatys2016image}, recent approaches~\cite{huang2018munit,ma2018exemplarmunit2,lee2018drit} split the generator encoder into a two-stream content and style encoder, where the content stream extracts the underlying structure, shape and main information to be preserved on the image, and the style one draws the rendering attributes it aims at transferring. These disentangled representations are similar in spirit with the CycleGAN cycle-consistency adversarial loss since they perform a cross-domain mapping for the style and content space. Consequently, it is difficult to perform fine-grained translations. In comparison, our proposed approach does not suffer in this regard, since we neither constrain the content nor the style distributions. Moreover, as the experiments will show, SMIT is suitable for both coarser translations and subtle local appearances \eg, art in-painting or facial expressions, respectively. 

\vspace{-0.3cm}
\paragraph{Continuous Interpolation}
On the one hand, Pumarola\ETAL{pumarola2018ganimation} introduced a cGAN framework that takes as input continuous rather than discrete labels. This approach enables the generation of examples with continuous labels at inference time, however, it does not handle diversity for the same input. On the other hand, for binary problems, Lee\ETAL{lee2018drit} and Huang\ETAL{huang2018munit} performed continuous interpolation between two styles in order to produce a pseudo-animated style transferring with images that belong to the same domain. Our work uses both target and style continuous interpolation.


\tref{table:related_work} summarizes our main differences with respect to the literature for either multi-label or multimodal translation. SMIT has richer capabilities that those of existing methods as we perform fine-grained local transformation, style transformation, continuous style interpolation, continuous label interpolation, and multi-label transferring using one single generator.

\section{\TITLE~(SMIT)}
Our final goal is to generate multi-attribute images with different styles using a single generator. As illustrated in \fref{fig:overview}, our method is an ensemble of three different networks: a generator, a discriminator, and a domain embedding (DE). The generator takes the source image as input and translates it. The discriminator does not only differentiate between real and fake samples, but it also approximates the output distribution of the real target by means of an auxiliary classifier. Finally, SMIT uses the DE to merge both target style and target labels into the generator.

\subsection{Problem Formulation}
Let $\mathcal{X}_{r}\in\mathbb{R}^{H\times{W}\times{3}}$ be the real image. $\mathcal{X}_{r}$ is encoded by a set of $N$ discrete or continuous labels $y_{r}\in\mathbb{R}^{N}$. Additionally, for each possible $\mathcal{X}_{r}$, there is an unknown style distribution $s_{r}\in\mathbb{R}^{S}$. Given a target label $y_{f}$, and a target style $s_{f}$, we want to learn a mapping function $\mathbb{G}$ to produce a fake image $\mathcal{X}_{f}$, without having access to the joint distribution $p(\mathcal{X}_{r}, \mathcal{X}_{f})$:
\vspace{-0.1cm}
\begin{equation}
    \mathbb{G}(\mathcal{X}_{r}, y_{f}, s_{f})\rightarrow{\mathcal{X}_{f}\in\mathbb{R}^{H\times{W}\times{3}}}
    \label{eq:smit}
\end{equation}
As it is common in cGANs~\cite{choi2017stargan,pumarola2018ganimation,chen2016infogan,radford2015unsupervised}, we have a discriminator $\mathbb{D}$ that outputs the source domain probability, \ie true or fake, and a classification/regression estimator, namely, $\mathbb{D}(\mathcal{X}_{f})\rightarrow{\{0,y_{f}\}}$ and $\mathbb{D}(\mathcal{X}_{r})\rightarrow{\{1,y_{r}\}}$.

\subsection{Model}

\paragraph{Generator ($\mathbb{G}$)}
We build upon the CycleGAN generator~\cite{zhu2017cyclegan}. It is inspired in an encoder-decoder architecture, which consists of down-sampling layers, residual blocks, and up-sampling layers. Importantly, we use Instance Normalization (IN)~\cite{dumoulin2017learned_in1,ulyanov2017improved_in2}, Adaptive Instance Normalization (AdaIN)~\cite{huang2017arbitraryadain}, and Layer Normalization (LN)~\cite{ba2016layer} for the three stages, respectively. The main reason we only use IN during the first stage and not in the up-sampling is because they introduce undesirable properties to the global mean and variance that are modified by AdaIN in the residual Layers.

\vspace{-0.2cm}
\paragraph{Domain Embedding (DE)}
We indirectly input the target attribute and the style randomness through AdaIN~\cite{huang2017arbitraryadain} weights. AdaIN normalization is computed from \eref{eq:adain}, where $x$ is the input and $z$ are the adaptive parameters. 

\vspace{-0.6cm}
\begin{gather}
    AdaIN(x, z) = z_{w}\frac{x - \mu(x)}{\sigma(x)} + z_{b} 
    \label{eq:adain}
    \\
    z = DE(y, s)
    \label{eq:de}
\end{gather}
\vspace{-0.3cm}

As the AdaIN parameters depend entirely on the number of feature maps of the input $x$, they are agnostic to both style and label domains, which makes the generator entirely label and style independent. This key property makes SMIT highly suitable for transfer learning, addressing a drawback of cGANs in real-world scenarios.

It is important to mention that since the style and label dimensions may differ from the $z$ dimensions, we use a projection embedding representation to encode style and label inputs to a fixed size suitable for AdaIN (\eref{eq:de}).

We remark that the DE does not require any training scheme, instead it is inspired by Language Modeling methods~\cite{margffoy2018dmn,devlin2018bert,liu2017recurrent,mccann2017learned,peters2018deep} that uses random initialization to map the input to a space embedding distribution. Particularly, we use a simple random embedding, \ie a fully connected layer to map from style and labels concatenation to the AdaIN parameters. Our rationale is as follows: By always ensuring different $z$, we guarantee different normalization parameters, which means different fake images. We study the DE behaviour in more detail in \sref{results_delearning}.

\vspace{-0.2cm}
\paragraph{Discriminator ($\mathbb{D}$)}
As previously stated, the discriminator has two outputs: source domain (\textit{src}) and auxiliary classifier (\textit{cls}). First, we use the idea of patch-GAN~\cite{isola2017pix2pix}, to tell whether the source is fake or true based on a patch rather than a single number ($\mathbb{D}_{src}$). Second, we have a binary cross entropy loss function for the conditional labels ($\mathbb{D}_{cls}$). If continuous labels are used, then a regression objective loss should be applied. However, as we will discuss \sref{sec:interpolation}, our approach is capable of generating continuous labels even if it was trained with discrete ones.

\vspace{-0.2cm}
\subsubsection{Training Framework}
In order to approximate function $\mathbb{G}$ in \eref{eq:smit}, we split our general loss function for clarity.

\vspace{-0.3cm}
\paragraph{Adversarial Loss}
We use the recently introduced averaged Relativistic Adversarial Loss (RGAN)~\cite{jolicoeur2018ragan} and the hinge version~\cite{miyato2018spectral} loss to train the adversarial loss. RGAN relies on the idea that the discriminator not only estimates whether images are real or fake, but it also estimates the probability that the given real images are more realistic than the fake ones. 
\vspace{-0.2cm}
\begin{gather}
    L_D = \mathbb{D}_{src}(\mathcal{X}_{r}) - ||\mathbb{D}_{src}(\mathcal{X}_{f})||_{1} \nonumber
    \\
    L_G = \mathbb{D}_{src}(\mathcal{X}_{f}) - ||\mathbb{D}_{src}(\mathcal{X}_{r})||_{1} \nonumber
    \\
    \mathcal{L}_{adv} = L_D + L_G \label{eq:adversarial}
\end{gather}
\vspace{-0.9cm}

\paragraph{Conditional Loss}
The adversarial loss does not include any regularization for the conditional labels, yet the generator must be able to produce both realistic and conditioned images. To solve this issue, we define the conditional loss as:
\vspace{-0.2cm}
 \begin{align}
    \mathcal{L}_{cls} = \mathbb{D}_{cls}(\mathcal{X}) \log(y) + (1-\mathbb{D}_{cls}(\mathcal{X})) \log (1-y)
    \label{eq:classification}
\end{align}

\vspace{-0.4cm}
\paragraph{Recovery Loss}
In order to produce $\mathcal{X}_{f}$, we jointly input the target label and the target style. Therefore, the cycle consistency loss employed to recover the original image can be naively defined as:
\vspace{-0.2cm}
 \begin{align}
    \mathcal{X}_{r} \approx \mathcal{X}_{rec} = \mathbb{G}(\mathbb{G}(\mathcal{X}_{r}, y_{f}, s_{f}), y_{r},s_{r}) \nonumber
\end{align}
Note that the original style ($s_{r}$) is an unknown parameter. Nonetheless, we assume that $s_{r}$ is drawn from a known normal distribution, and therefore reformulate the reconstruction loss by adding a different random style $s_{f}'$. We assume random styles during the whole training process. Thus, we compute the reconstruction or cycle consistency loss as:
\vspace{-0.2cm}
\begin{gather}
    \mathcal{X}_{rec} = \mathbb{G}(\mathbb{G}(\mathcal{X}_{r}, y_{f}, s_{f}), y_{r},s_{f}') \nonumber
    \\
    \mathcal{L}_{rec} = ||\mathcal{X}_{r} - \mathcal{X}_{rec}||_{1}
    \label{eq:reconstruction}
\end{gather}

\vspace{-0.7cm}
\paragraph{Attention Loss}
Until this point, there is no guarantee that the output of our generator will preserve background details \eg, the underlying structure, or the identity of a person. To solve this particular issue, we regularize our model with the unsupervised attention mechanism proposed by Pumarola\ETAL{pumarola2018ganimation}. We add a new and parallel layer to the generator output ($\mathcal{X}_{f}$) that works as the attention mask ($\mathcal{M}$).

The attention loss encourages fake images to change only certain regions with respect to the real input, and it is decomposed by the following terms: 
\vspace{-0.2cm}
\begin{gather}
    [\mathcal{X}_{f}\in\mathbb{R}^{H\times{W}\times{3}},~\mathcal{M}\in\mathbb{R}^{H\times{W}}] = \mathbb{G}(\mathcal{X}_{r}, y_{f}, s_{f})
    \nonumber
    \\ 
    \mathcal{X}_{f}=\mathcal{M} \boldsymbol{\cdot} \mathcal{X}_{r} + (1 - \mathcal{M}) \boldsymbol{\cdot} \mathcal{X}_{f}
    \nonumber
    \\
    \mathcal{L}_{attn} = ||\mathcal{M}||_{1}
    \label{eq:attention}
\end{gather}  
\vspace{-0.7cm}

\vspace{-0.2cm}
\paragraph{Identity Loss}
To further stabilize the training framework, we regularize our model with the identity loss that is defined as follows: 
\vspace{-0.2cm}
\begin{align}
    \mathcal{L}_{idt} = ||\mathcal{X}_{r} - (\mathbb{G}(\mathcal{X}_{r}, y_r, s_{f}''))||_{1}
    \label{eq:identity}
\end{align}

\vspace{-0.7cm}
\paragraph{Overall Loss}
We define our full objective function in \eref{eq:loss}, as the weighed sum of the previous losses:
\vspace{-0.2cm}
\begin{equation}
    \mathcal{L} = \lambda_{adv}\mathcal{L}_{adv} + \lambda_{cls}\mathcal{L}_{cls} + \lambda_{rec}\mathcal{L}_{rec} 
    + \lambda_{attn}\mathcal{L}_{attn}
    + \lambda_{idt}\mathcal{L}_{idt}
    \label{eq:loss}
\end{equation}  

Remarkably, our method does not require style regularization~\cite{huang2018munit,lee2018drit} since we use a training framework that can easily bypass it.

\section{Experimental Setup}

\begin{figure*}[t]
\vspace{-0.2cm}
\begin{center}
   \includegraphics[height=5cm,width=0.9\linewidth]{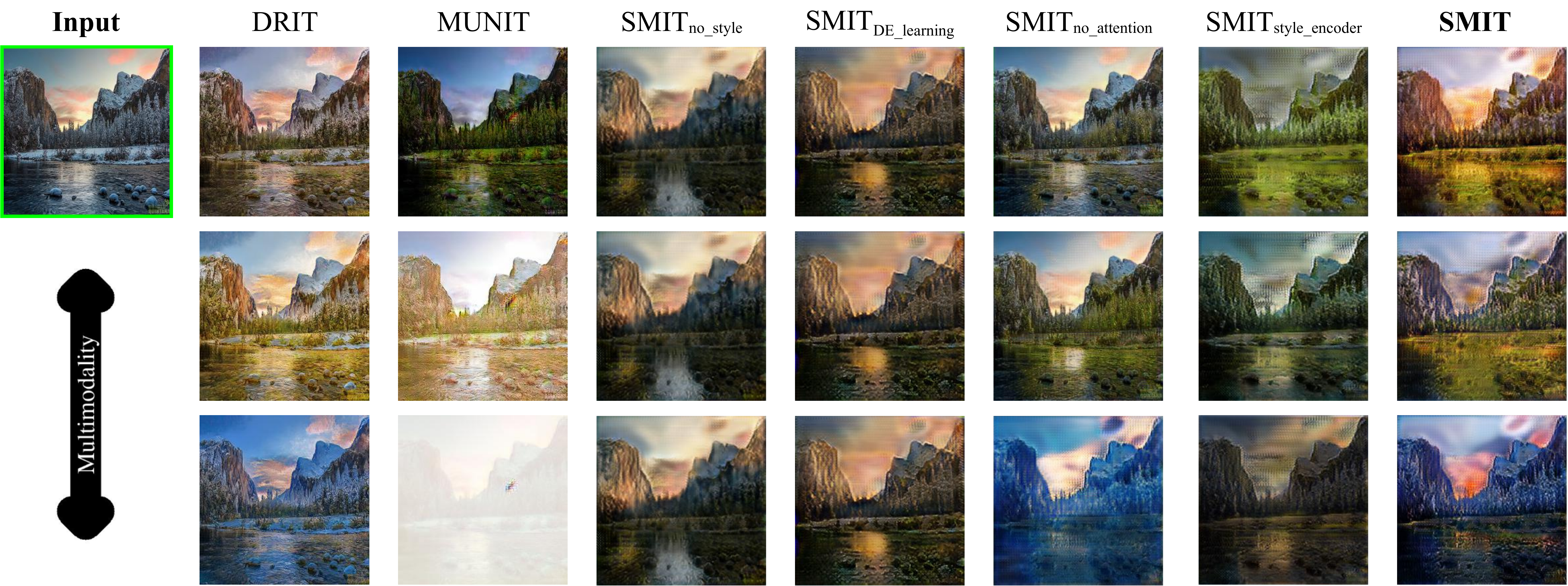}
\end{center}
\vspace{-0.1cm}
   \caption{\textbf{Ablation experiments.} Qualitative comparisons over the Yosemite dataset~\cite{isola2017pix2pix}. Given the same input, we report the output for the related work~\cite{lee2018drit,huang2018munit} and for each ablation experiment. Each row depicts different styles.}
\label{fig:ablation}
\vspace{-0.3cm}
\end{figure*}

\label{sec:results}
We validate our method over several and very different datasets and tasks, such as instance facial synthesis~\cite{liu2015CelebA}, emotion recognition~\cite{langner2010RafD}, Yosemite summer$\leftrightarrow$winter~\cite{isola2017pix2pix}, and edges-to-object generation~\cite{isola2017pix2pix}.

In the supplementary material, we extend our qualitative results to painters~\cite{anoosheh2018combogan}, Alps seasons~\cite{anoosheh2018combogan}, RafD~\cite{langner2010RafD}, BP4D~\cite{zhang2014bp4d}, EmotionNet~\cite{fabian2016emotionet}, and full CelebA~\cite{liu2015CelebA} with 40 attributes.


\subsection{Evaluation Metrics}
\paragraph{Diverse Translation}
The LPIPS metric~\cite{zhang2018lpisp} allows us to quantify the similarity between two different images. LPIPS computes the L2 distance between pairs of deep features (\eg, AlexNet, VGG, etc) images.

\paragraph{Multi-label Translation}
Besides the LPIPS score, we also compute
the Inception Score (IS)~\cite{salimans2016improved} that is a popular score for I2I problems. The IS employs an Inception Network~\cite{szegedy2016rethinking} to classify fake images and thus rank them according to their scores with respect to the prior distribution. Additionally, we report the Conditional Inception Score (CIS)~\cite{huang2018munit} that quantifies both high quality and diverse mapping.

\subsection{Evaluation Framework}
Given the unique nature of our approach, we unfold the quantitative evaluation into two different schemes: multimodal evaluation, and multi-label evaluation.

\vspace{-0.4cm}
\paragraph{Multimodal Evaluation}
We directly use MUNIT~\cite{huang2018munit} and DRIT~\cite{lee2018drit} to compare our method in GAN-based disentangled representations. For fair comparison under this setting, we work within the same datasets Edges~\cite{isola2017pix2pix} and Yosemite~\cite{zhu2017cyclegan}. To this end, we train MUNIT and DRIT and report the corresponding LPIPS over the whole test set. 

We use the LPIPS score to measure the diversity of the generated images. As there is no standard evaluation framework for the diversity in GAN-based problems, we use a set of two metrics. First, as in MUNIT, we compute the diversity one-vs-all across the entire dataset (D), using the diversity in the real data as a reference. Then, we use one single fixed style to produce the cross-mapping in order to compute the diversity along the entire fake dataset. Second, as in DRIT, given a single image, we measure the partial diversity (PD) across different modalities (20 different styles) and report the average and standard deviation over each image, over the whole set.

\vspace{-0.4cm}
\paragraph{Multi-label Evaluation}
Additionally, for purely multi-label I2I methods, we train an Inception network~\cite{szegedy2016rethinking} on a RafD train set (90\%) and report the IS and CIS over the remaining test set (10\%). We retrain StarGAN and GANimation~\cite{pumarola2018ganimation} under exactly the same settings in order to make a fair comparison.

\subsection{Implementation Details}
We use an ensemble of three different convolutional networks: Generator, Discriminator, and a Domain Embedding (DE). 

Similar to previous methods~\cite{huang2018munit,lee2018drit}, we assume the style to be drawn from a prior Gaussian distribution with 0 mean and identity variance, namely $\mathcal{N}(0,I)$. Therefore, the DE takes this 20-dimensional style vector and the $N$-dimensional target domain (one hot encoded) as inputs to produce the corresponding AdaIN number of parameters.


We provide a more detailed description of the architecture of our networks and training details in the supplementary material.

\section{Results}
We quantitatively and qualitatively demonstrate the effectiveness of SMIT in several settings. First, we perform ablation experiments, then we show qualitative results over different datasets, and finally we perform an extensive quantitative evaluation and compare our results against the state-of-the-art. 

\subsection{Ablation Study}
We establish different baselines that define the main components of our framework: DE learning, removing the style randomness, adding style regularization, and removing the attention mechanism. We perform a qualitative and quantitative comparison for each of them, and we report our findings in \fref{fig:ablation} and \tref{table:ablation}, respectively.

\vspace{-0.3cm}
\paragraph{DE learning} \label{results_delearning}
Studying DE parameters is one of our main interests as it is the only controller between the style and labels, and the mapped image. We observe that the generator can easily fall in mode collapse if the DE weights are learned, thus producing almost the same images for different styles. In order to overcome this problem, we analyze the DE contribution to the general system either with learned or fixed random parameters. As we can see in \fref{fig:ablation}, SMIT\textsubscript{DE\_learning}, learning the DE parameters leads to full mode collapse, since the style has a negligible impact on the AdaIN generator parameters. This behaviour is due to the fact that the gradients that come from the auxiliary classifier force the domain embedding to produce stable outputs, and therefore the same output thanks to the lack of specialized and per domain style regularization. Conversely, by establishing fixed weights on the DE, we guarantee diversity, \ie, from \eref{eq:adain} we observe that for different \textit{scale} and \textit{bias}, we ensure different behaviour on the normalization, hence different outputs. 

\begin{figure*}[t]
\vspace{-0.2cm}
\begin{center}
   \includegraphics[trim={0 0 0 3.9cm}, clip, width=0.9\linewidth]{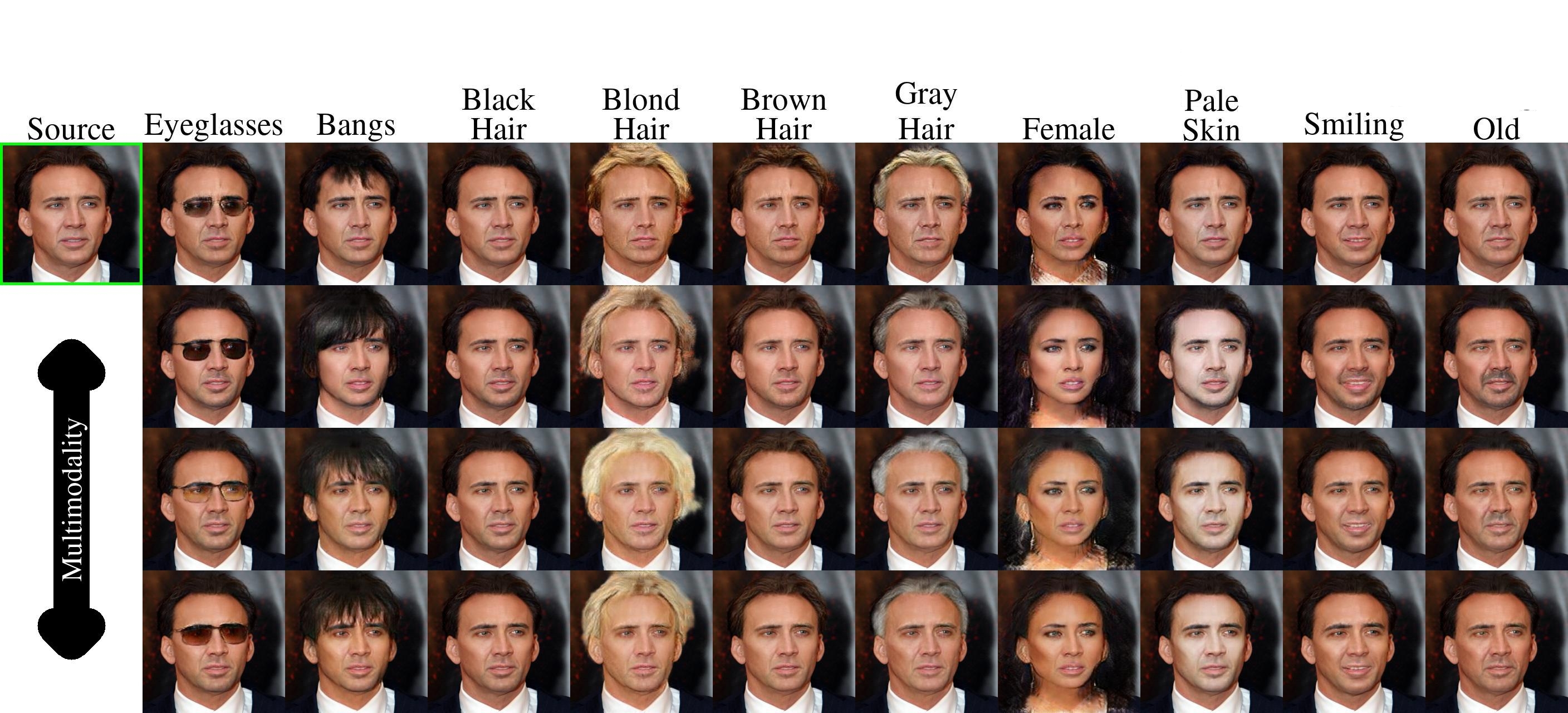}
   \caption{\textbf{Qualitative results for facial analysis mapping~\cite{liu2015CelebA}.} Example results for an image in the wild. For each attribute (column), we show the corresponding translation for four different modalities (rows).}
\label{fig:qualitative}
\end{center}
\vspace{-0.5cm}
\end{figure*}

\begin{table}
\setlength{\tabcolsep}{4.6pt}
\begin{center}
\resizebox{0.8\linewidth}{!}{
\begin{tabular}{|c||c|c|}
\hline
& \multicolumn{2}{c|}{Yosemite~\cite{isola2017pix2pix}} \\
\cline{2-3}
 & D & PD \\
\hline
\rowcolor[HTML]{\colortable} 
SMIT\textsubscript{no\_style}            & 0.412$\pm$0.046 & - \\
SMIT\textsubscript{DE\_learning}         & 0.413$\pm$0.044 & 0.004$\pm$0.003 \\
\rowcolor[HTML]{\colortable} 
SMIT\textsubscript{no\_atention}         & 0.406$\pm$0.041 & 0.105$\pm$0.071 \\
SMIT\textsubscript{style\_encoder}       & 0.418$\pm$0.043 & 0.133$\pm$0.063 \\
\rowcolor[HTML]{\colortable} 
\hline
\textbf{SMIT}              & \textbf{0.419$\pm$0.048} & \textbf{0.145$\pm$0.072} \\
\hline
\end{tabular}
}
\caption{\textbf{Ablation quantitative evaluation.} We report the diversity (D) and the partial diversity (PD) for every ablation study in our method.}
\label{table:ablation}
\end{center}
\vspace{-0.8cm}
\end{table}

Among all the datasets~\cite{liu2015CelebA,isola2017pix2pix,fabian2016emotionet,anoosheh2018combogan,zhu2017cyclegan,langner2010RafD} in which we validate our system, we observed that for datasets that contain a small number of samples with only two different domains (\eg, Yosemite~\cite{zhu2017cyclegan}, $\sim$1k images per domain), there is a decline in the quality of the fake images when the DE has fixed random parameters. More precisely, even though the auxiliary classifier is highly confident after a few iterations, and the generator learns to fool the discriminator, the generator produces pixelation in the images. Nevertheless, given the simplicity of the dataset, pixelated images fulfill the conditions to fool the discriminator, \ie fake images are realistic enough and they fall into the statistical representation of the labels. We further study this behaviour by combining two different settings: DE training (no pixelation and deterministic) and DE fixed (pixelation and stochastic). We split the AdaIN parameters into different small networks with different behaviours (learned or fixed weights), which share the input (target domain and style). We found that learning either small or big parts of the AdaIN layers induces mode collapse to the whole system. Nonetheless, with enough training iterations the pixelation issue is nuanced and not too evident. Surprisingly, we observed that the partial diversity metric (PD) is higher for highly pixelated images than smooth yet diverse ones. This finding indicates that the partial diversity is not related to both quality and diversity, but only to diversity at any cost (\eg, change in color, pixelation, etc). 

{\small
\begin{table*}
\vspace{-0.1cm}
\begin{center}
\resizebox{\linewidth}{!}{
\begin{tabular}{|c||c|c||c|c||c|c||c|}
\hline
 & \multicolumn{2}{c||}{Edges2Shoes~\cite{isola2017pix2pix}} & \multicolumn{2}{c||}{Edges2Handbags~\cite{isola2017pix2pix}} & \multicolumn{2}{c||}{Yosemite~\cite{isola2017pix2pix}} & \# Parameters \\
              \cline{2-7}
              & D & PD & D & PD & D & PD & (Generator) \\
\hline
\rowcolor[HTML]{\colortable} 
CycleGAN~\cite{zhu2017cyclegan}       & {0.272$\pm$0.048} & - & {0.293$\pm$0.081} & - & 0.272$\pm$0.048 & - & 2x11.4M \\
DRIT~\cite{lee2018drit}           & {0.237$\pm$0.149} & {0.028$\pm$0.030} & {0.296$\pm$0.181} & {0.056$\pm$0.060} & {0.398$\pm$0.038} & 0.126$\pm$0.019 & 2x21.3M \\
\rowcolor[HTML]{\colortable} 
MUNIT~\cite{huang2018munit}          & {0.295$\pm$0.051} & {0.077$\pm$0.057} & {0.365$\pm$0.052} & {0.123$\pm$0.067} & 0.335$\pm$0.045 & {0.208$\pm$0.034} & 2x15.0M \\
\textbf{SMIT (ours)}  & {0.303$\pm$0.058} & {0.072$\pm$0.056} & {0.367$\pm$0.048} & {0.096$\pm$0.072} & \textbf{0.437$\pm$0.041} & {0.145$\pm$0.072} & \textbf{8.4M} \\
\hline
Real Data      & {0.313$\pm$0.052} & - & {0.374$\pm$0.051} & - & {0.447$\pm$0.049} & - & - \\
\hline
\end{tabular}
}
\caption{\textbf{Multimodal quantitative evaluation.} We report the LPIPS score to compare the diversity (D) and partial diversity (PD) with respect to the multimodal approaches. Better results are boldfaced according to their significant values.}
\label{table:resultsD}
\end{center}
\vspace{-0.5cm}
\end{table*}
}

This form of coupling style and domain information is in line with~\cite{li2017demystifying,gatys2016image} as to use global statistics is better suited for the purpose of style transferring, rather than spatially connected features (\eg, concatenating the image and the labels) as other methods usually employ~\cite{lee2018drit,choi2017stargan,pumarola2018ganimation}.

\paragraph{No Style}
By removing the style, our network behaves on a fully deterministic way since fixed labels always impose the same statistics over the generator (\fref{fig:ablation} and \tref{table:ablation}, SMIT\textsubscript{no\_style}). 

\vspace{-0.3cm}
\paragraph{Style Encoder}
MUNIT~\cite{huang2018munit} and DRIT~\cite{lee2018drit} share a common practice by using a style encoder, where they regularize the style or noise previously injected in the generator. We also evaluate the necessity of such a mechanism. To this end, we deploy a separate network for style encoding ($\mathbb{S}$), whose purpose is to extract the style that is injected to fake images, \ie computing $s_{f}'\approx\mathbb{S}(\mathcal{X}_{f})$. As we depict in \fref{fig:ablation} (SMIT\textsubscript{style\_encoder}) and \tref{table:ablation}, there are no qualitative or quantitative differences by using this regularizer. However, the style encoder is a different network as big as the discriminator, so it increases the training time and memory consumption. Moreover, we argue that having a fixed random embedding as DE is enough to produce diversity because we force the generator to always produce different images regardless of the lack of regularization in the style. Therefore, the style encoder is not performing a critical role within our system. It is worth noting that the style encoder in conjunction with the DE-training has no effect on the diversity.

Due to the nature of multi-label problems, the style regularization is unhelpful in its simple form because of the high label entanglement. Thus, for any style encoding, it would require different styles for different labels using as many domain embeddings as domains, and perform cycle-consistency in a way that styles are tied to labels, which is difficult in practice.

\paragraph{Attention Mask}
We observe that the attention mechanism plays a critical role for the entire training scheme for those fine-grained datasets \eg, CelebA, EmotionNet, BP4D. Without this loss, our framework takes the easiest way in the translation process, \ie uniformly changing the color of the input (\fref{fig:ablation}, MUNIT). We argue that with enough iterations, this undesirable property leads to higher partial diversity due to diversity in color.
\\\\
Furthermore, our Domain Embedding differs from the Multi-Layer Perceptron (MLP) proposed by MUNIT~\cite{huang2018munit} as they use domain-specific yet trainable networks in order to transform from the style vector representation to the AdaIN number of parameters, which prevents the mode collapse problem. Note that we only use a single Domain Embedding regardless the multi-domain nature. 

\subsection{Qualitative Results}
We now proceed to highlight the SMIT capabilities over the CelebA dataset. In \fref{fig:qualitative}, we demonstrate the effectiveness of our method for 10 different attributes, switching one attribute at a time (columns) for different styles (rows). From these transformations, we observe that our model is indeed learning a fully continuous representation for the attributes, as it generalizes across different modalities either for subtle or broader transformations such as eyeglasses or smiling, or gender or hair colors, respectively. Similarly, in the supplementary material we depict different emotion translations and compare against state-of-the-art methods.

Moreover, in the supplementary material, we show that, given fixed style and fixed labels, our model is able to generate always the same attributes for different people, \ie the same eyeglasses, bangs, etc. We also report the attention mask visualizations. Additionally, we show translations for painters~\cite{anoosheh2018combogan}, Alps~\cite{anoosheh2018combogan}, RafD~\cite{langner2010RafD}, edges2objects~\cite{isola2017pix2pix}, BP4D~\cite{zhang2014bp4d}, EmotionNet~\cite{fabian2016emotionet}, and full CelebA~\cite{liu2015CelebA} datasets. We also depict qualitative differences with StarGAN, GANimation, and FaceApp~\cite{faceapp} over the CelebA dataset. 

\begin{table}
\vspace{-0.3cm}
\begin{center}
\resizebox{\linewidth}{!}{
\begin{tabular}{|c||c|c|c|c|}
\hline
& \multicolumn{4}{c|}{RafD~\cite{langner2010RafD}} \\
\cline{2-5}
 & CIS & IS & D & PD \\
\hline
\rowcolor[HTML]{\colortable} 
StarGAN~\cite{choi2017stargan}          & 1.00$\pm$0.00 & 1.66$\pm$0.38 & 0.15$\pm$0.01 & - \\
GANimation\cite{pumarola2018ganimation}       & 1.00$\pm$0.00 & 1.51$\pm$0.33 & 0.16$\pm$0.01 & - \\
\rowcolor[HTML]{\colortable} 
\textbf{SMIT (ours)}  & \textbf{1.25$\pm$0.06} & \textbf{2.51$\pm$0.70} & \textbf{0.17$\pm$0.01} & \textbf{0.004$\pm$0.001} \\
\hline
Real Data       & - & 1.18$\pm$0.18 & 0.16$\pm$0.01 & - \\
\hline
\end{tabular}
}
\caption{\textbf{Multi-label quantitative evaluation.} We report the results for Inception Score (IS), Conditioned Inception Score (CIS), and LPIPS diversity metric (D and PD), for multi-label frameworks.}
\label{table:resultsML}
\vspace{-0.8cm}
\end{center}
\end{table}

\paragraph{Interpolations} \label{sec:interpolation}
Following common practice within Multimodal Image-to-Image translation methods~\cite{lee2018drit,huang2018munit}, where we assume that each style is randomly sampled from a normal probability distribution, our method also benefits from style interpolation going from one style to another by performing a spherical interpolation. 

Even though the labels are binary attributes at train time, the DE transforms them into a higher dimensional representation given by the number of channels in the AdaIN layers. Inserting the labels in the form of continuous labels into the generator is of importance as we can easily perform continuous inference before the DE. The absence or presence of any label is correlated with different representations in the AdaIN parameters.

In the supplementary material, we show visualizations for style and label interpolation. Note that we do not explicitly train with continuous labels.

\subsection{Quantitative Results}
Next, we quantitatively compare SMIT with respect to the literature. We separate our experiments into two strategies due to the lack of both multi-label and multimodal translation methods.

\paragraph{Multimodal Evaluation}
As we depict in \tref{table:resultsD}, we compare directly with DRIT and MUNIT over \textit{edges2shoes}, \textit{edges2handbags} and \textit{Yosemite} datasets. 

Our method produces higher LPIPS for the entire test set (D), and competitive results across partial diversity (PD) with respect to the state-of-the-art since there is no significant differences with MUNIT. We hypothesize that MUNIT's~\cite{huang2018munit} good performance in the PD score is because this method is focused on color transformation and rendering rather than texture or content (\fref{fig:ablation}, MUNIT column). MUNIT constrains the content latent space, producing thus highly diverse mappings across a batch, and low general diversity if the style is fixed. As we retrain DRIT and MUNIT, it is worth to mention that DRIT's poor performance on edges2shoes and edges2handbags is due to the lack of diversity for object$\rightarrow$edge mapping. 

Remarkably, due to the reduced number of parameters (\tref{table:resultsD}, number of parameters), SMIT takes less computational resources than baseline approaches to training, that is SMIT fits four times the batch size used in DRIT~\cite{lee2018drit} and MUNIT~\cite{huang2018munit}, using one Titan X GPU.

We provide more quantitative results for each domain independently in the supplementary material.

\paragraph{Multi-label Evaluation}
\tref{table:resultsML} shows our results for StarGAN, GANimation, and SMIT. For each image, we perform 7 different translations (ignoring the ground truth translation). As we expected, StarGAN and GANimation obtain a constant CIS (1.0) and high IS scores, which indicates their lack of diversity but good qualitative translations. SMIT significantly overcomes related methods in diversity and image quality. Note that SMIT also outperforms the IS and D for the real images, demonstrating thus the effectiveness in both quality and diversity beyond the original dataset. In the supplementary material, we discriminate CIS and IS over each label independently.

Even though StarGAN and GANimation use a single generator and share a similar number of parameters, it is important to remark that they reshape the label vector into the input image size. This issue arises in high-resolution image to image translation as neither the number of parameters nor the computational time are negligible. By contrast, SMIT is suitable either for low or high resolution as it is label-agnostic dependent.

\section{Conclusions}

In this paper, we presented a novel, robust yet simple method for automatically performing stochastic image-to-image translation for multiple domains using a single generator. We demonstrated the capability of our approach with respect to the state-of-the-art in both disentangled and multi-label scenarios by achieving jointly high quality and diversity representations for both coarse or fine-grained translations. Moreover, SMIT is directly suitable for either multimodal interpolation or continuous interpolation in style and label intensity domains, respectively. 

{\small

}

\clearpage
\begin{appendices}
\appendix

\section*{\large{SMIT: \TITLE~-~Supplemental Material}}

\section{Network Architecture} 
\label{appendix:network}
We show SMIT network architectures for the generator, domain controller, and discriminator in \tref{table:generator}, \ref{table:domain_embedding}, and \ref{table:discriminator}, respectively.

\subsection{Generator}
The generator is an ensemble of down-sampling, residual and up-sampling stages. Each block is composed by a tuple of convolution, normalization and ReLU layers. The normalization layer can take the form of any of the following terms: None, instance normalization (IN), Adaptative Instance Normalization (AdaIN), or Layer Normalization (LN). 

\subsection{Domain Embedding}
The Domain Embedding is a target domain and style embedding projection using a fully connected layer. First, the input is the concatenation of the target domain ($\mathbb{N}_d$) and the target random style ($\mathbb{S}$) that we set 20-dimensional. Second, the output is the corresponding AdaIN number of parameters in the generator. Therefore, as we have 6 residual blocks in the generator, and two AdaIN layers per residual block, then we have 6.144 AdaIN fixed parameters.

\subsection{Discriminator}
Resembling the generator blocks, the discriminator also has three layers per block: convolution, normalization and Leaky ReLU (LReLU) with a negative slope of 0.01. Each layer is normalized with Spectral Normalization (SN).

For the Multi-scale discriminator, we use three different image sizes: 256, 128, and 64. The three networks only differ in the last hidden output: for each discriminator we enforce that the last hidden layer has $2\times2$ output size

\section{Additional Qualitative Results}
\label{appendix:qualitative_results}
\maxdeadcycles=200
\extrafloats{100}

\begin{figure*}[t!]
\begin{center}
   \includegraphics[width=0.8\linewidth]{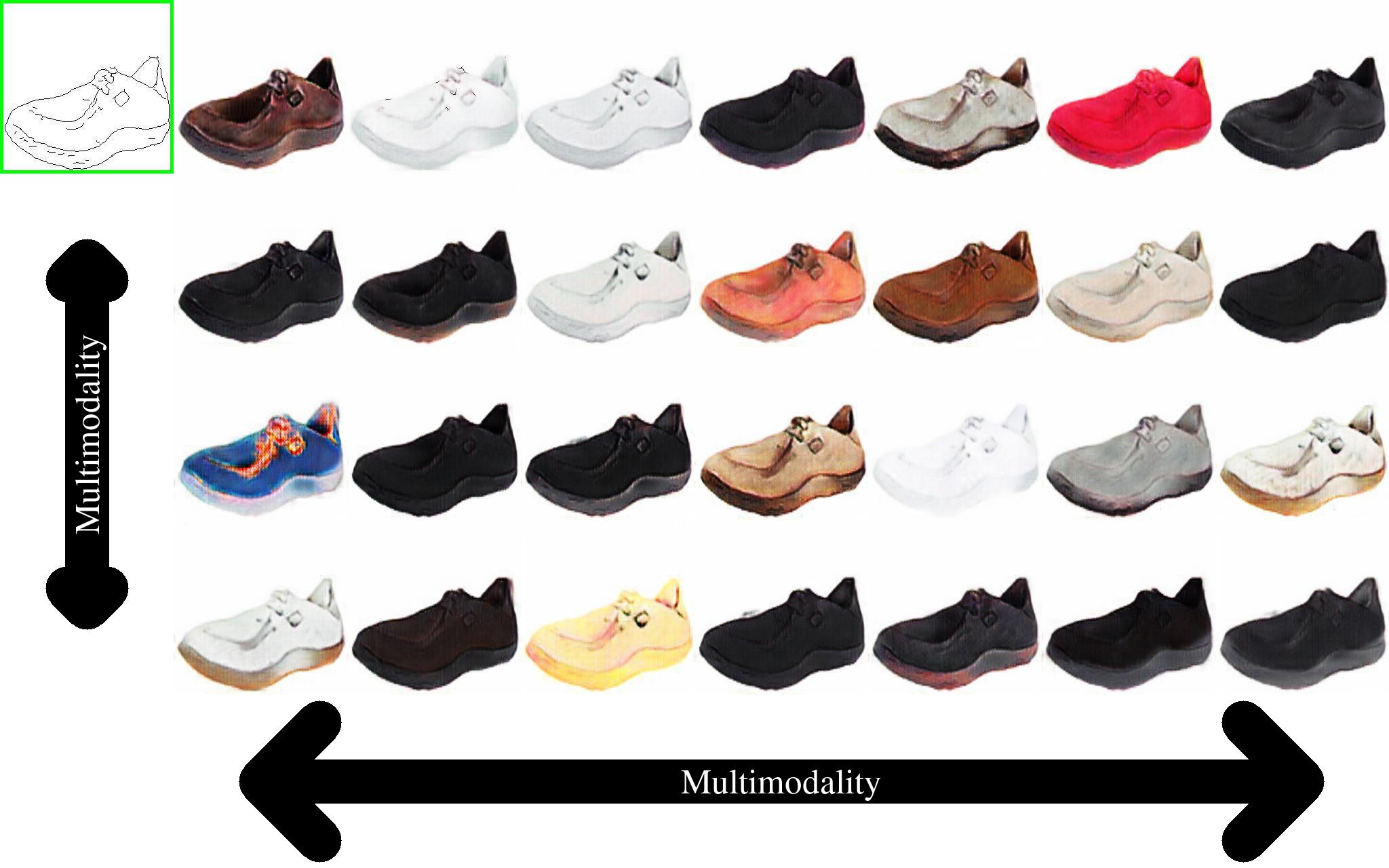}
\end{center}
   \caption{\textbf{SMIT qualitative results for edges2shoes.} Using a sketch from the test set as input (green box), we display diverse outputs across rows and columns.}
\label{fig:shoes0}
\end{figure*}

\begin{figure*}[t!]
\begin{center}
   \includegraphics[width=0.8\linewidth]{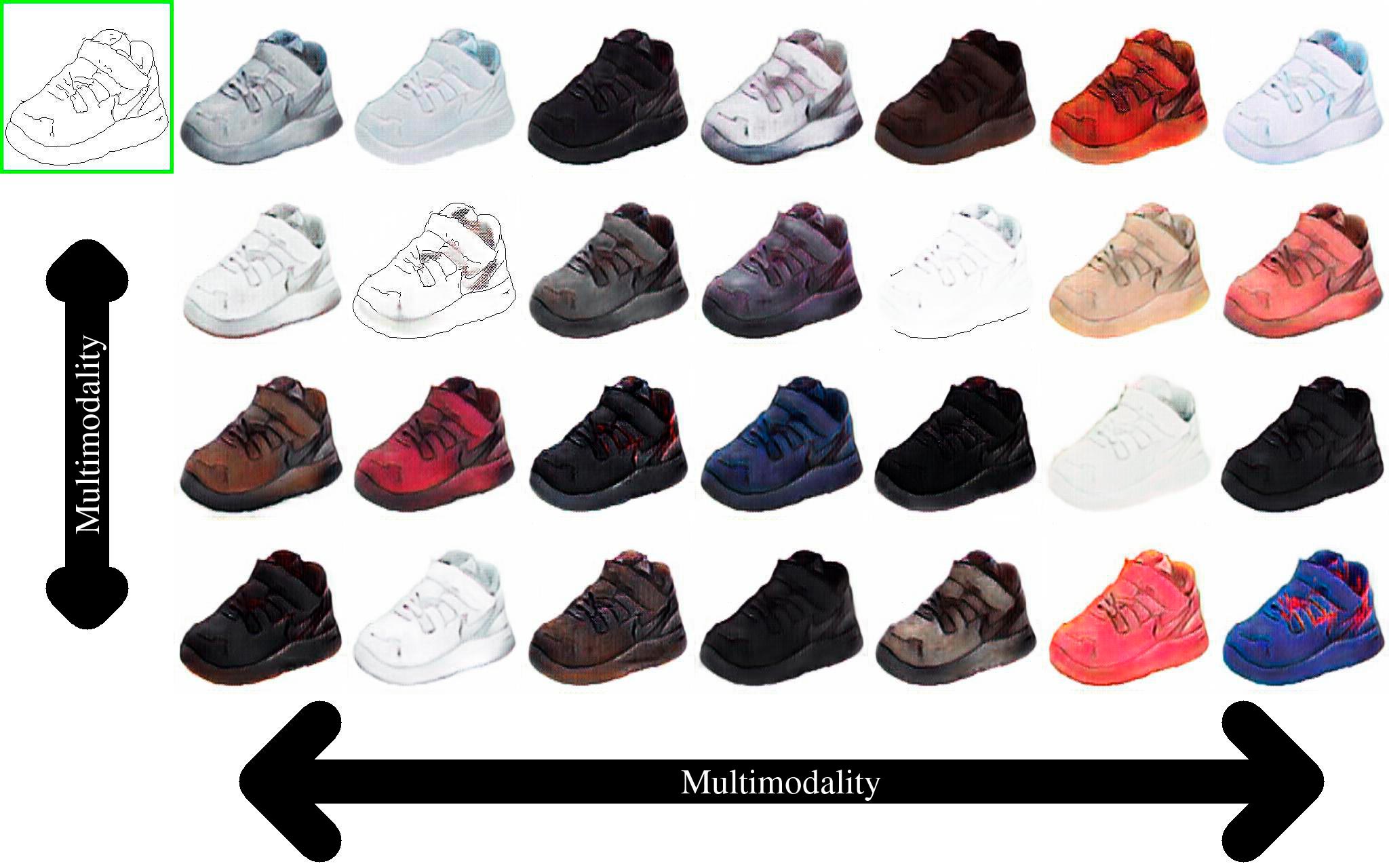}
\end{center}
   \caption{\textbf{SMIT qualitative results for edges2shoes.} Using a sketch from the test set as input (green box), we display diverse outputs across rows and columns.}
\label{fig:shoes1}
\end{figure*}

\begin{figure*}[t!]
\begin{center}
   \includegraphics[width=0.8\linewidth]{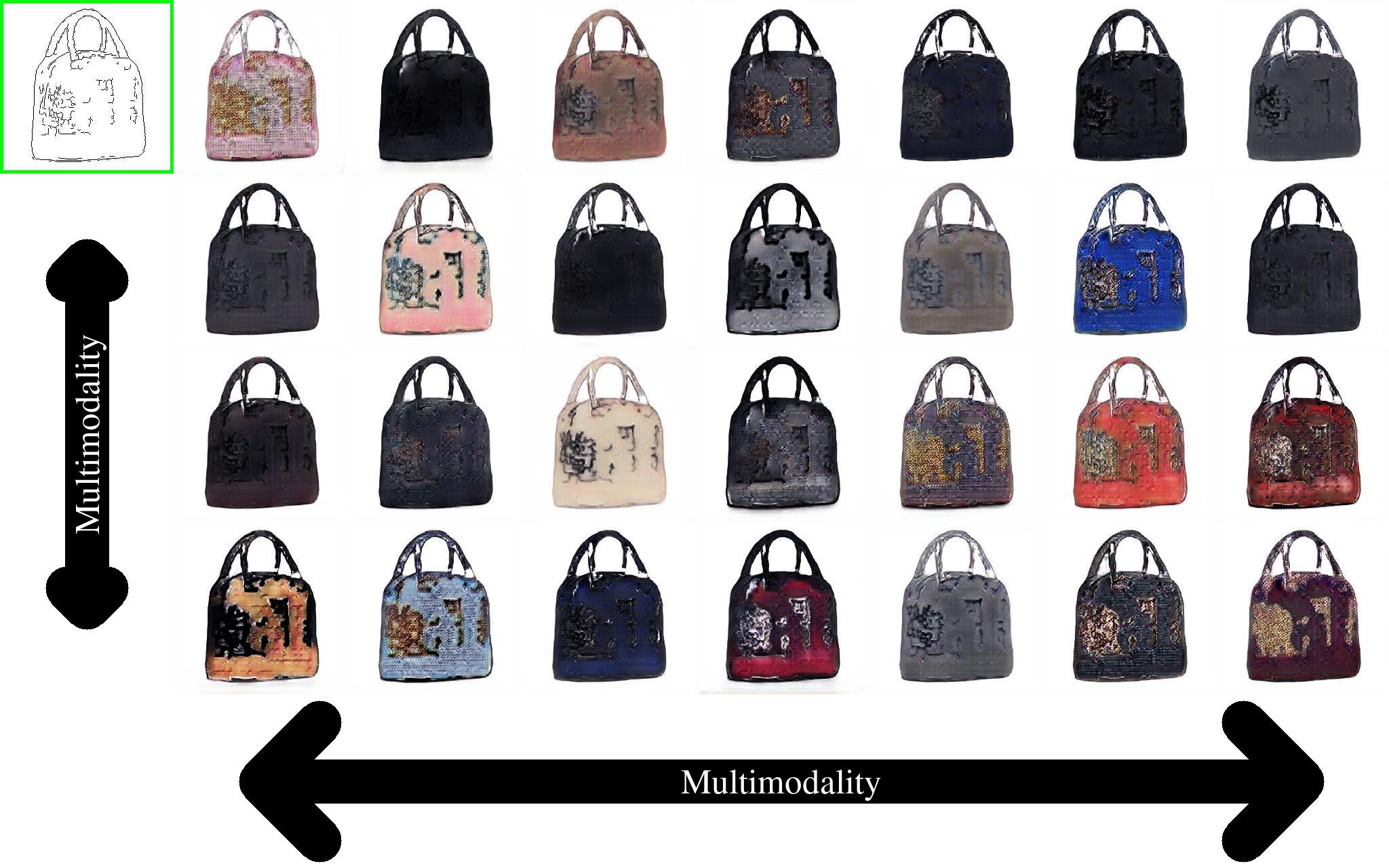}
\end{center}
   \caption{\textbf{SMIT qualitative results for edges2handbags.} Using a sketch from the test set as input (green box), we display diverse outputs across rows and columns.}
\label{fig:handbags0}
\end{figure*}

\begin{figure*}[t!]
\begin{center}
   \includegraphics[width=0.8\linewidth]{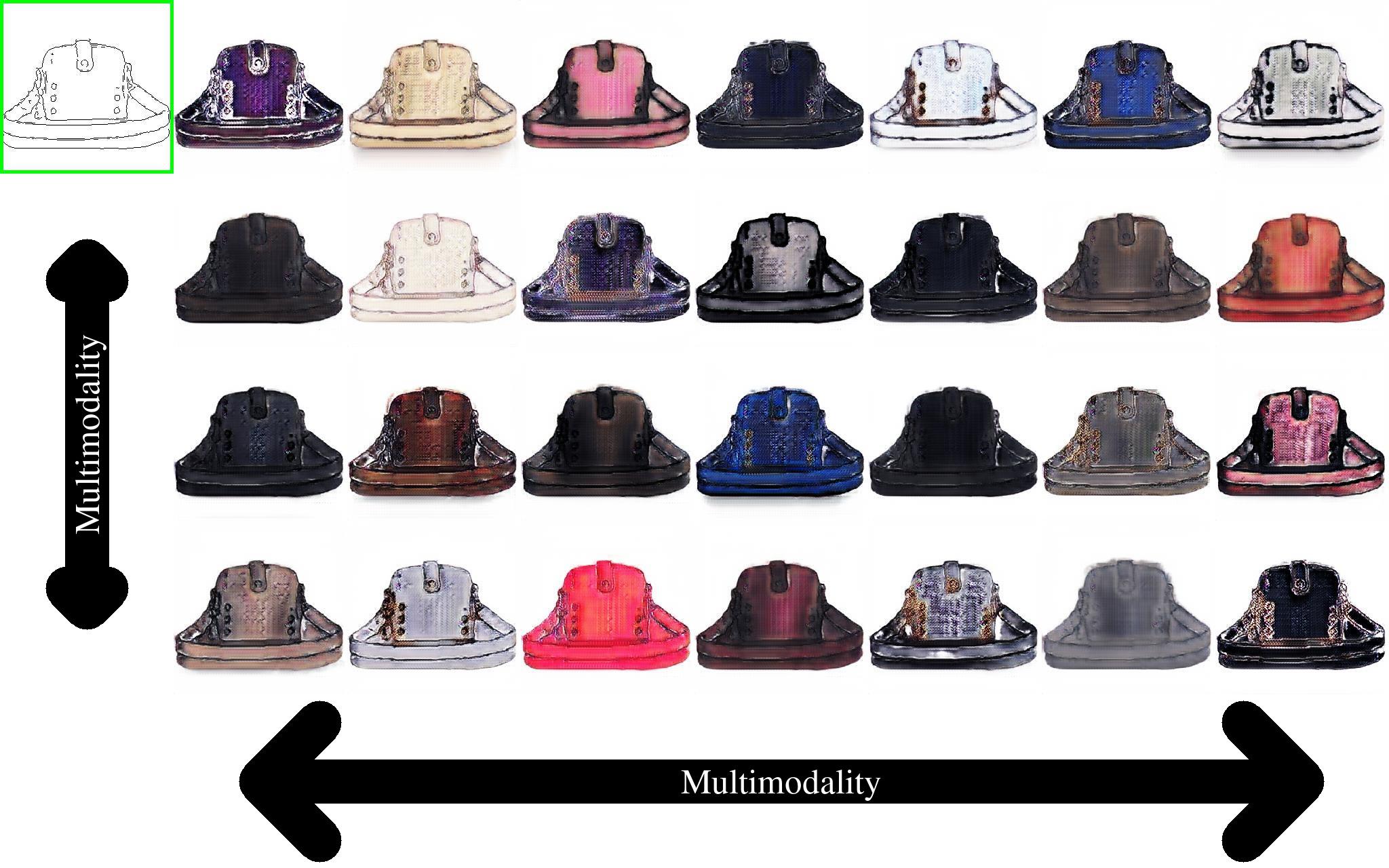}
\end{center}
   \caption{\textbf{SMIT qualitative results for edges2handbags.} Using a sketch from the test set as input (green box), we display diverse outputs across rows and columns.}
\label{fig:handbags1}
\end{figure*}

\begin{figure*}[t!]
\begin{center}
   \includegraphics[width=\linewidth]{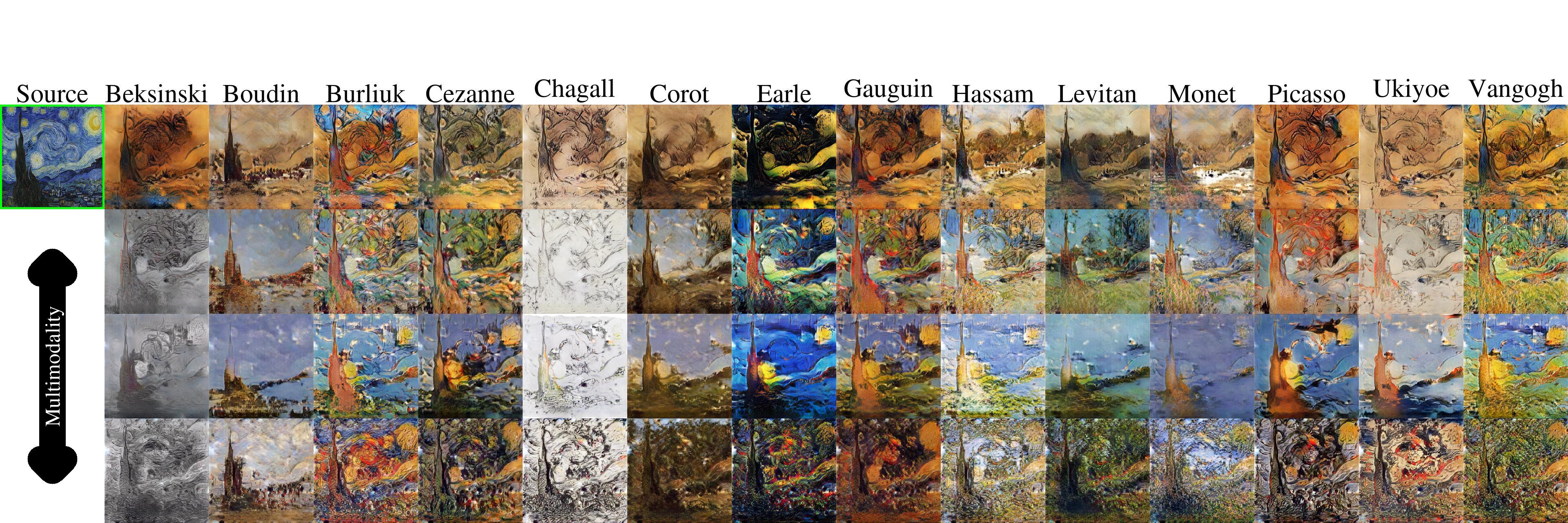}
\end{center}
   \caption{\textbf{SMIT qualitative results for painters.} Using the famous starry night as input (image in the wild), we show the corresponding transformations for every painter (columns) for different modalities (rows), including Van Gogh himself.}
\label{fig:painters0}
\end{figure*}

\begin{figure*}[t!]
\begin{center}
   \includegraphics[width=\linewidth]{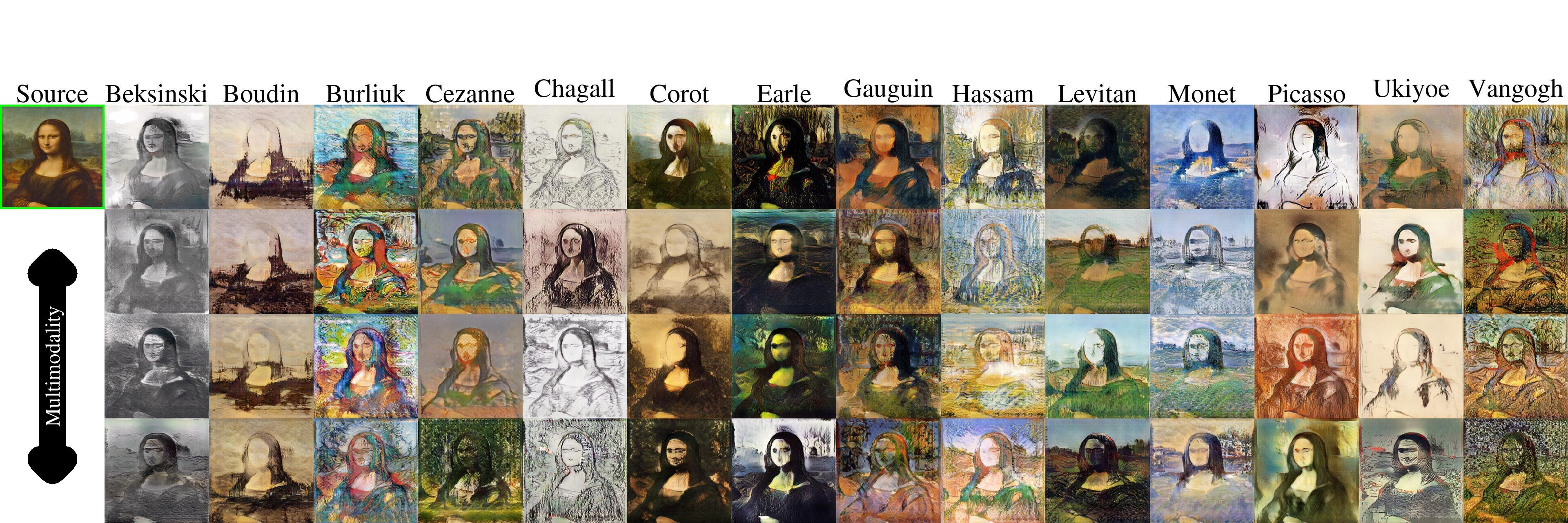}
\end{center}
   \caption{\textbf{SMIT qualitative results for painters.} Using the famous La Joconde as input (image in the wild), we show the corresponding transformations for every painter (columns) for different modalities (rows).}
\label{fig:painters1}
\end{figure*}

\begin{figure*}[t!]
\begin{center}
   \includegraphics[width=\linewidth]{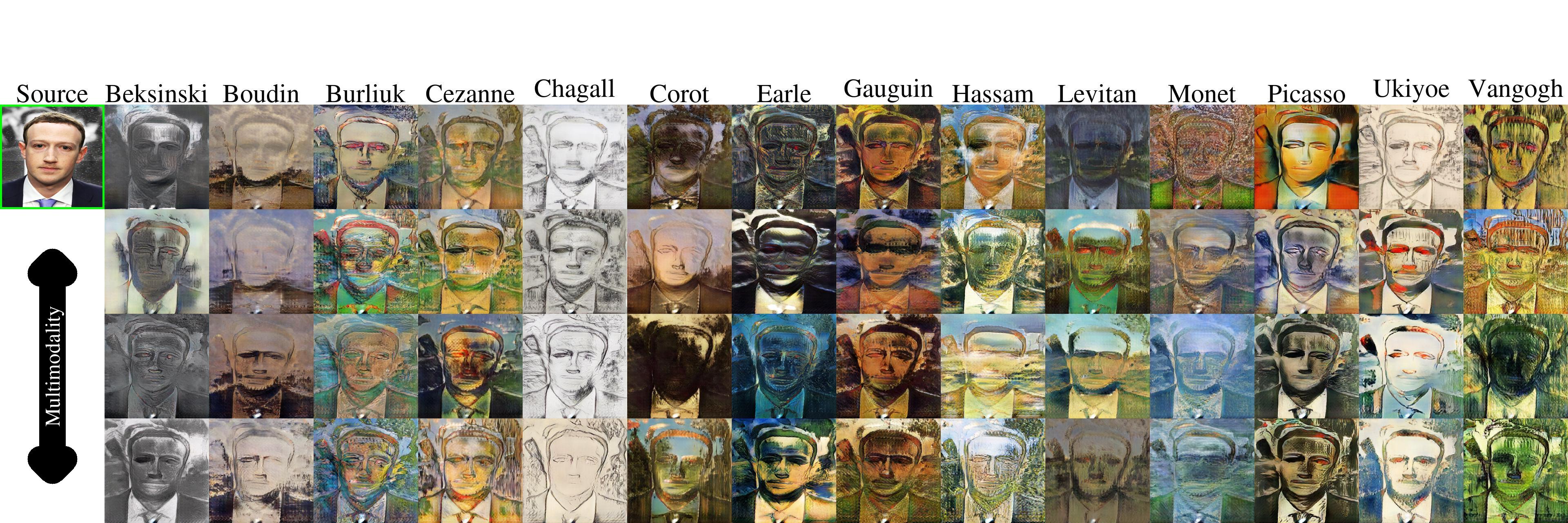}
\end{center}
   \caption{\textbf{SMIT qualitative results for painters.} Using a regular in-the-wild image as input (image in the wild), we show the corresponding transformations for every painter (columns) for different modalities (rows).}
\label{fig:painters2}
\end{figure*}

\begin{figure*}[t!]
\begin{center}
   \includegraphics[width=0.8\linewidth]{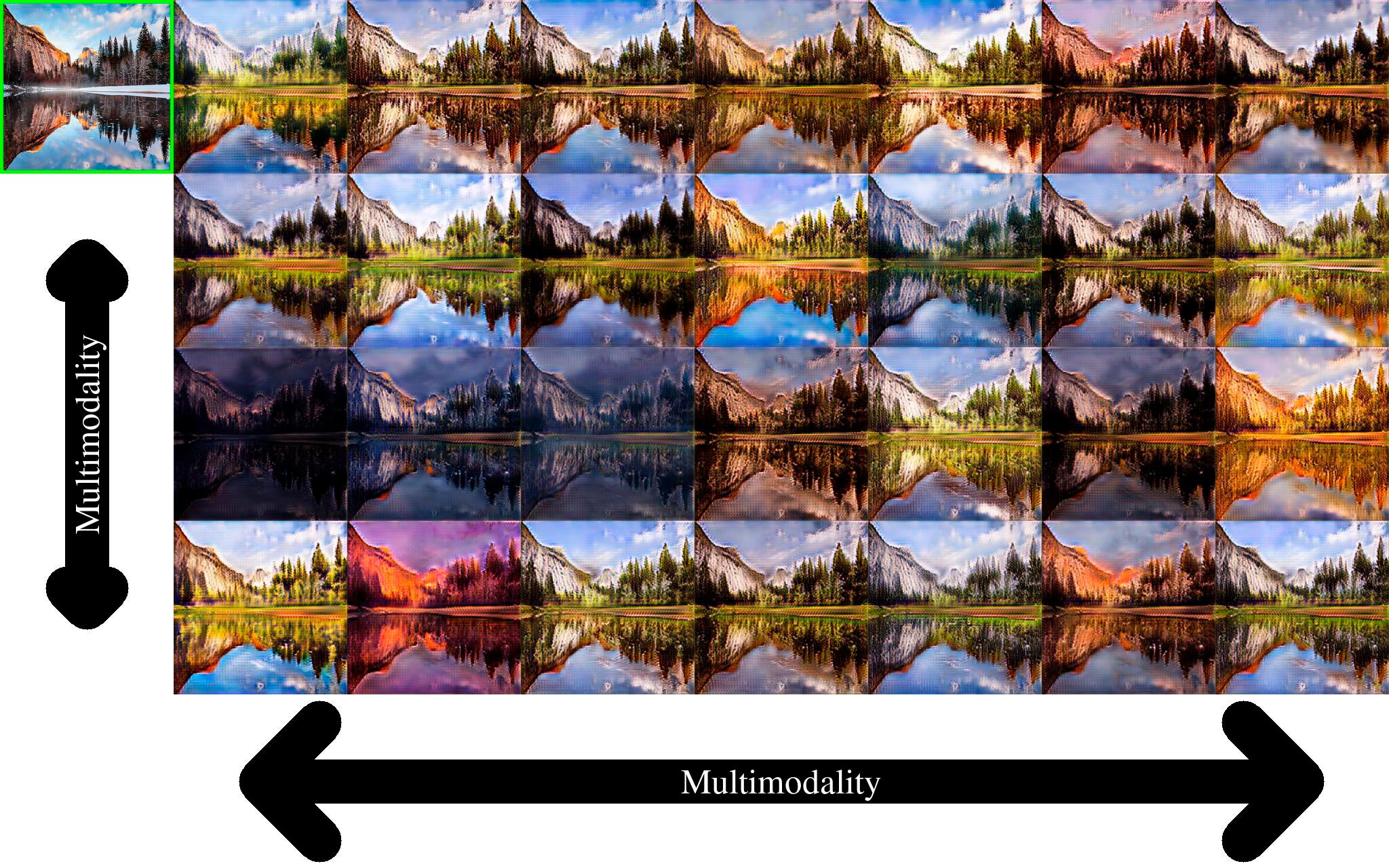}
\end{center}
   \caption{\textbf{SMIT qualitative results for Yosemite.} Using a winter image from the test set as input (green box), we show the corresponding summer transformations across different modalities (rows and columns).}
\label{fig:yosemite0}
\end{figure*}

\begin{figure*}[t!]
\begin{center}
   \includegraphics[width=0.8\linewidth]{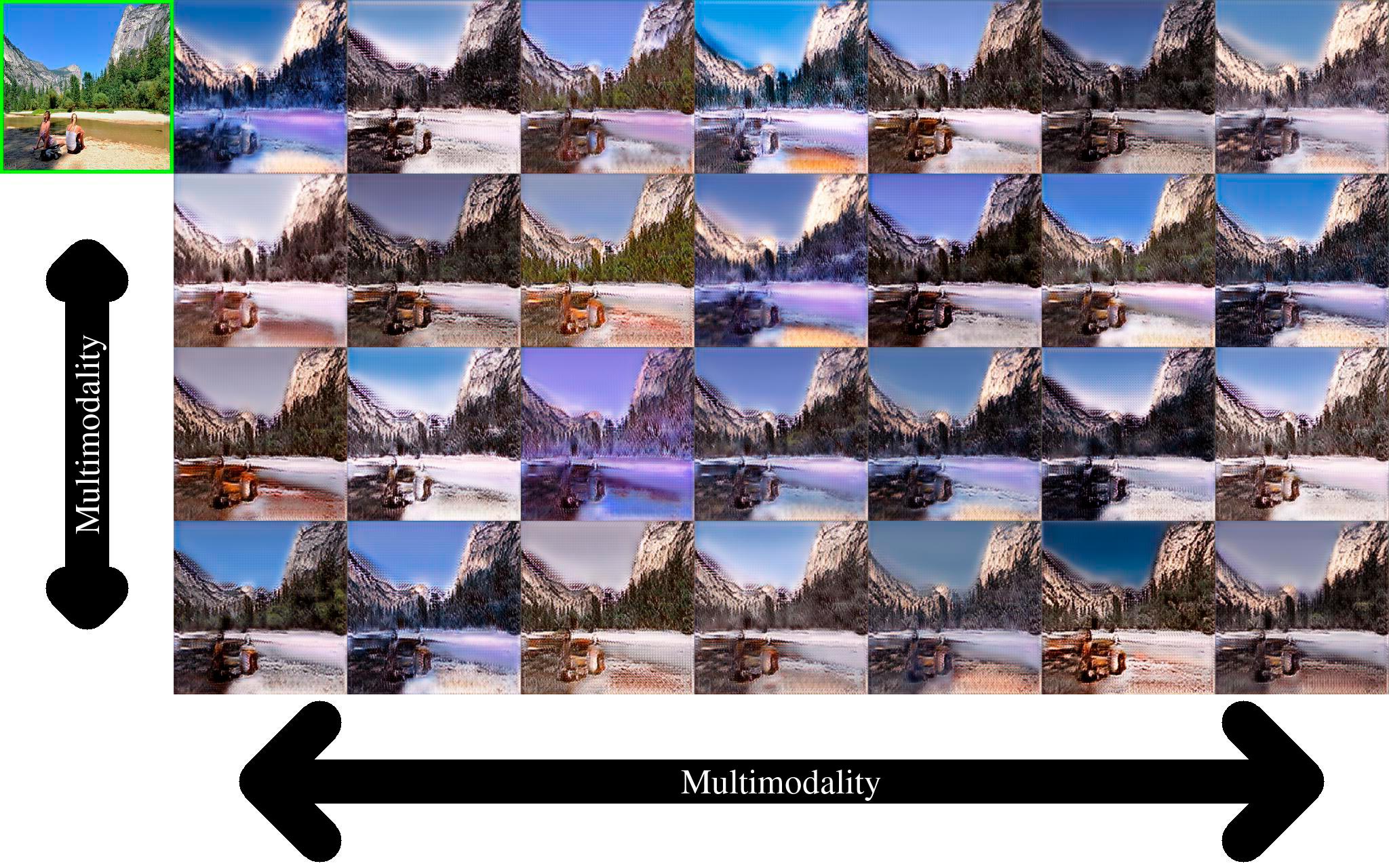}
\end{center}
   \caption{\textbf{SMIT qualitative results for Yosemite.} Using a summer image from the test set as input (green box), we show the corresponding winter transformations across different modalities (rows and columns).}
\label{fig:yosemite1}
\end{figure*}

\begin{figure*}[t!]
\begin{center}
   \includegraphics[width=0.55\linewidth]{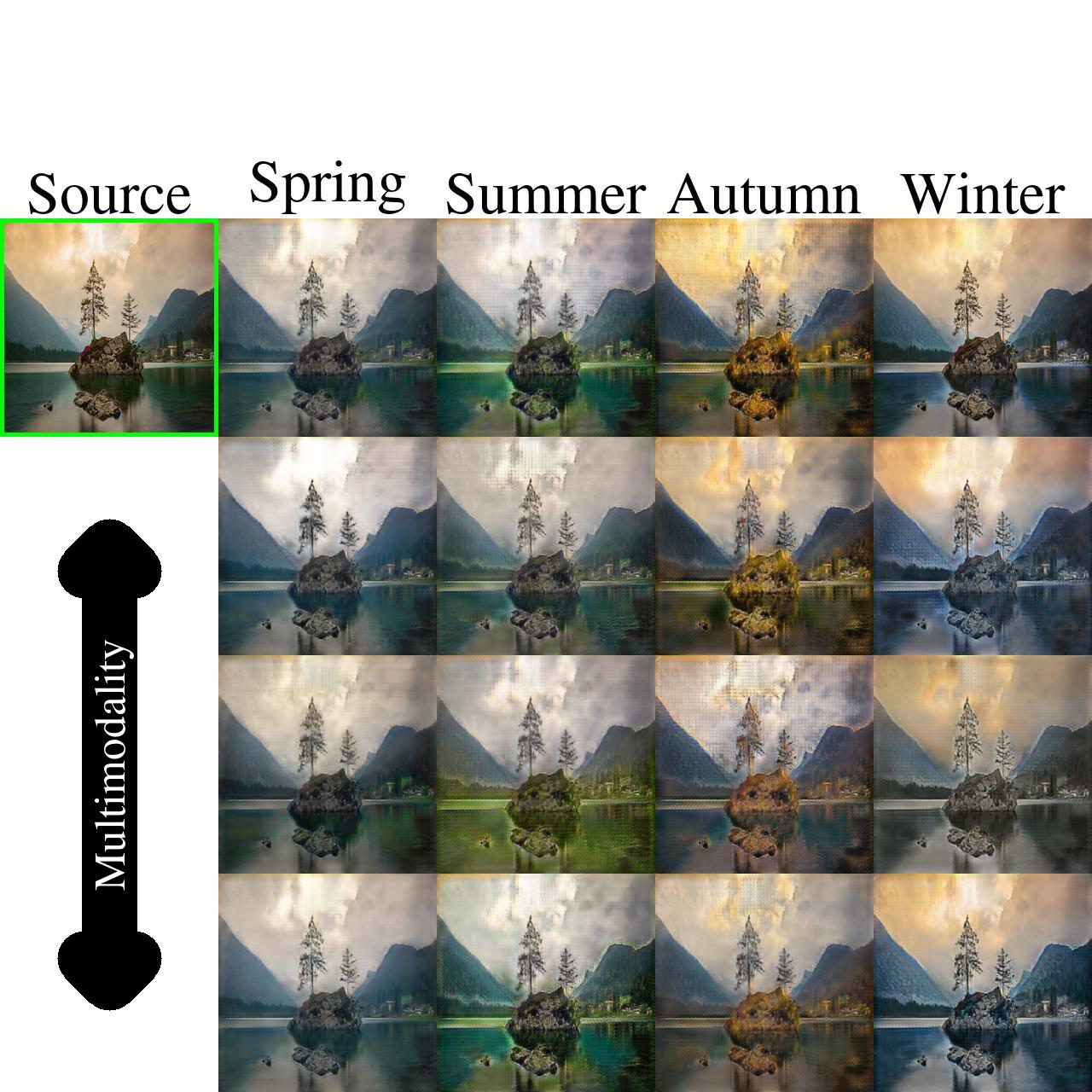}
\end{center}
   \caption{\textbf{SMIT qualitative results for Alps Season.} Using an image in the wild as input (green box), we show the corresponding season transformations (columns) across different modalities (rows).}
\label{fig:alps0}
\end{figure*}

\begin{figure*}[t!]
\begin{center}
   \includegraphics[width=0.55\linewidth]{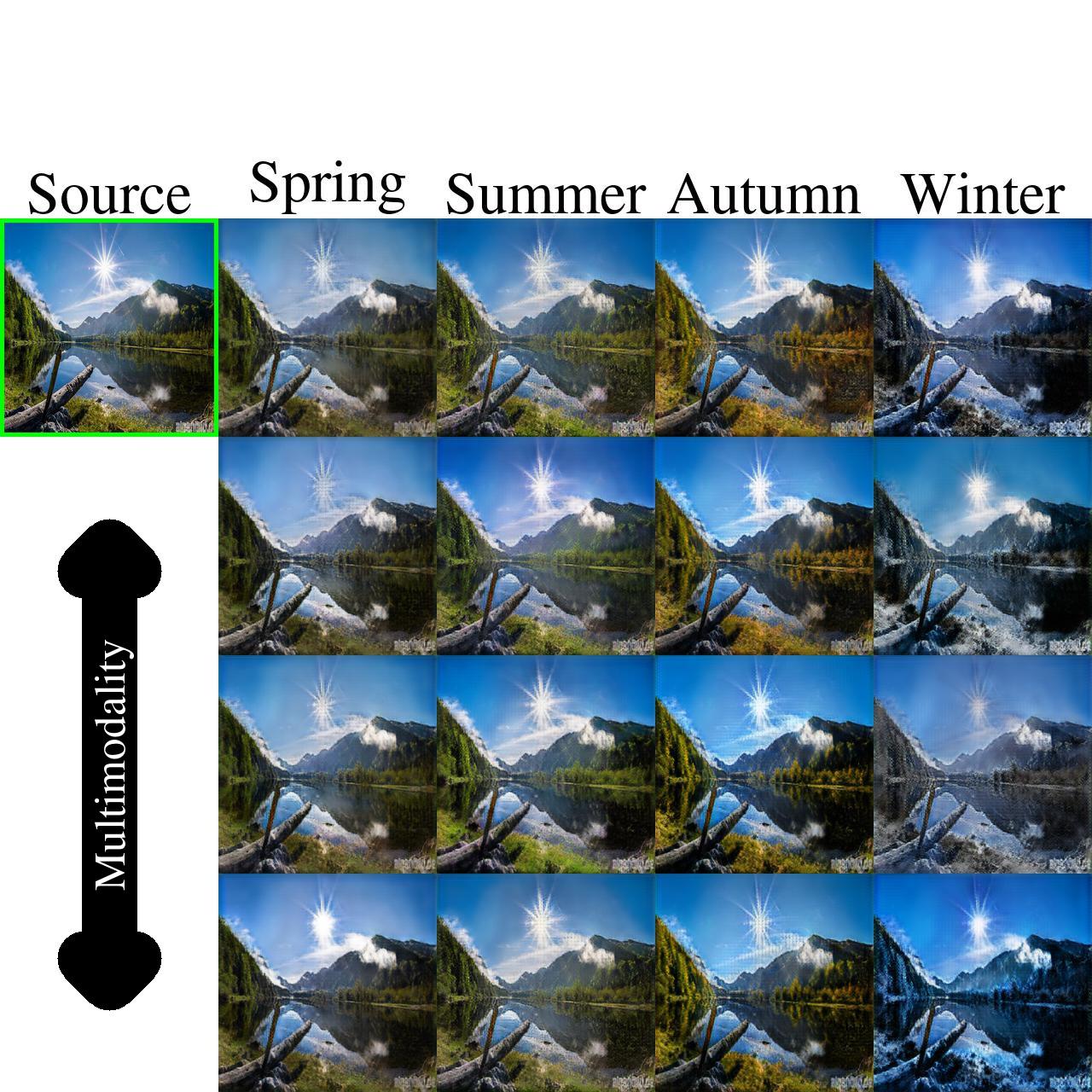}
\end{center}
   \caption{\textbf{SMIT qualitative results for Alps Season.} Using an image from the test set in the wild as input (green box), we show the corresponding season transformations (columns) across different modalities (rows).}
\label{fig:alps1}
\end{figure*}

\begin{figure*}[t!]
\begin{center}
   \includegraphics[width=0.55\linewidth]{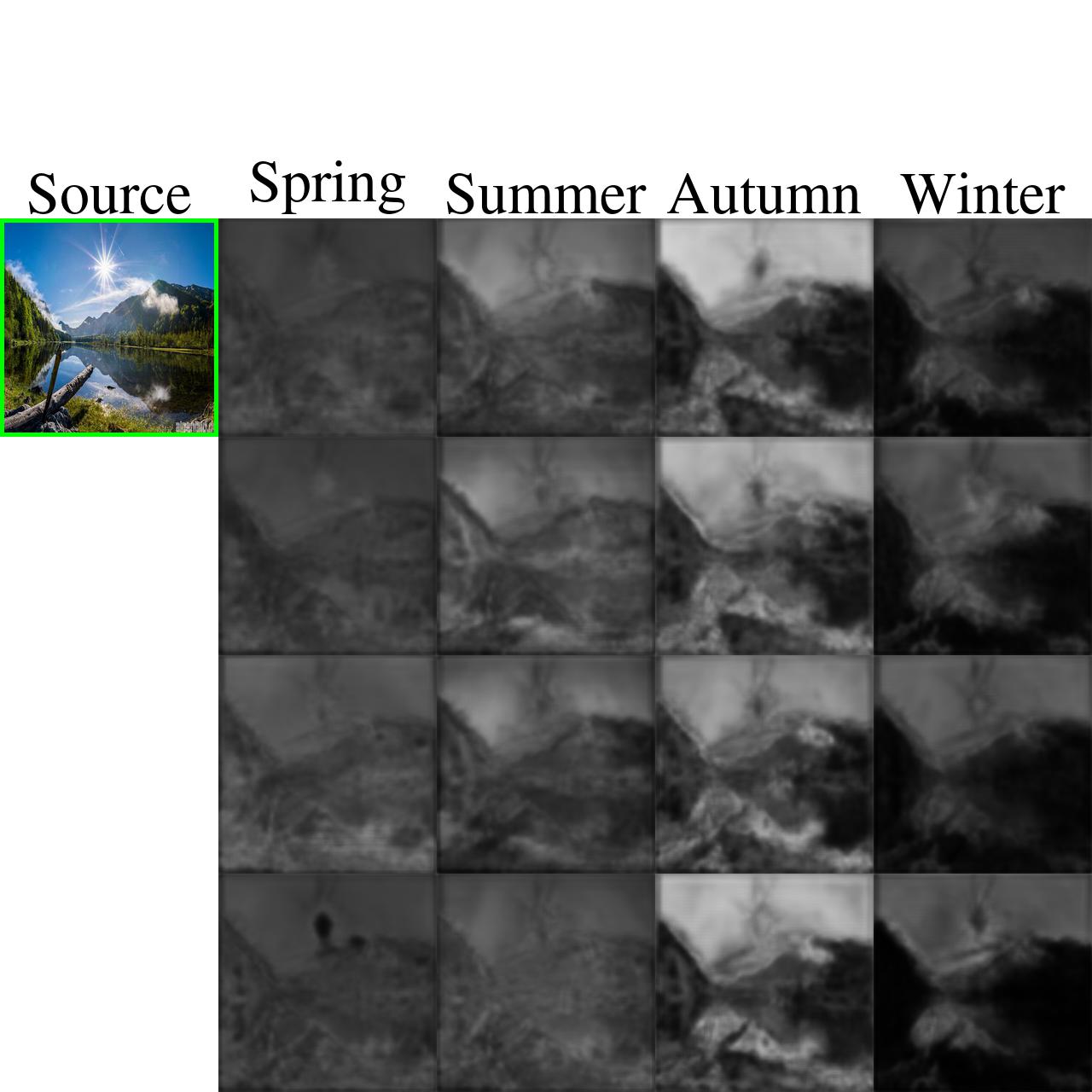}
\end{center}
   \caption{\textbf{SMIT qualitative attention results for Alps Season.} We show the attention masks that produces \fref{fig:alps1}. Black regions represent the changes with respect to the input.}
\label{fig:alps1_attn}
\end{figure*}

\begin{figure*}[t!]
\begin{center}
   \includegraphics[width=\linewidth]{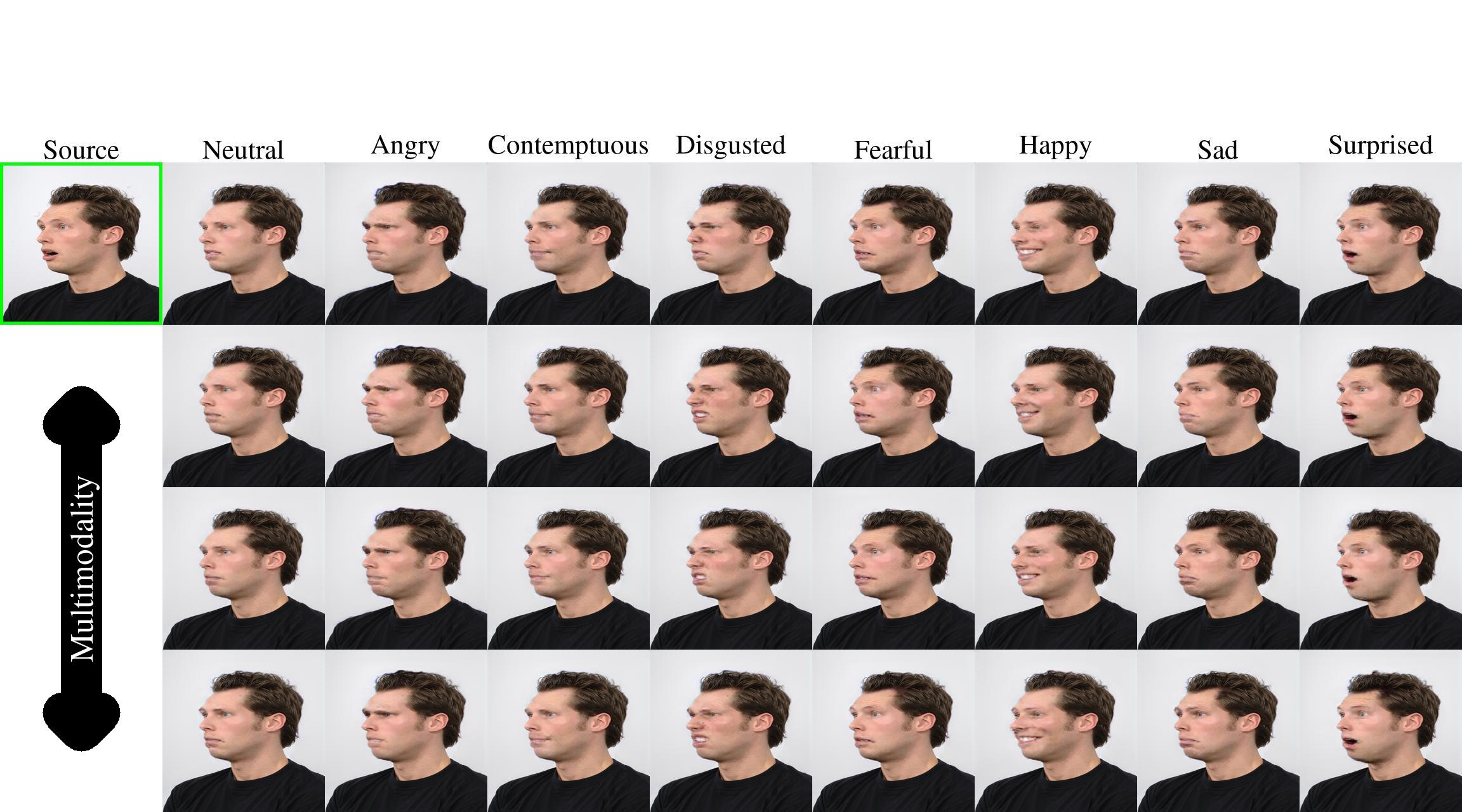}
\end{center}
   \caption{\textbf{SMIT qualitative results for RafD (emotions).} Using a surprised person from the test set as input (green box), we show the corresponding emotion generation (columns) for different modalities (rows).}
\label{fig:rafd0}
\end{figure*}

\begin{figure*}[t!]
\begin{center}
   \includegraphics[width=\linewidth]{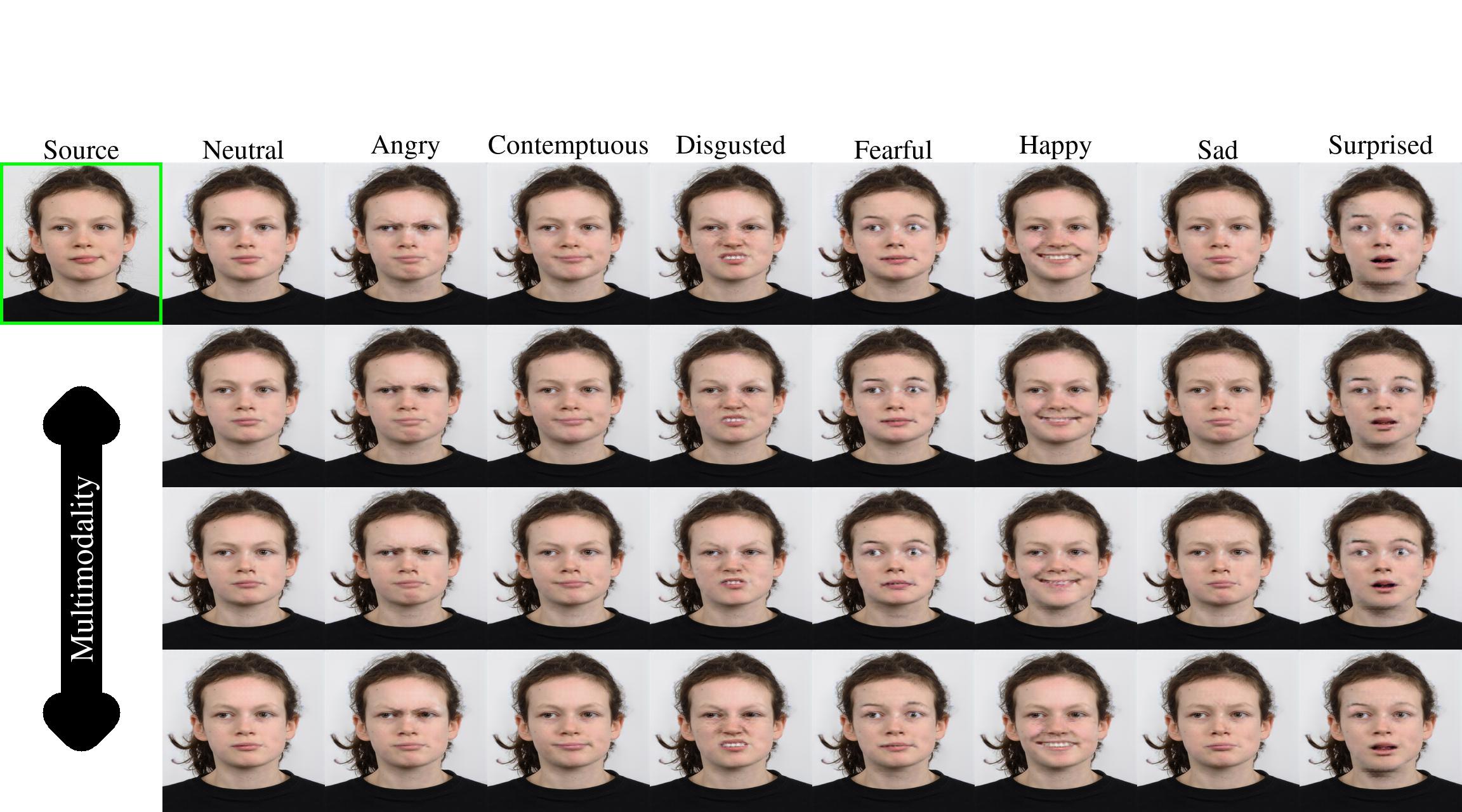}
\end{center}
   \caption{\textbf{SMIT qualitative results for RafD (emotions).} Using a contemptuous person from the test set as input (green box), we show the corresponding emotion generation (columns) for different modalities (rows).}
\label{fig:rafd1}
\end{figure*}

\begin{figure*}[t!]
\begin{center}
   \includegraphics[width=\linewidth]{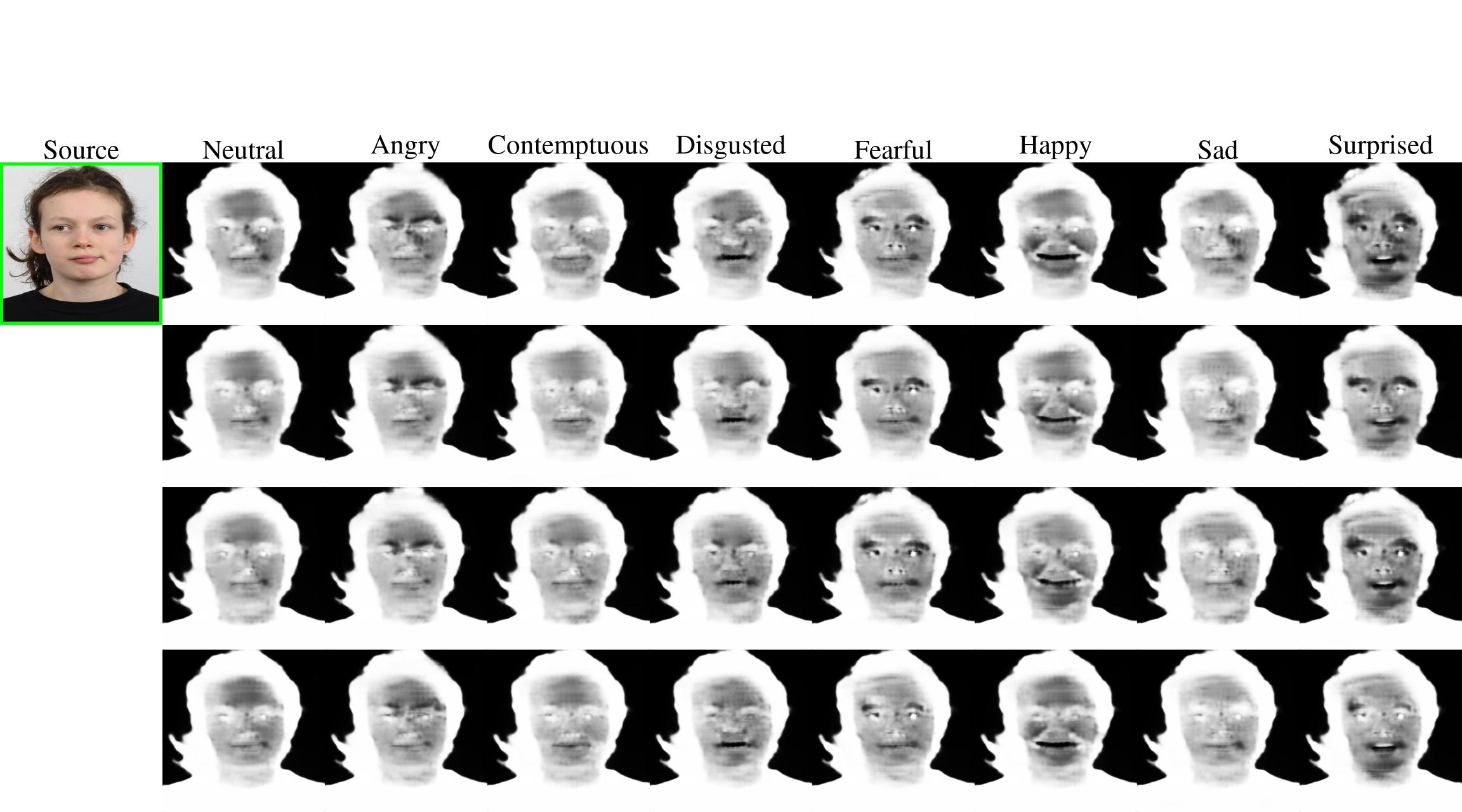}
\end{center}
   \caption{\textbf{SMIT qualitative attention results for RafD.} We show the attention masks that produces \fref{fig:rafd1}. Black regions represent the changes with respect to the input.}
\label{fig:rafd1_attn}
\end{figure*}

\begin{figure*}[t!]
\begin{center}
   \includegraphics[width=\linewidth]{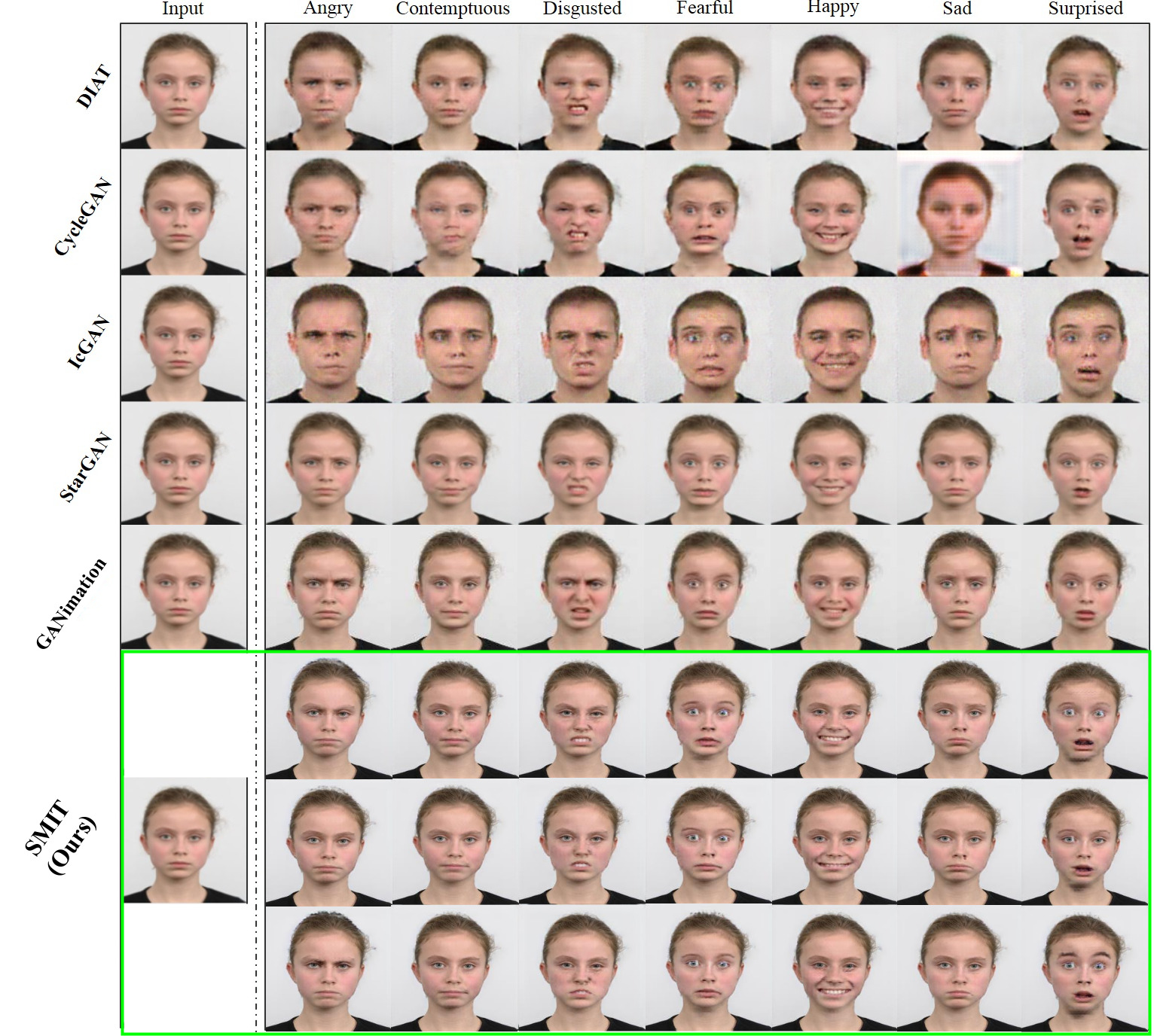}
\end{center}
   \caption{\textbf{Qualitative comparison against state-of-the-art works.} We depict differences in the qualitity of our multimodal and multi-label method (rows in the green box) against deterministic approaches (other rows). We extract the upper side of the image from GANimation paper.}
\label{fig:rafd_comparison}
\vspace{3cm}
\end{figure*}

\begin{figure*}[t!]
\begin{center}
   \includegraphics[width=\linewidth]{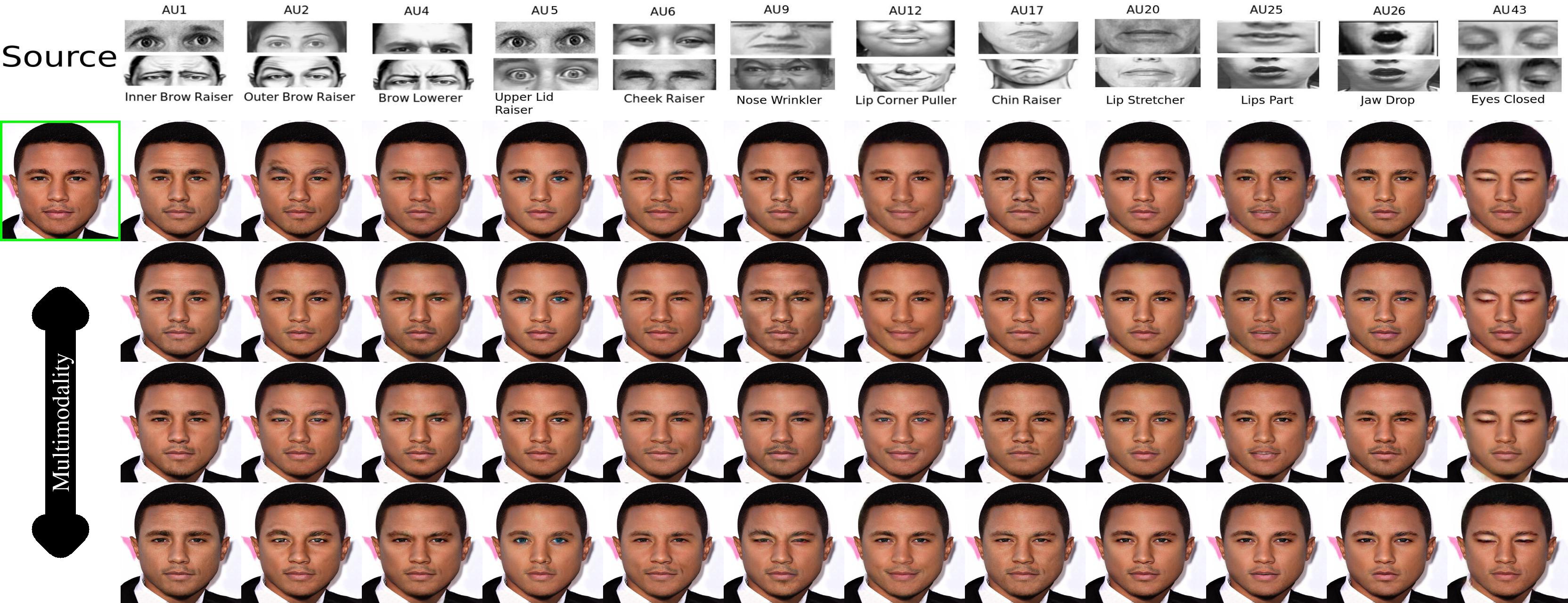}
\end{center}
   \caption{\textbf{SMIT qualitative results for EmotionNet (facial expressions).} Using an image in the wild as input (green box), we show the corresponding facial expression (columns) swapping (with respect to the input) for different modalities (rows).}
\label{fig:emotionnet0}
\vspace{3cm}
\end{figure*}

\begin{figure*}[t!]
\begin{center}
   \includegraphics[width=\linewidth]{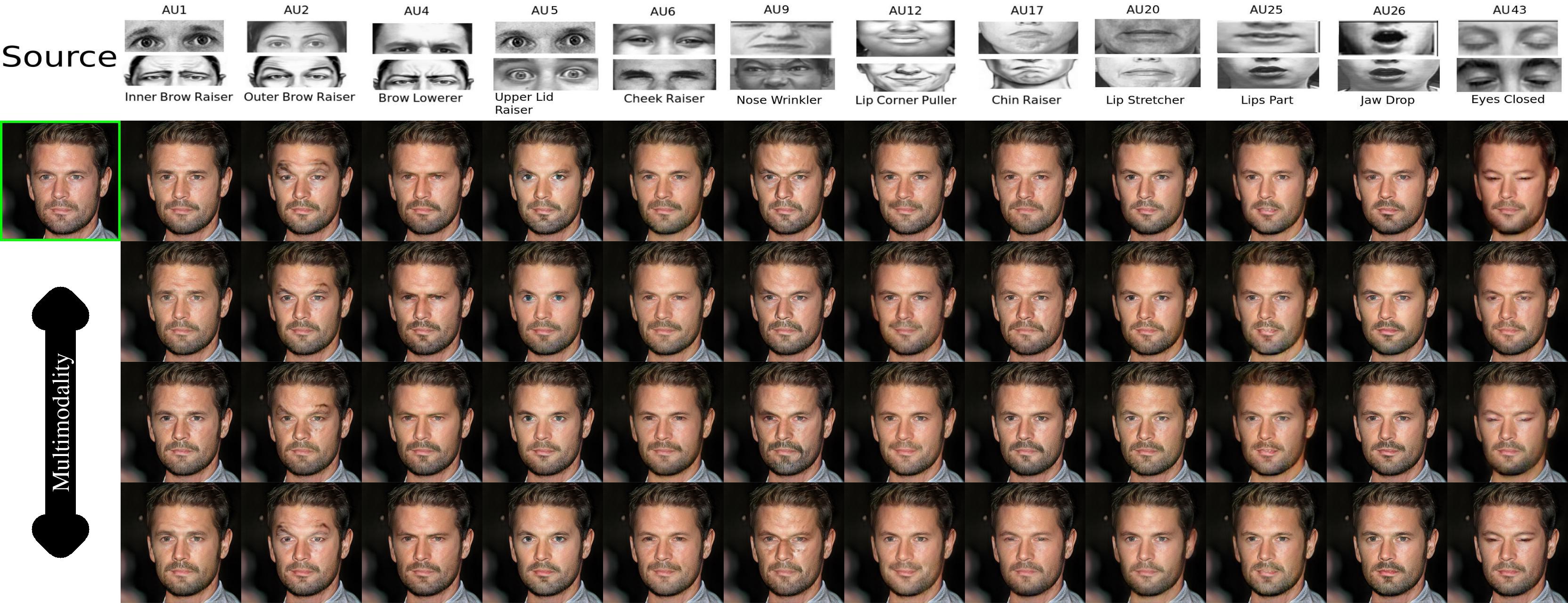}
\end{center}
   \caption{\textbf{SMIT qualitative results for EmotionNet (facial expressions).} Using an image in the wild as input (green box), we show the corresponding facial expression (columns) swapping (with respect to the input) for different modalities (rows).}
\label{fig:emotionnet1}
\end{figure*}

\begin{figure*}[t!]
\begin{center}
   \includegraphics[width=\linewidth]{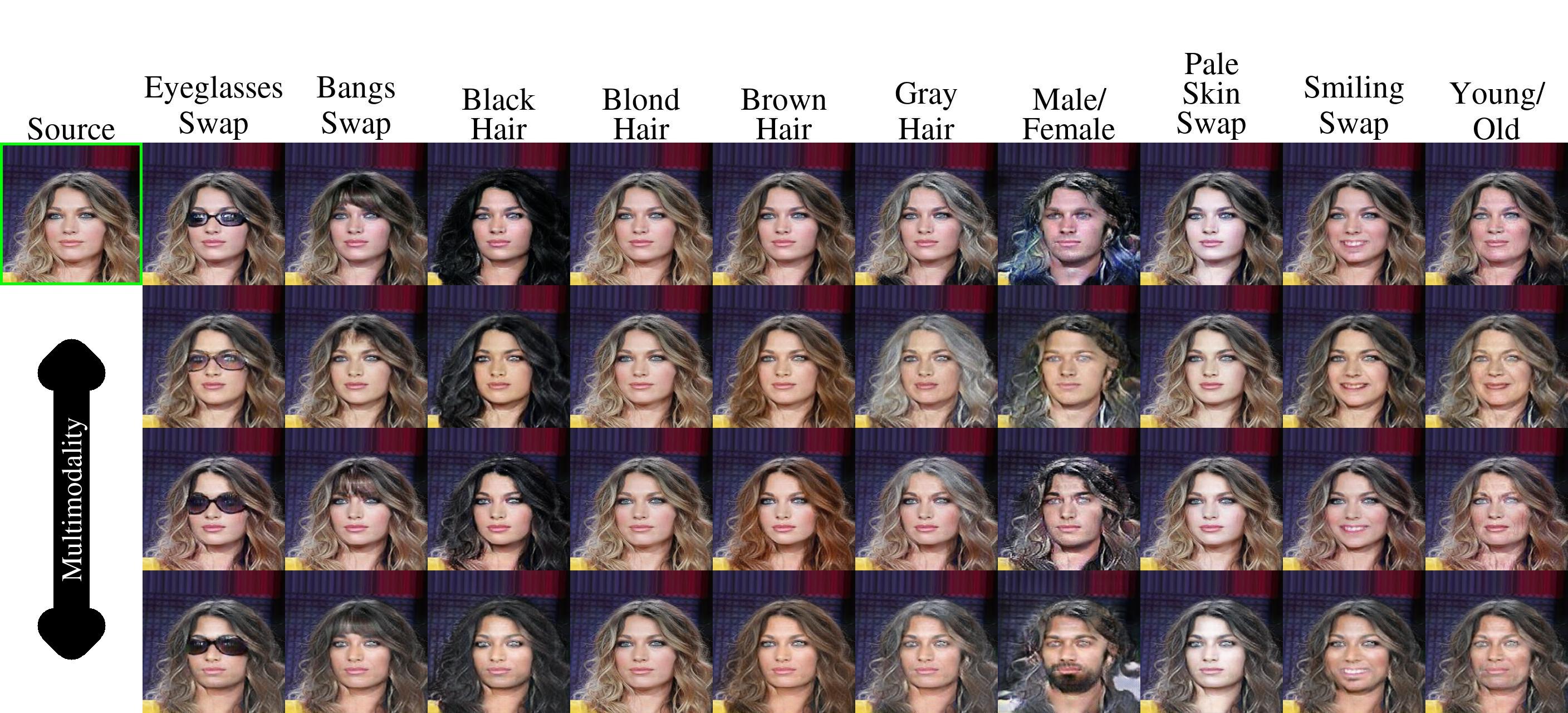}
\end{center}
   \caption{\textbf{SMIT qualitative results for CelebA with 10 attributes.} Using an image from the test set as input (green box), we show the corresponding attributes (columns) swapping (with respect to the input) for different modalities (rows).}
\label{fig:celeba101}
\end{figure*}

\begin{figure*}[t!]
\begin{center}
   \includegraphics[width=\linewidth]{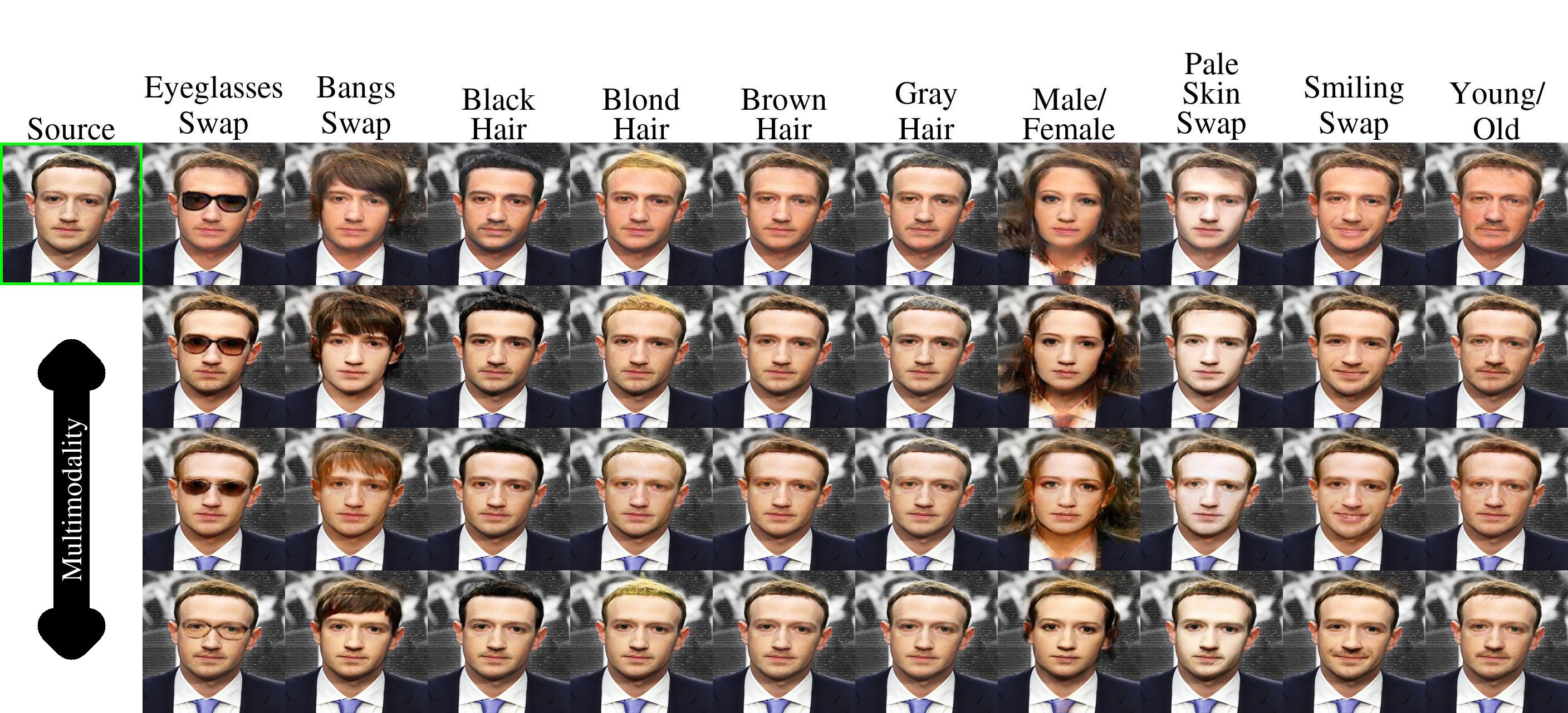}
\end{center}
   \caption{\textbf{SMIT qualitative results for CelebA with 10 attributes.} Using an image in the wild as input (green box), we show the corresponding attributes (columns) swapping (with respect to the input) for different modalities (rows).}
\label{fig:celeba100}
\end{figure*}

\begin{figure*}[t!]
\begin{center}
   \includegraphics[width=\linewidth]{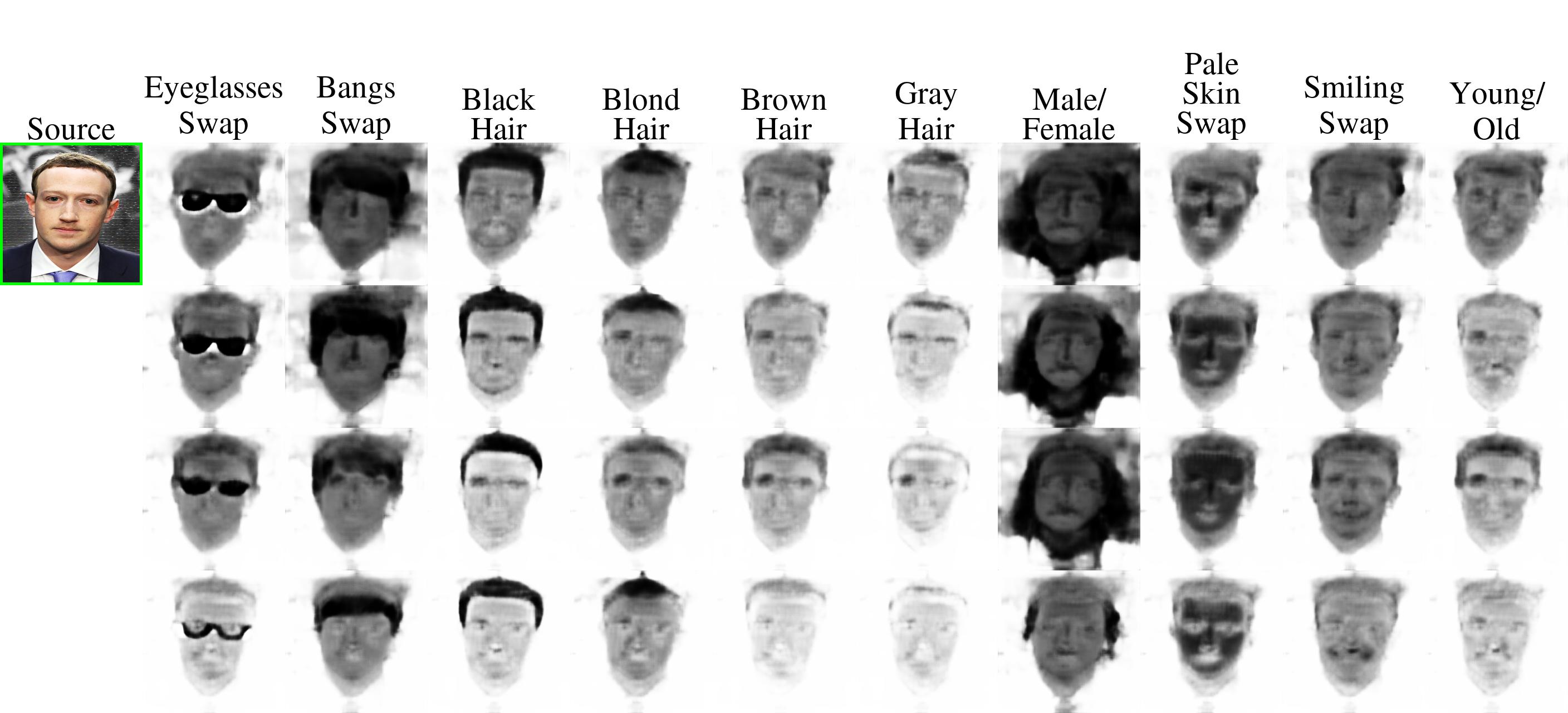}
\end{center}
   \caption{\textbf{SMIT qualitative attention results for CelebA with 10 attributes.} These images are the attention maps that produces \fref{fig:celeba100}. Black regions represent the changes with respect to the input.}
\label{fig:celeba100_attn}
\end{figure*}

\begin{figure*}[t!]
\begin{center}
   \includegraphics[width=0.85\linewidth]{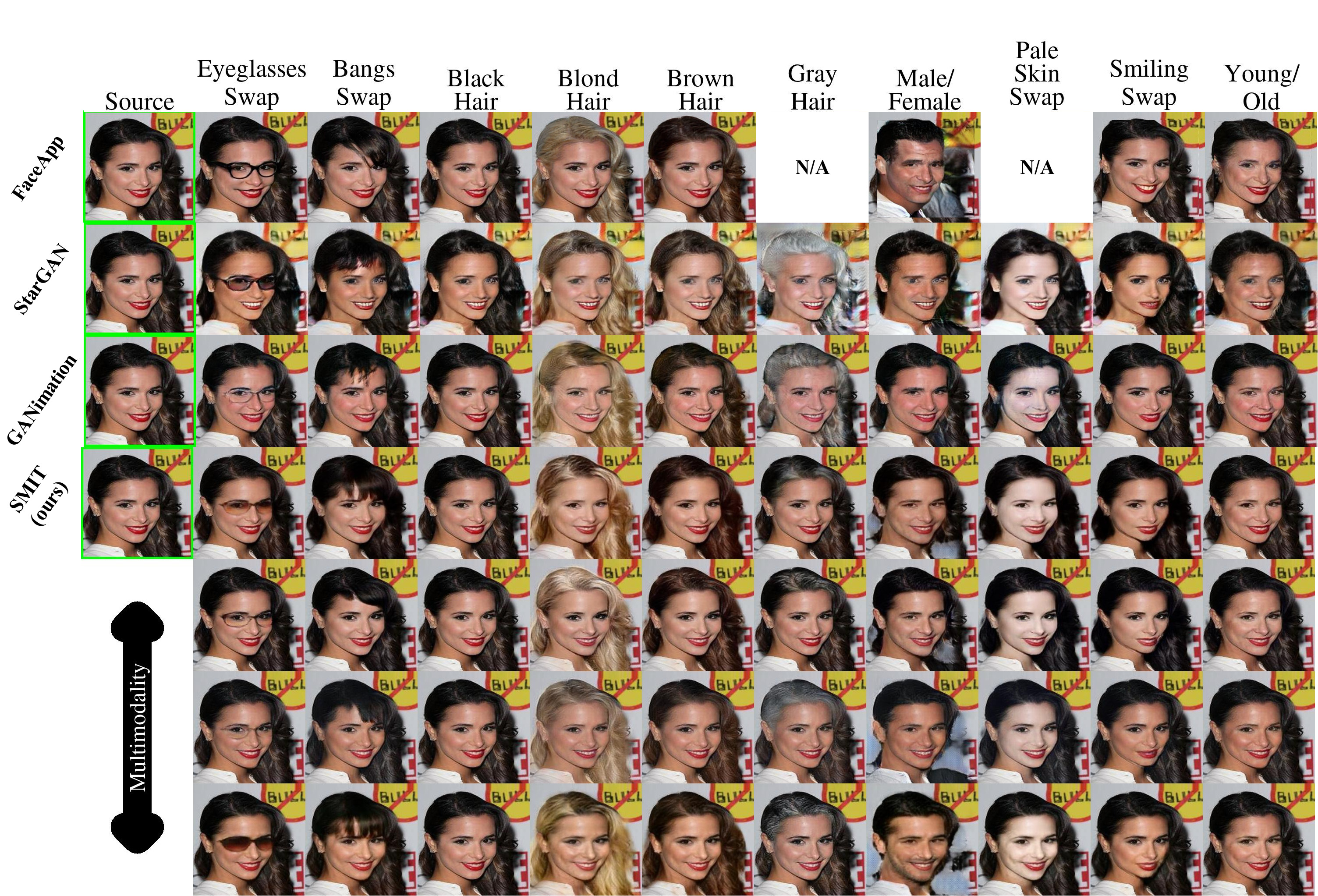}
\end{center}
   \caption{\textbf{SMIT qualitative results for CelebA with 10 attributes.} Using an image from the test set as input, we show qualitative comparison against state-of-the-art methods in multi-label image-to-image translation. We retrain StarGAN and GANimation for these results.}
\label{fig:celeba_comparison}
\end{figure*}

\begin{figure*}[t!]
\begin{center}
   \includegraphics[width=\linewidth]{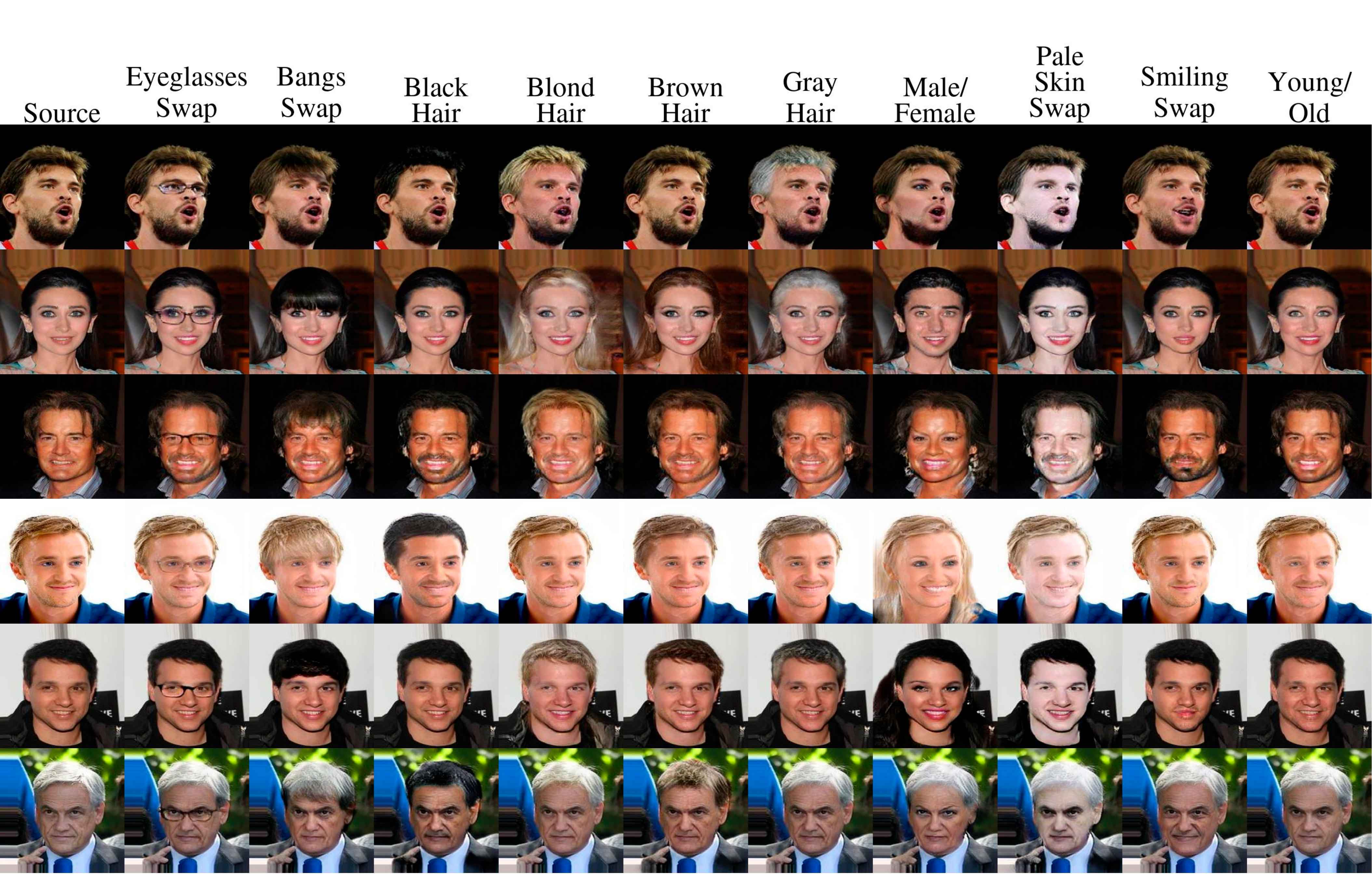}
\end{center}
   \caption{\textbf{SMIT qualitative results for CelebA with 10 attributes.} Using an image in the wild as input (green box) and a fixed modality, we show a strong similarity within the attributes (columns) for different people (rows).}
\label{fig:celeba_samestyle}
\end{figure*}

\begin{figure*}[t!]
\begin{center}
   \includegraphics[trim={0 0 244cm 0},clip,width=\linewidth]{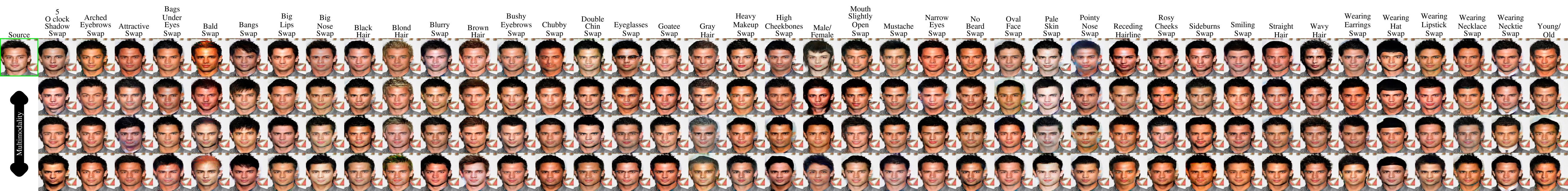}
   \includegraphics[trim={122cm 0 122cm 0},clip,width=\linewidth]{supplemental_figures/celeba400.jpg}
   \includegraphics[trim={244cm 0 0 0},clip,width=\linewidth]{supplemental_figures/celeba400.jpg}   
\end{center}
   \caption{\textbf{SMIT qualitative results for CelebA with full attributes (40).} Using an image from the test set as input (green box), we show the corresponding attributes (columns) swapping (with respect to the input) for different modalities (rows). In this framework, the identity loss plays a critical role by means of constraining the identity of the person at every modality.}
\label{fig:celeba400}
\end{figure*}

\begin{figure*}[t!]
\begin{center}
   \includegraphics[trim={0 0 244cm 0},clip,width=\linewidth]{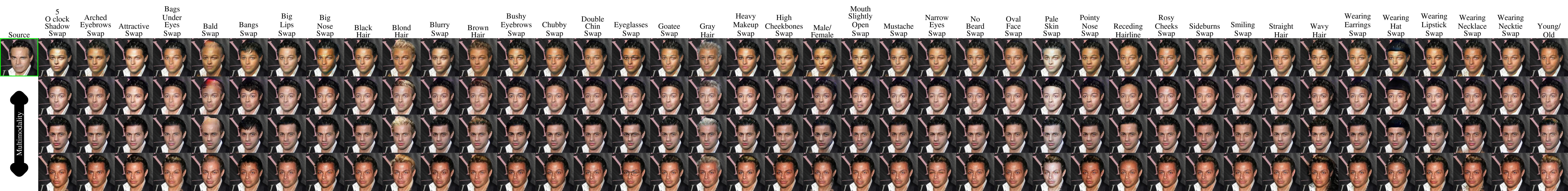}
   \includegraphics[trim={122cm 0 122cm 0},clip,width=\linewidth]{supplemental_figures/celeba401.jpg}
   \includegraphics[trim={244cm 0 0 0},clip,width=\linewidth]{supplemental_figures/celeba401.jpg}   
\end{center}
   \caption{\textbf{SMIT qualitative results for CelebA with full attributes (40).} Using an image from the test set as input (green box), we show the corresponding attributes (columns) swapping (with respect to the input) for different modalities (rows).}
\label{fig:celeba401}
\end{figure*}

We qualitatively show SMIT multimodal representations for different random styles over edges2shoes (\fref{fig:shoes0} and \ref{fig:shoes1}), edges2handbags (\fref{fig:handbags0} and \ref{fig:handbags1}), Yosemite (\fref{fig:yosemite0} and \ref{fig:yosemite1}), Alps seasons (\fref{fig:alps0}, \ref{fig:alps1}, and \ref{fig:alps1_attn}), RafD (\fref{fig:rafd0}, \ref{fig:rafd1}, \ref{fig:rafd1_attn}, and \ref{fig:rafd_comparison}), painters (\fref{fig:painters0}, \ref{fig:painters1} and \ref{fig:painters2}), EmotionNet (\fref{fig:emotionnet0} and \ref{fig:emotionnet1}), 11-CelebA (\fref{fig:celeba101}, \ref{fig:celeba100}, \ref{fig:celeba100_attn}, \ref{fig:celeba_comparison} and \ref{fig:celeba_samestyle}), and 40-CelebA (\fref{fig:celeba400} and \ref{fig:celeba401}). 

\subsection{Multimodal Interpolation}
\begin{figure*}[t!]
\begin{center}
  \includegraphics[width=0.8\linewidth]{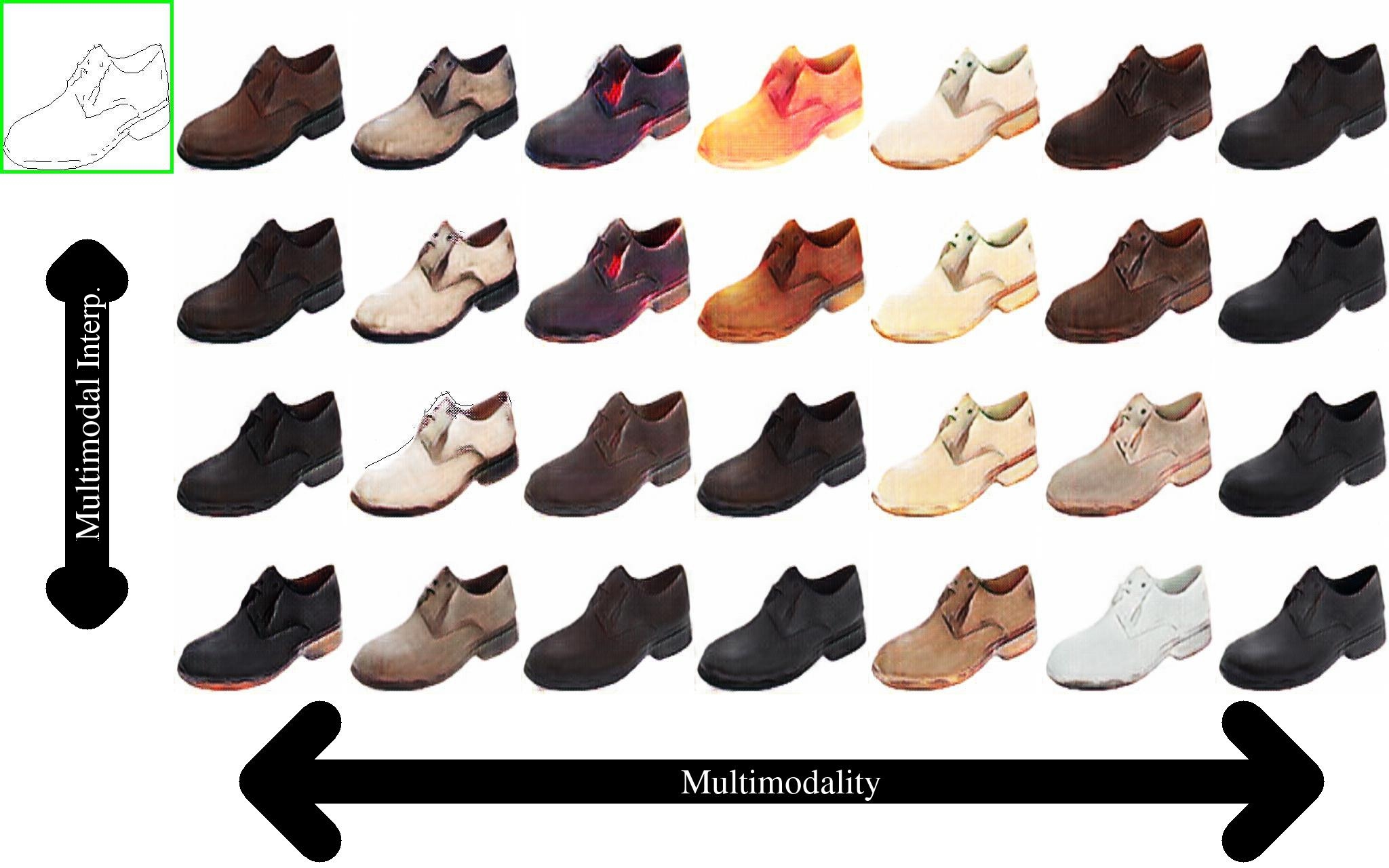}
\end{center}
   \caption{\textbf{SMIT multimodal interpolation results for edges2shoes.} Using an image from the test set as input (green box), we show the style interpolation between the first and last row for different modalities (columns).}
\label{fig:shoes_style_interp}
\end{figure*}

\begin{figure*}[t!]
\begin{center}
  \includegraphics[width=0.8\linewidth]{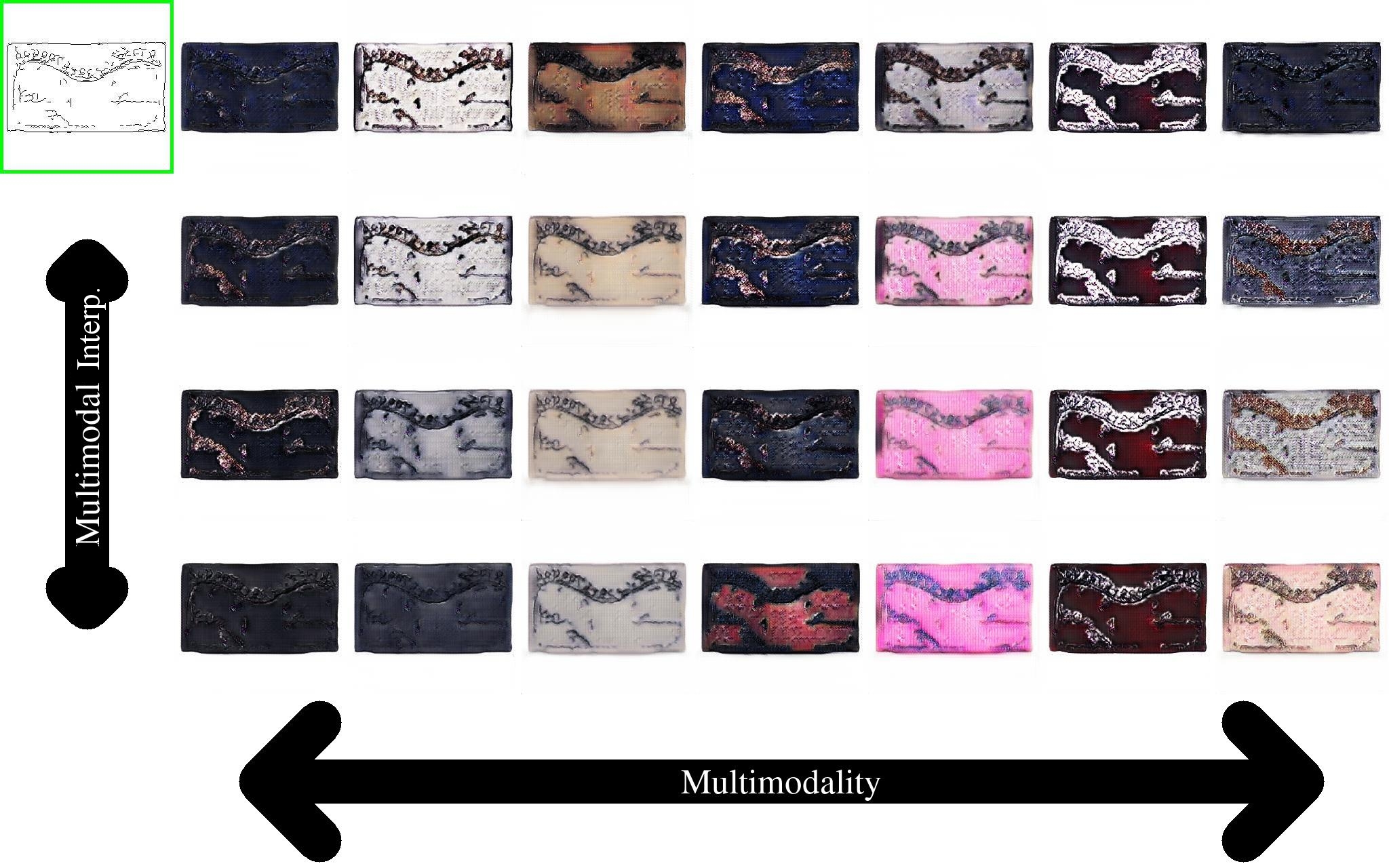}
\end{center}
   \caption{\textbf{SMIT multimodal interpolation results for edges2handbags.} Using an image from the test set as input (green box), we show the style interpolation between the first and last row for different modalities (columns).}
\label{fig:handbags_style_interp}
\end{figure*}

\begin{figure*}[t!]
\begin{center}
  \includegraphics[width=0.8\linewidth]{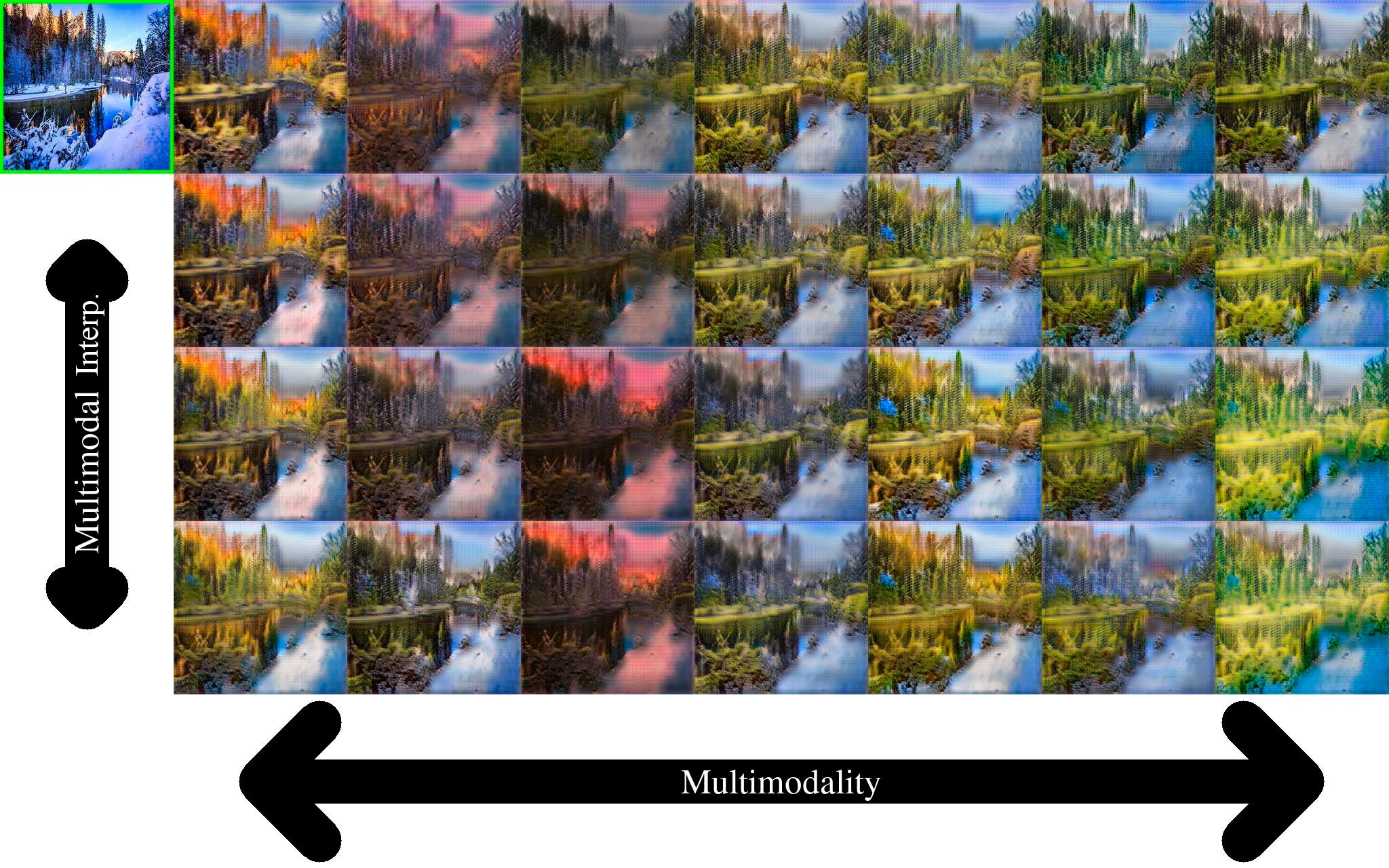}
\end{center}
   \caption{\textbf{SMIT multimodal interpolation results for Yosemite.} Using a winter image from the test set as input (green box), we show the summer style interpolation between the first and last row for different modalities (columns).}
\label{fig:yosemite_style_interp}
\end{figure*}

\begin{figure*}[t!]
\begin{center}
  \includegraphics[width=0.6\linewidth]{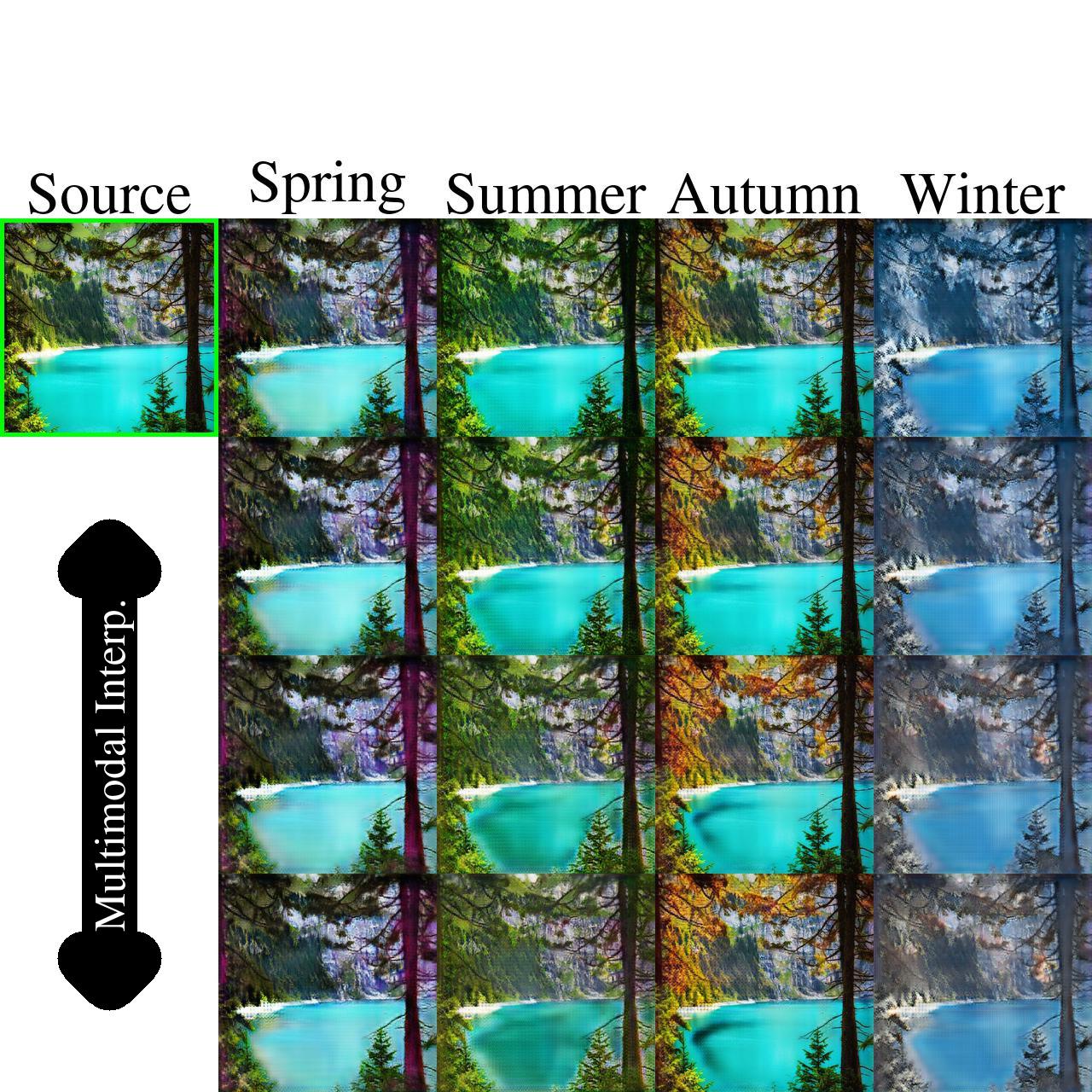}
\end{center}
   \caption{\textbf{SMIT multimodal interpolation results for Alps Season.} Using an image from the test set as input (green box), we show the style interpolation between the first and last row for different season domains (columns).}
\label{fig:alps_style_interp}
\end{figure*}

\begin{figure*}[t!]
\begin{center}
  \includegraphics[trim={0 0 0 3.8cm},clip,width=\linewidth]{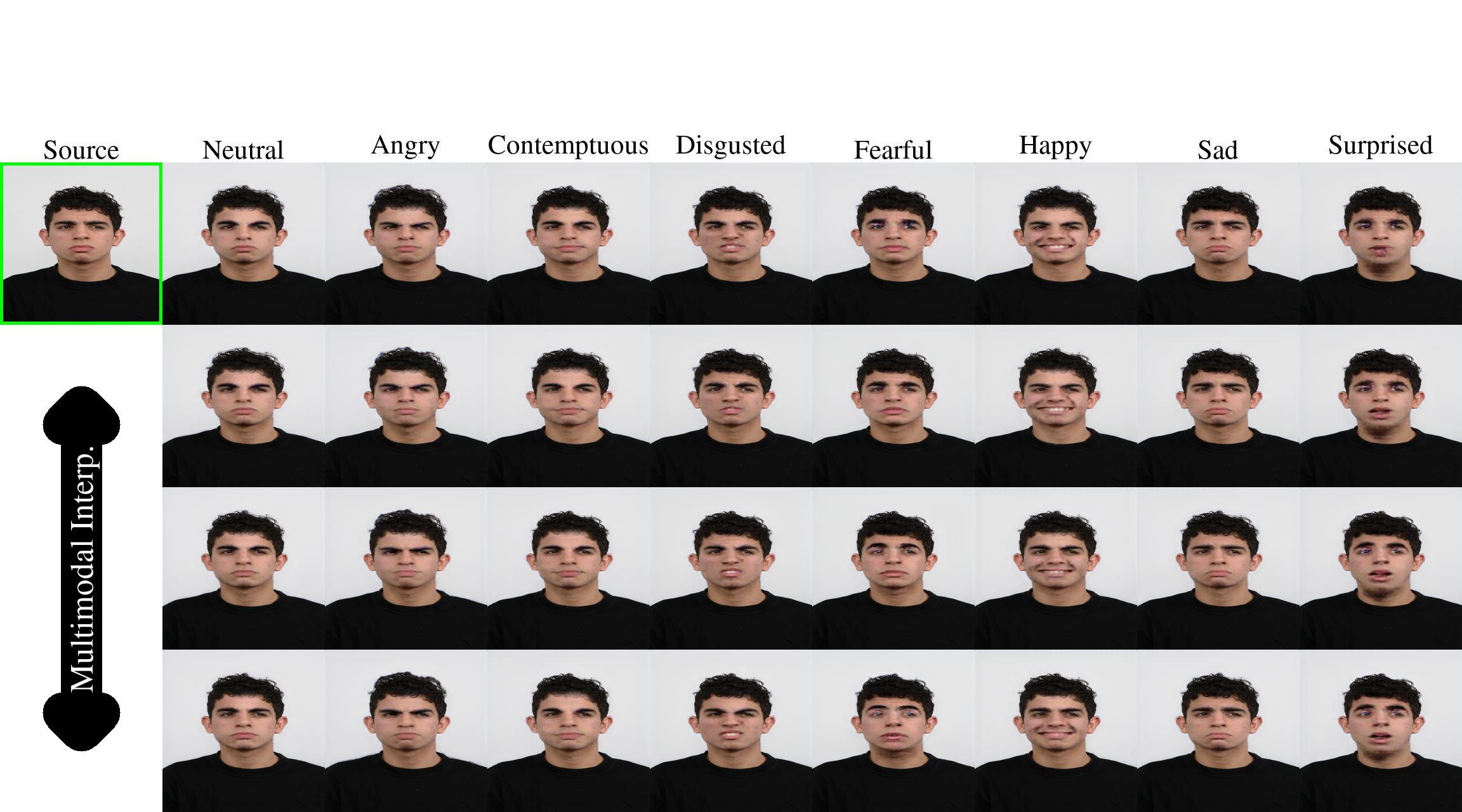}
\end{center}
   \caption{\textbf{SMIT multimodal interpolation results for RafD.} Using an image from the test set as input (green box), we show the emotion style interpolation between the first and last row for different domains (columns).}
\label{fig:rafd_style_interp}
\end{figure*}

\begin{figure*}[t!]
\begin{center}
  \includegraphics[trim={0 0 0 3.8cm},clip,width=\linewidth]{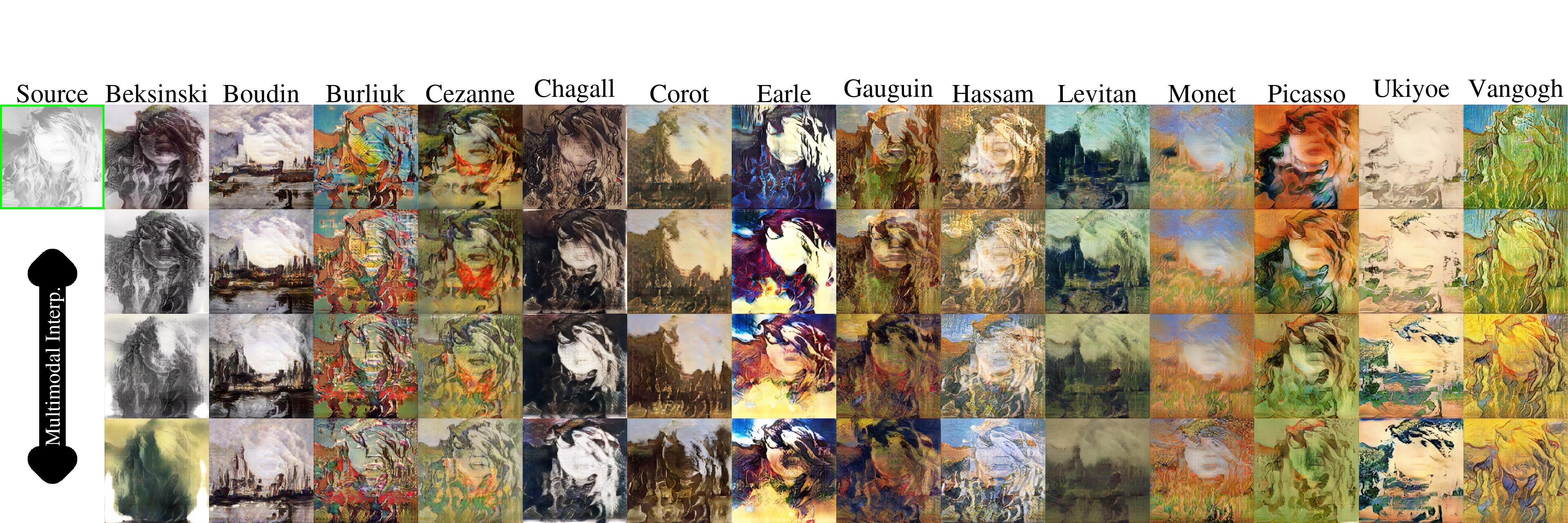}
\end{center}
   \caption{\textbf{SMIT multimodal interpolation results for painters.} Using an image in the wild as input (green box), we show the style interpolation between the first and last row for different painter domains (columns).}
\label{fig:painters_style_interp}
\vspace{3cm}
\end{figure*}

\begin{figure*}[t!]
\begin{center}
  \includegraphics[width=\linewidth]{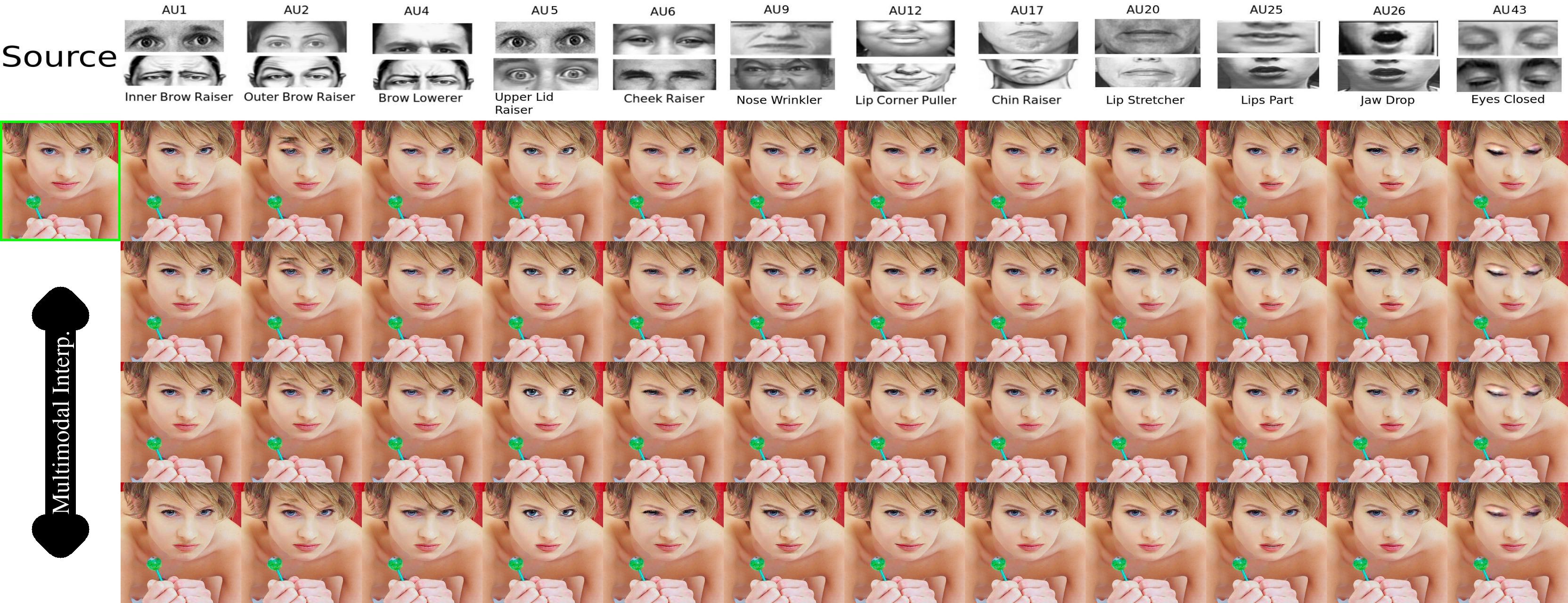}
\end{center}
   \caption{\textbf{SMIT multimodal interpolation results for EmotionNet.} Using an image from the test set as input (green box), we show the facial expression style interpolation between the first and last row for different domains (columns).}
\label{fig:emotionnet_style_interp}
\vspace{3cm}
\end{figure*}

\begin{figure*}[t!]
\begin{center}
  \includegraphics[width=\linewidth]{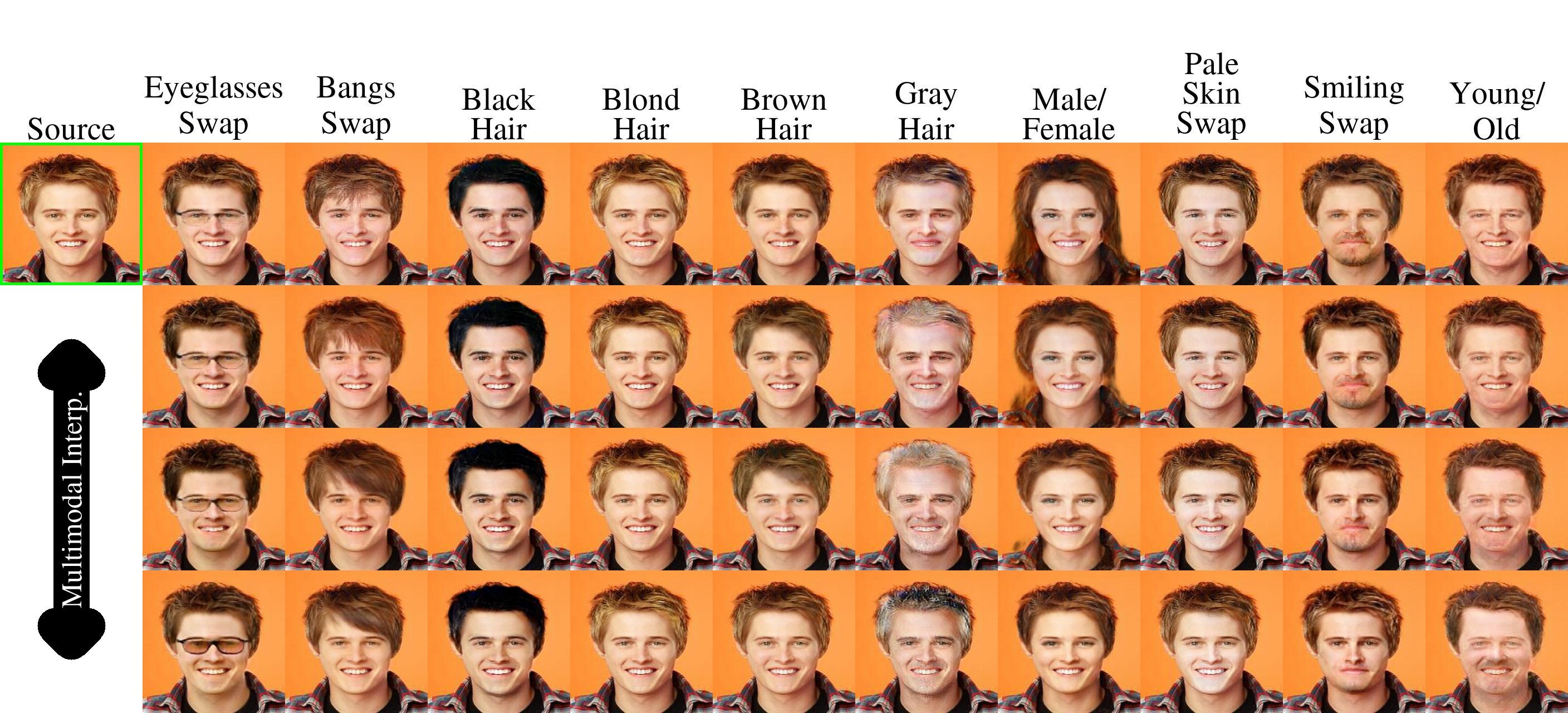}
\end{center}
   \caption{\textbf{SMIT multimodal interpolation results for CelebA.} Using an image from the test set as input (green box), we show the style interpolation between the first and last row for different attribute domains (columns).}
\label{fig:celeba10_style_interp}
\end{figure*}

\Cref{fig:shoes_style_interp,fig:handbags_style_interp,fig:yosemite_style_interp,fig:alps_style_interp,fig:rafd_style_interp,fig:painters_style_interp,fig:emotionnet_style_interp,fig:celeba10_style_interp} depict spherical interpolation between two random styles (first and last row).

\subsection{Label Interpolation}
\begin{figure*}[t!]
\begin{center}
  \includegraphics[width=\linewidth]{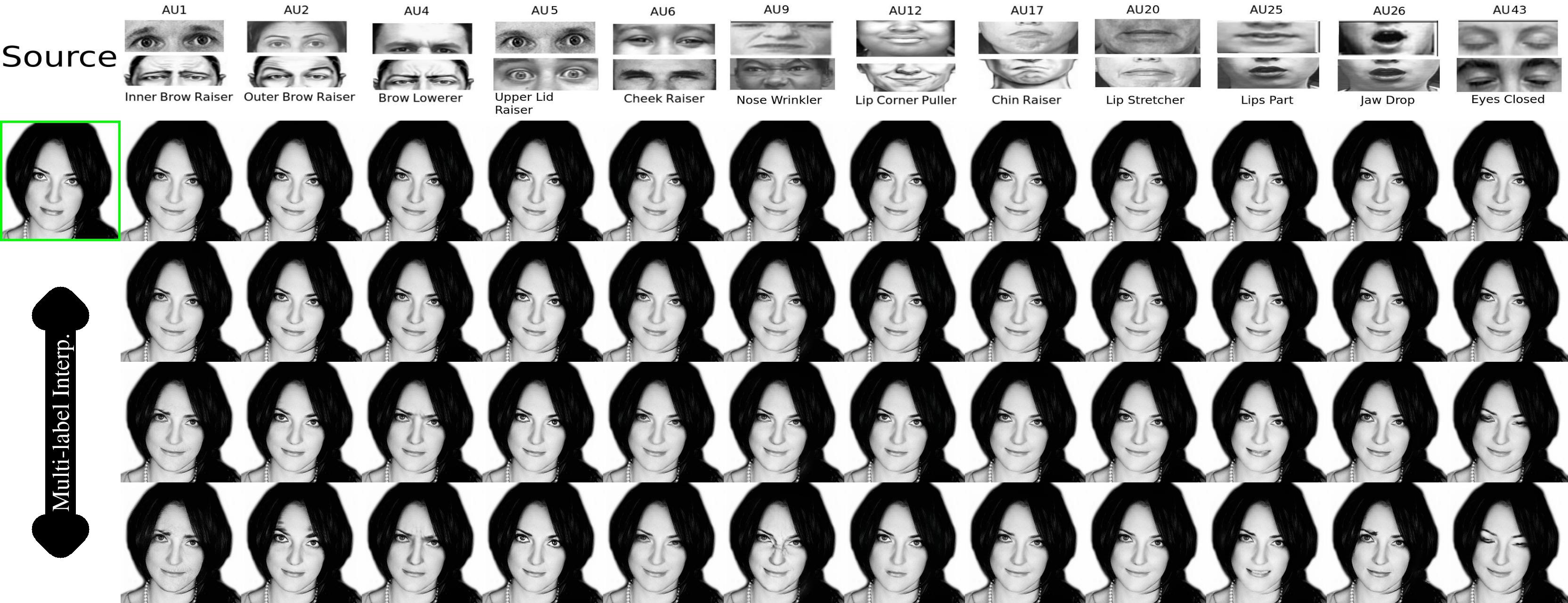}
\end{center}
   \caption{\textbf{SMIT label interpolation results for EmotionNet.} Using an image from the test set as input (green box) and a fixed modality, we show the facial expression label continuous interpolation between the first and last row for different domains (columns).}
\label{fig:emotionnet_label_interp}
\end{figure*}

\begin{figure*}[t!]
\begin{center}
  \includegraphics[width=\linewidth]{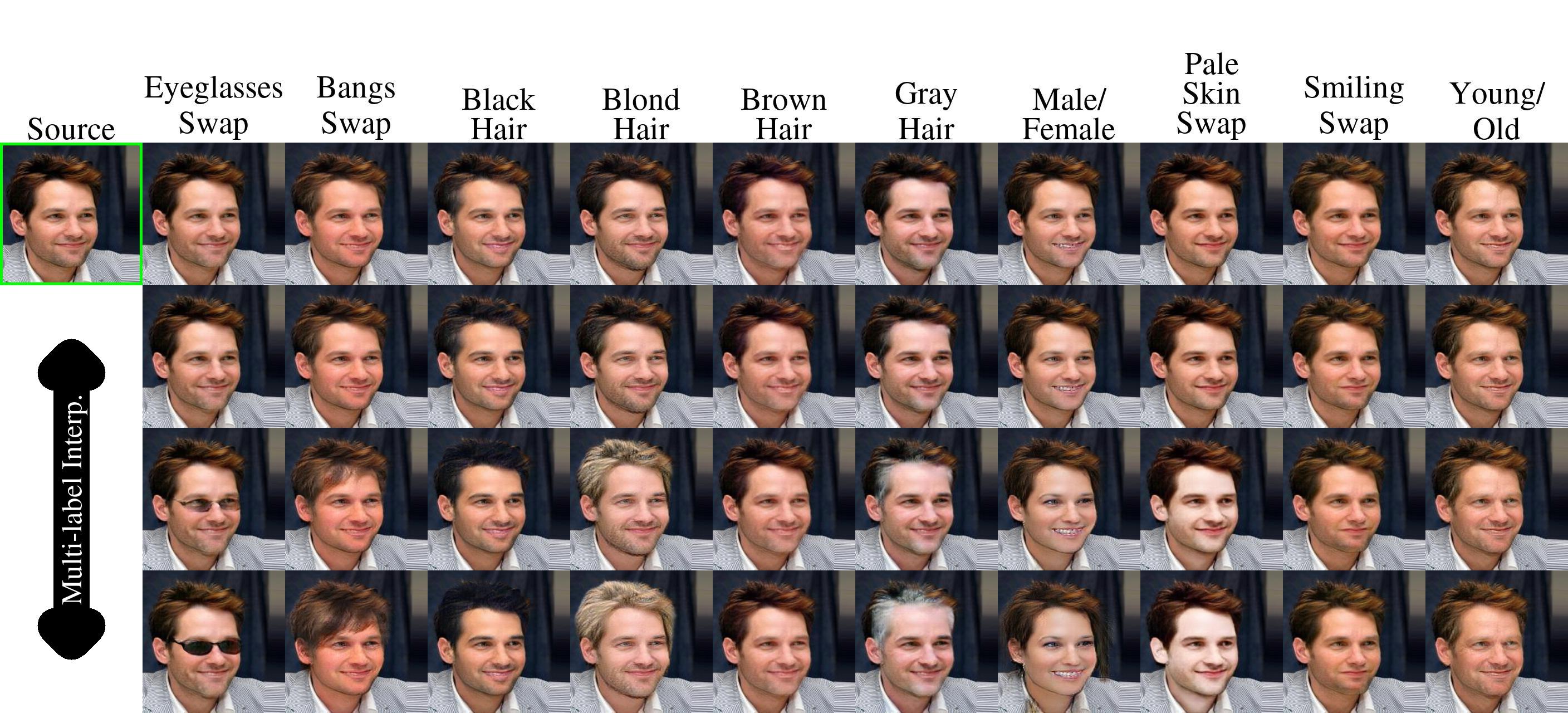}
\end{center}
   \caption{\textbf{SMIT label interpolation results for CelebA with 10 attributes.} Using an image from the test set as input (green box) and a fixed modality, we show the attribute label continuous interpolation between the first and last row for different domains (columns).}
\label{fig:celeba10_label_interp}
\vspace{3cm}
\end{figure*}

\begin{figure*}[t!]
\begin{center}
   \includegraphics[width=\linewidth]{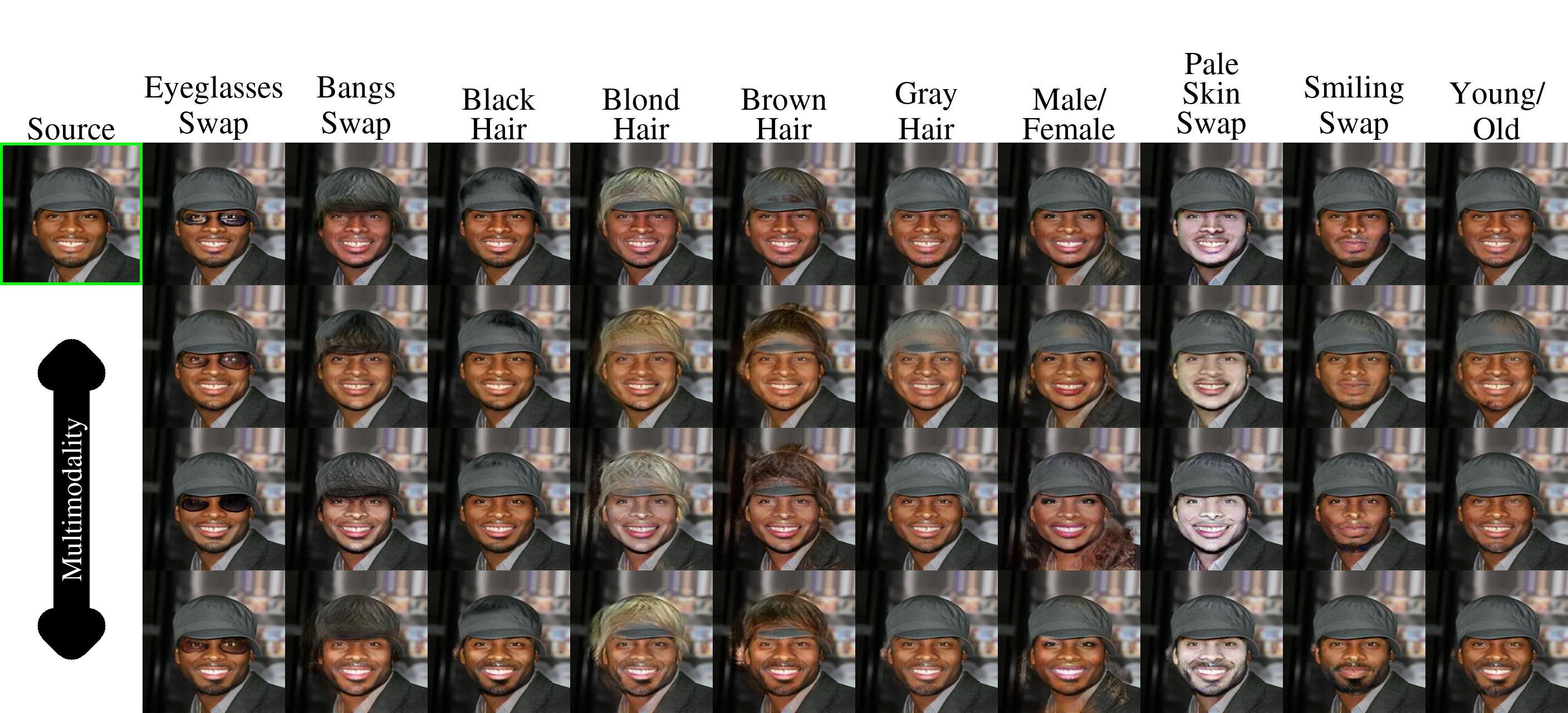}
\end{center}
   \caption{\textbf{SMIT difficult cases for CelebA with 10 attributes.} Using an image in the wild as input (green box), we show the corresponding attributes (columns) swapping (with respect to the input) for different modalities (rows).}
\label{fig:celeba_difficult0}
\end{figure*}

\begin{figure*}[t!]
\begin{center}
   \includegraphics[width=\linewidth]{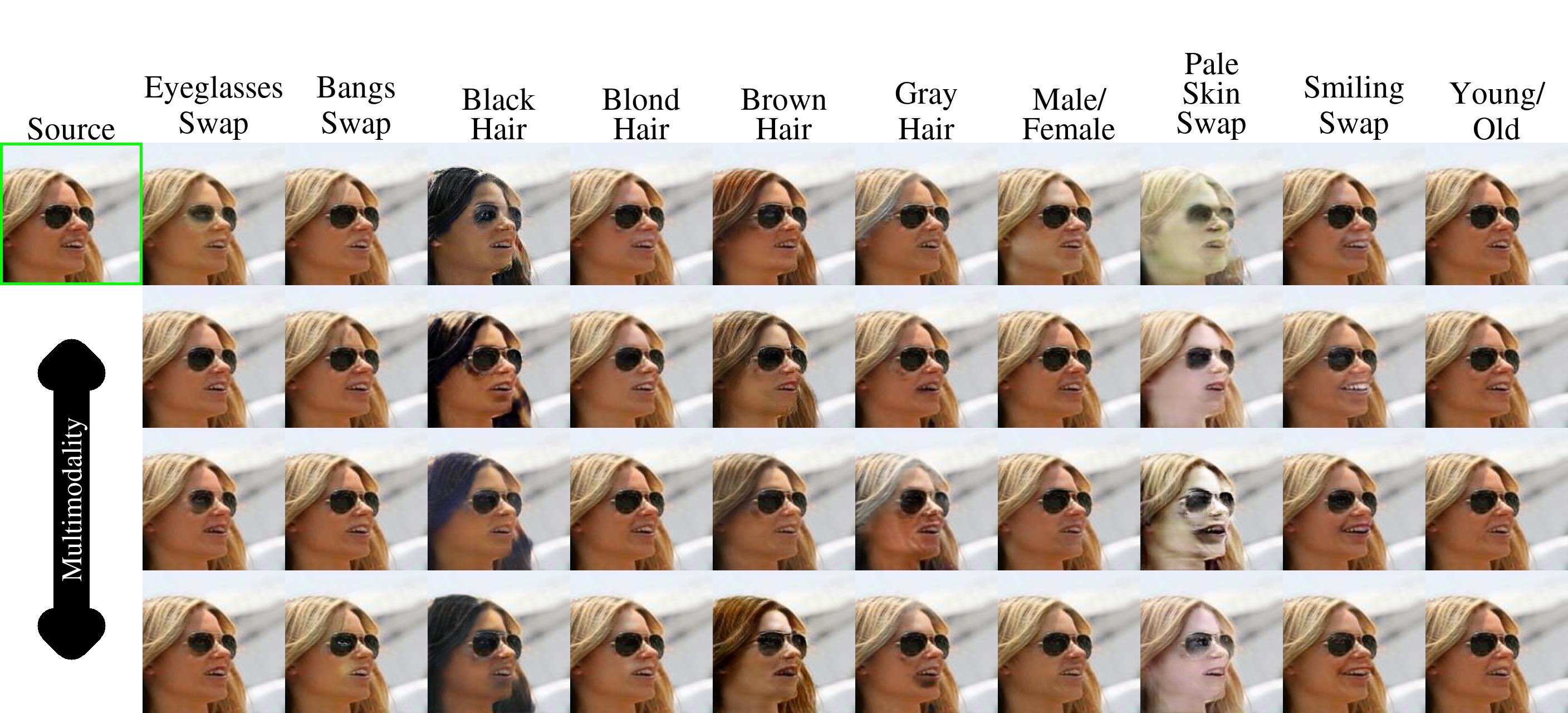}
\end{center}
   \caption{\textbf{SMIT difficult cases for CelebA with 10 attributes.} Using an image in the wild as input (green box), we show the corresponding attributes (columns) swapping (with respect to the input) for different modalities (rows).}
\label{fig:celeba_difficult1}
\end{figure*}

\begin{figure*}[t!]
\begin{center}
   \includegraphics[width=\linewidth]{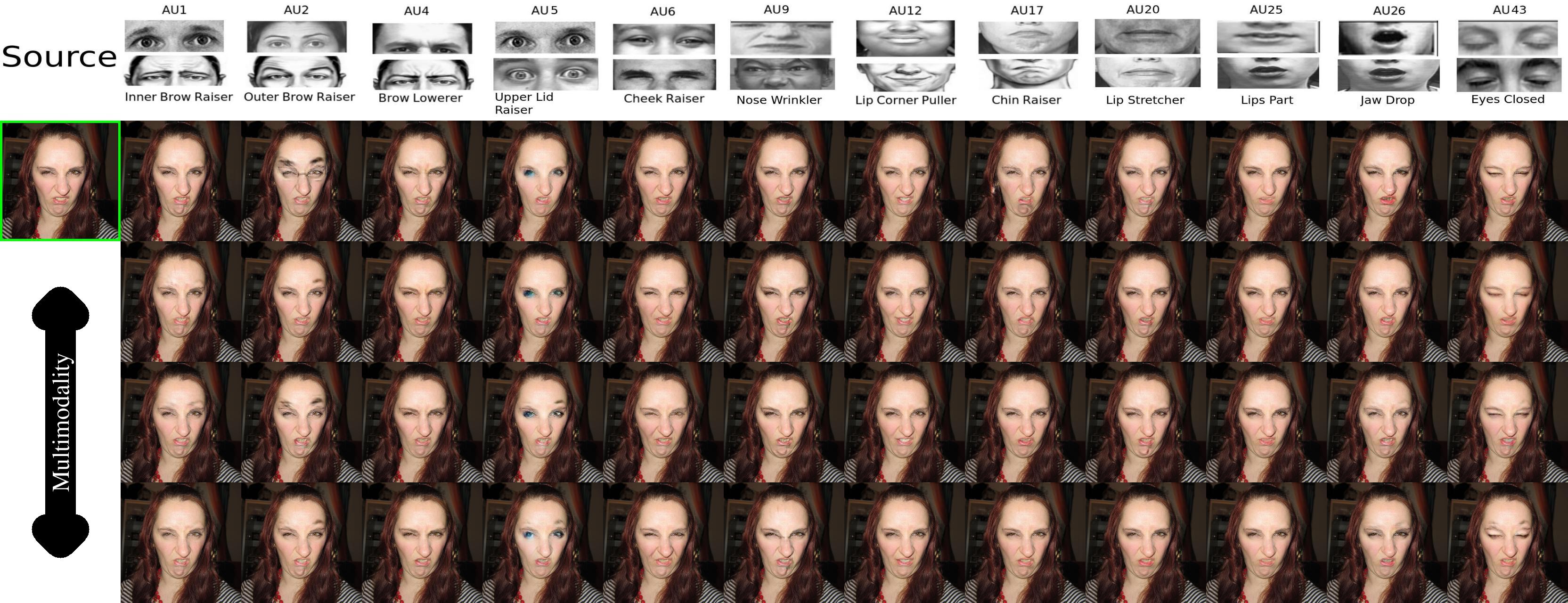}
\end{center}
   \caption{\textbf{SMIT difficult cases for EmotionNet.} Using an image in the wild as input (green box), we show the corresponding attributes (columns) swapping (with respect to the input) for different modalities (rows).}
\label{fig:emotionnet_difficult0}
\vspace{3cm}
\end{figure*}

\begin{figure*}[t!]
\begin{center}
   \includegraphics[width=\linewidth]{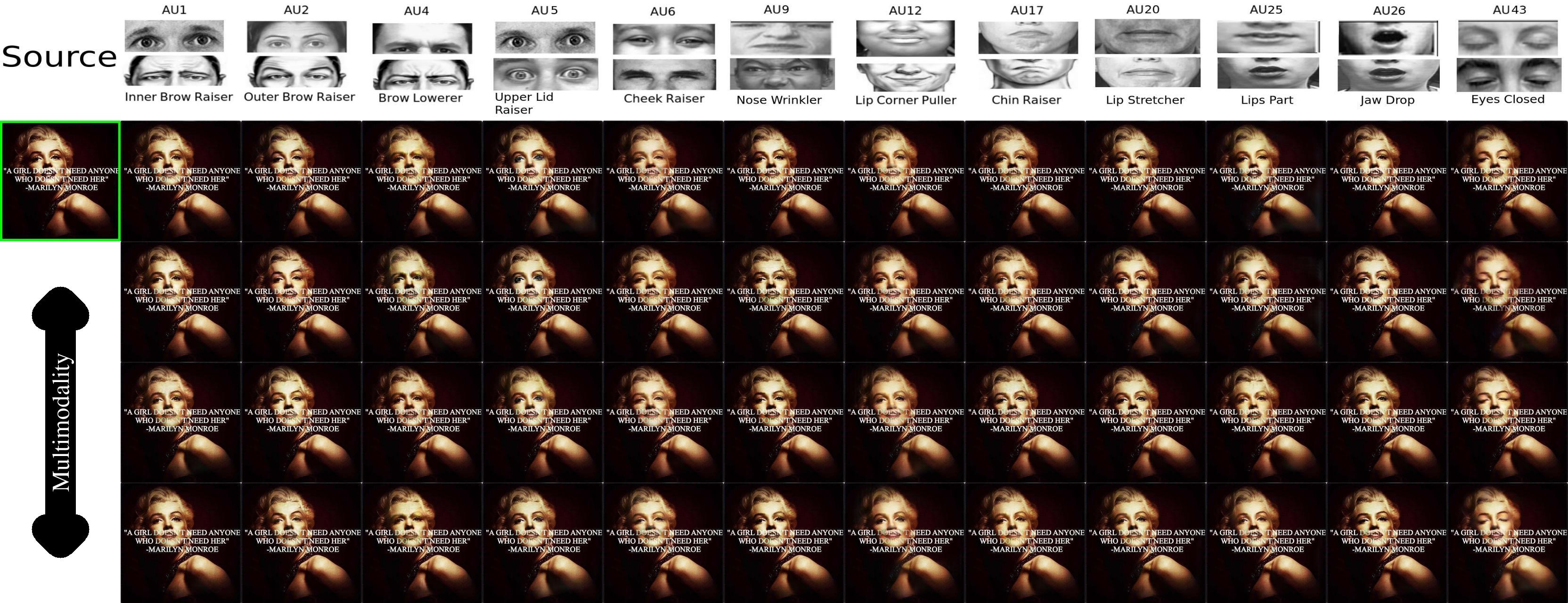}
\end{center}
   \caption{\textbf{SMIT difficult cases for EmotionNet.} Using an image in the wild as input (green box), we show the corresponding attributes (columns) swapping (with respect to the input) for different modalities (rows).}
\label{fig:emotionnet_difficult1}
\end{figure*}

\Cref{fig:emotionnet_label_interp,fig:celeba10_label_interp} present continuous label interpolation for multi-label frameworks. 
\\\\
In \Cref{fig:celeba_difficult0,fig:celeba_difficult1,fig:emotionnet_difficult0,fig:emotionnet_difficult1}, we also show SMIT qualitative visualization for difficult cases over CelebA and EmotionNet datasets. 

\section{Additional Quantitative Results}
\label{appendix:quantitative_results}
Next, we quantitatively report the evaluation metrics for each domain independently over edges2shoes, edges2handbags, edges2object, and Yosemite datasets, in \tref{table:shoes}, \ref{table:handbags}, \ref{table:objects}, and \ref{table:yosemite}, respectively.

Furthermore, in \tref{table:rafd_cis}, \ref{table:rafd_is}, \ref{table:rafd_d}, and \ref{table:rafd_pd} we discriminate the multi-label results in Conditioned Inception Score (CIS), Inception Score (IS), LPIPS Diversity (D), and LPIPS Partial Diversity (PD) over RafD dataset, respectively. 
 
\begin{table*}[t!]
\setlength{\tabcolsep}{7pt}
\renewcommand{\arraystretch}{1.5}
\begin{center}
\begin{tabular}{c  c  c}
Part & Input $\rightarrow$ Output Shape & Layer Information \\
\hline \hline
\multirow{3}{*}{Down-sampling} 
& $(256, 256, 3) \rightarrow (256, 256, 32)$ & Conv2d(dim=32, kernel=7, stride=1, padding=3), IN, ReLU \\
& $(256, 256, 32) \rightarrow (128, 128, 64)$ & Conv2d(64, 4, 2, 1), IN, ReLU \\
& $(128, 128, 64) \rightarrow (64, 64, 128)$ & Conv2d(128, 4, 2, 1), IN, ReLU \\
& $(64, 64, 128) \rightarrow (32, 32, 256)$ & Conv2d(256, 4, 2, 1), IN, ReLU \\
\Xhline{1.0pt}
\multirow{6}{*}{Bottleneck} & $(32, 32, 256) \rightarrow (32, 32, 256)$ & Residual Block: Conv2d(256, 3, 1, 1), AdaIN, ReLU\\
 & $(32, 32, 256) \rightarrow (32, 32, 256)$ &  Residual Block: Conv2d(256, 3, 1, 1), AdaIN, ReLU \\
 & $(32, 32, 256) \rightarrow (32, 32, 256)$ &   Residual Block: Conv2d(256, 3, 1, 1), AdaIN, ReLU  \\
 & $(32, 32, 256) \rightarrow (32, 32, 256)$ &   Residual Block: Conv2d(256, 3, 1, 1), AdaIN, ReLU  \\
 & $(32, 32, 256) \rightarrow (32, 32, 256)$ &   Residual Block: Conv2d(256, 3, 1, 1), AdaIN, ReLU  \\
 & $(32, 32, 256) \rightarrow (32, 32, 256)$ &   Residual Block: Conv2d(256, 3, 1, 1), AdaIN, ReLU  \\
\Xhline{1.0pt}
\multirow{2}{*}{Up-sampling} 
 & $(32, 32, 256) \rightarrow (64, 64, 128)$ & Nearest Upsampling (2x), Convd2d(128, 3, 1, 1), LN, ReLU \\
 & $(64, 64, 128) \rightarrow (128, 128, 64)$ & Nearest Upsampling (2x), Convd2d(64, 3, 1, 1), LN, ReLU \\
 & $(128, 128, 64) \rightarrow (256, 256, 32)$ & Nearest Upsampling (2x), Convd2d(32, 3, 1, 1), LN, ReLU \\ 
 \Xhline{0.5pt}

Fake Output ($\mathcal{X}_{f}$) & $(256, 256, 32) \rightarrow (256, 256, 3)$ & Conv2d(3, 7, 1, 3), None, Tanh \\
Attention mask ($\mathcal{M})$ & $(256, 256, 32) \rightarrow (256, 256, 1)$ & Conv2d(1, 7, 1, 3), None, Sigmoid \\
\hline
\hline
\end{tabular}
\end{center}
\caption{\textbf{SMIT Generator network architecture.}}
\label{table:generator}
\end{table*}

\begin{table*}[t!]
\setlength{\tabcolsep}{7pt}
\renewcommand{\arraystretch}{1.7}
\begin{center}
\begin{tabular}{c c c}
Layer & Input $\rightarrow$ Output Shape & Layer Information \\
\hline \hline
Embedding Projection & $(20+\mathbb{N}_d) \rightarrow (6144) $ & FullyConnected(dim=$6144$)\\
\hline \hline
\end{tabular}
\end{center}
\caption{\textbf{SMIT Domain Embedding network architecture.}}
\label{table:domain_embedding}
\end{table*}


\begin{table*}[t!]
\setlength{\tabcolsep}{7pt}
\renewcommand{\arraystretch}{1.7}
\begin{center}
\begin{tabular}{c c c}
Layer & Input $\rightarrow$ Output Shape & Layer Information \\
\hline \hline
Input Layer & $(256, 256, 3) \rightarrow (128, 128, 32) $ & Conv2d(dim=32, kernel=4, stride=2, padding=1), SN, LReLU \\
\Xhline{1.0pt}
Hidden Layer & $(128, 128, 32) \rightarrow (64, 64, 64) $ & Conv2d(64, 4, 2, 1), SN, LReLU \\
Hidden Layer & $(64, 64, 64) \rightarrow (32, 32, 128) $ & Conv2d(128, 4, 2, 1), SN, LReLU \\
Hidden Layer & $(32, 32, 128) \rightarrow (16, 16, 256) $ & Conv2d(256, 4, 2, 1), SN, LReLU \\
Hidden Layer & $(16, 16, 256) \rightarrow (8, 8, 512) $ & Conv2d(512, 4, 2, 1), SN, LReLU \\
Hidden Layer & $(8, 8, 512) \rightarrow (4, 4, 1024) $ & Conv2d(1024, 4, 2, 1), SN, LReLU \\
Hidden Layer & $(4, 4, 1024) \rightarrow (2, 2, 2048) $ & Conv2d(2048, 4, 2, 1), SN, LReLU \\
\Xhline{1.0pt}
Output Layer ($\mathbb{D}_{src}$) & $(2, 2, 2048) \rightarrow (2, 2, 1) $ & Conv2d(1, 3, 1, 1) \\
Output Layer ($\mathbb{D}_{cls}$) & $(2, 2, 2048) \rightarrow (1, 1, \mathbb{N}_d) $ & Conv2d($\mathbb{N}_d, 2, 1, 0$) \\
\hline \hline
\end{tabular}
\end{center}
\caption{\textbf{SMIT Discriminator network architecture.}}
\label{table:discriminator}
\end{table*}

\begin{figure*}[t!]
\begin{center}
   \includegraphics[width=\linewidth]{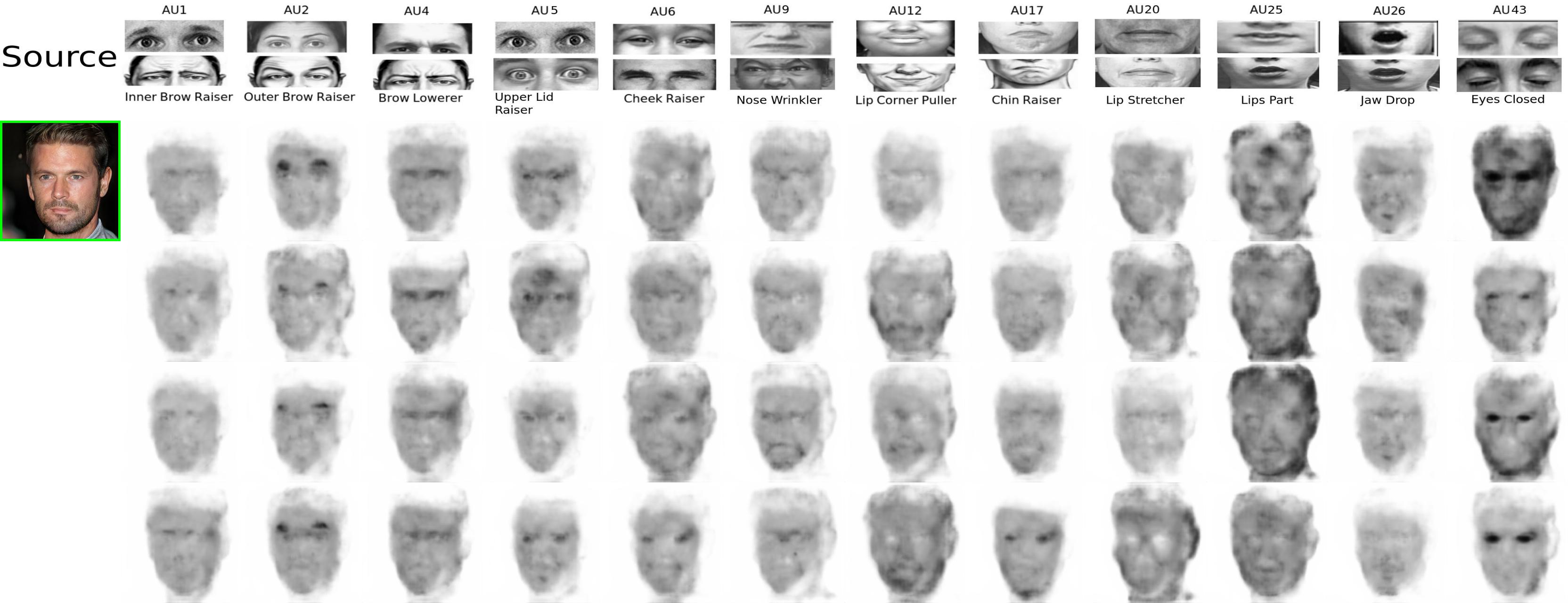}
\end{center}
   \caption{\textbf{SMIT qualitative attention results for EmotionNet.} These images are the attention maps that produces \fref{fig:emotionnet1}. Black regions represent the changes with respect to the input.}
\label{fig:emotionnet1_attn}
\vspace{3cm}
\end{figure*}

\begin{figure*}[t!]
\begin{center}
   \includegraphics[width=\linewidth]{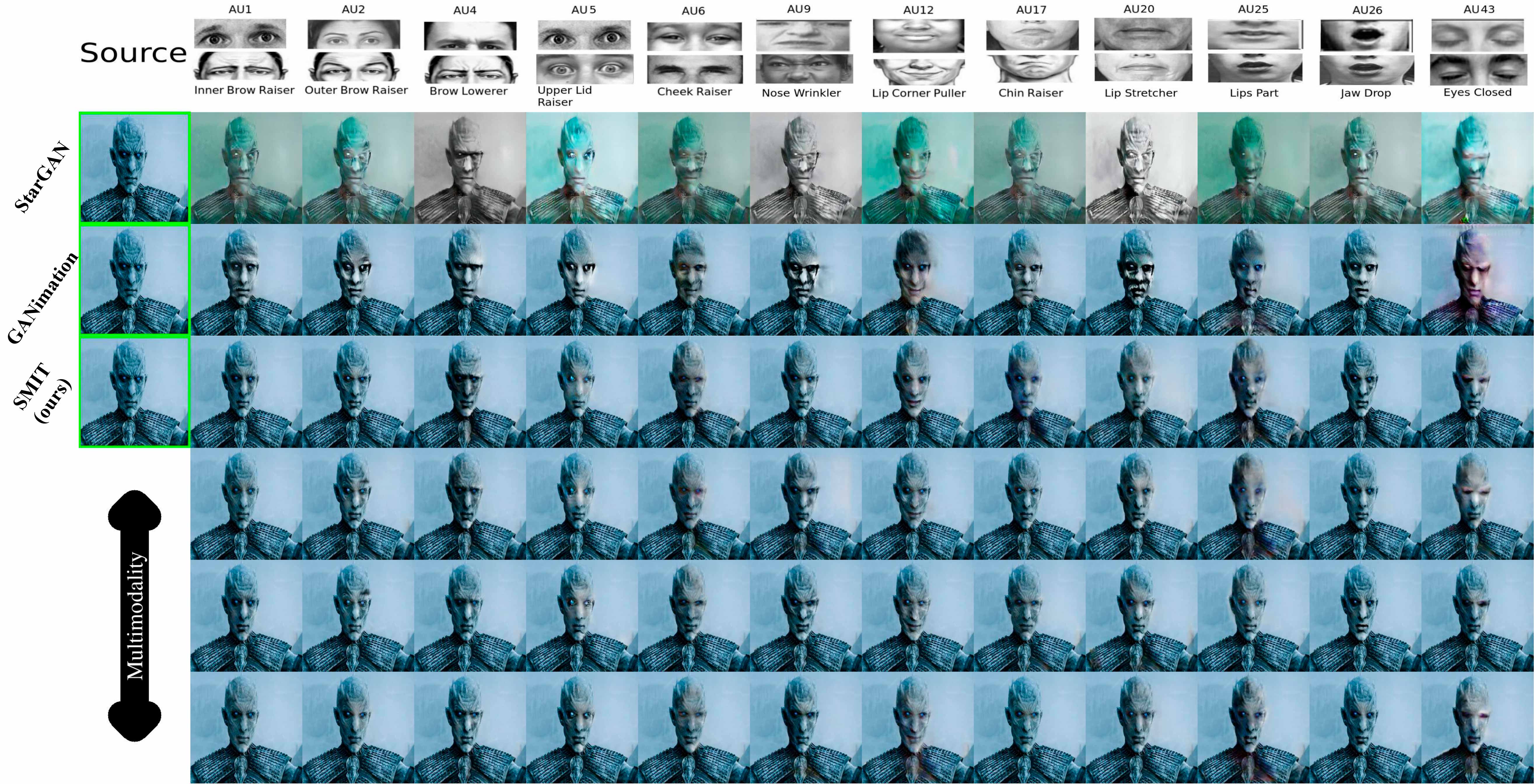}
\end{center}
   \caption{\textbf{SMIT qualitative results for EmotionNet.} Using an image in the wild as input, we show qualitative comparison against state-of-the-art methods in multi-label image-to-image translation. We retrain StarGAN and GANimation for these results.}
\label{fig:emotionnet_comparison}
\end{figure*}

\clearpage

\begin{figure*}[t!]
\begin{center}
  \includegraphics[width=\linewidth]{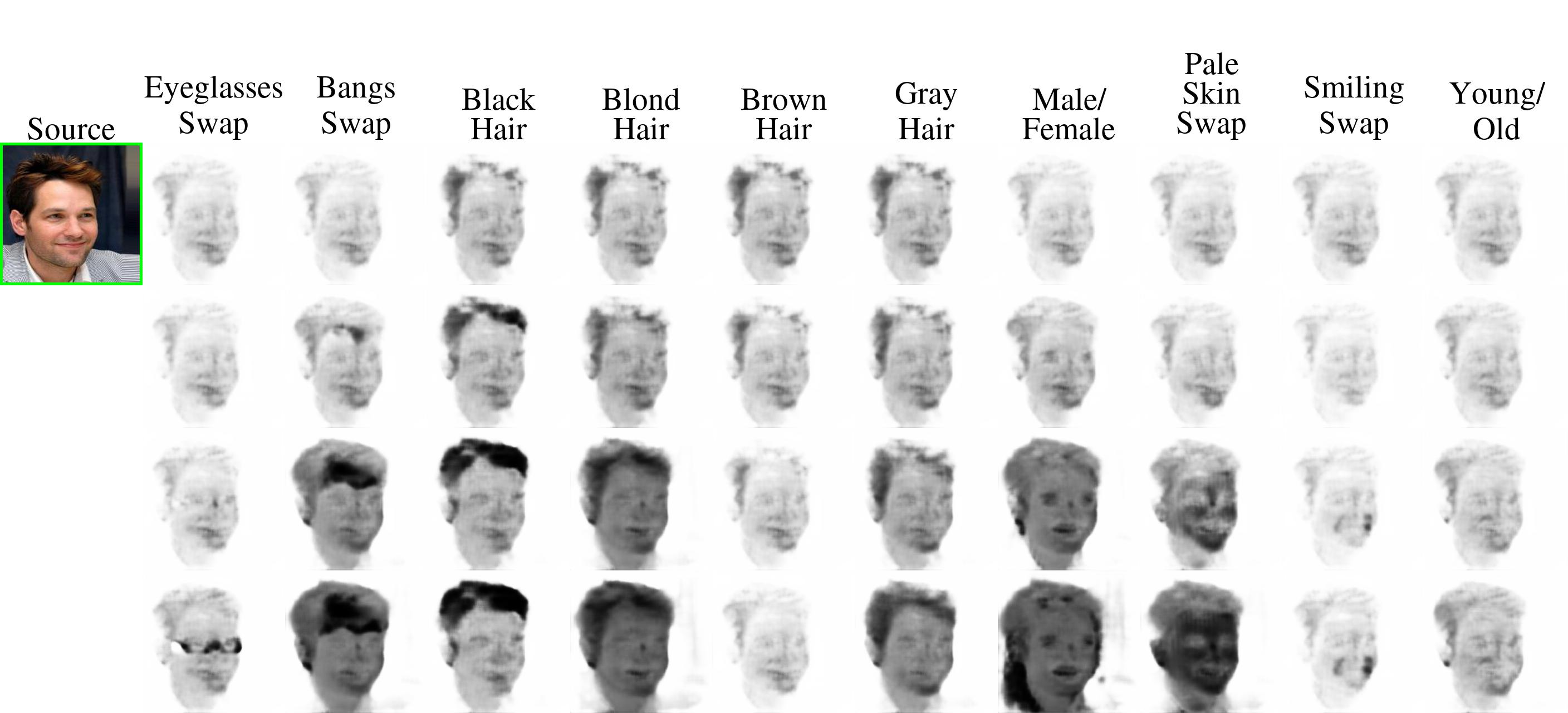}
\end{center}
   \caption{\textbf{SMIT label mask interpolation results for CelebA.} These attentions map represent the outputs of the \fref{fig:celeba10_label_interp}. Black regions represent the changes with respect to the input.}
\label{fig:celeba10_label_interp_attn}
\end{figure*}

\maxdeadcycles=200
\extrafloats{100}


\begin{table*}[t!]
\begin{center}
\begin{tabular}{|c||c|c|c|c|}
\hline
 & \multicolumn{4}{c|}{Edges2Shoes} \\
\cline{2-5}
& \multicolumn{2}{c|}{Edges} & \multicolumn{2}{c|}{Shoes} \\              
\cline{2-5}
              & D & PD & D & PD \\
\hline
\rowcolor[HTML]{\colortable} 
CycleGAN       & 0.269$\pm$0.046 & - & \textbf{0.275$\pm$0.050} & - \\
DRIT           & 0.000$\pm$0.000 & 0.000$\pm$0.000 & 0.243$\pm$0.052 & 0.056$\pm$0.017 \\
\rowcolor[HTML]{\colortable} 
MUNIT          & 0.269$\pm$0.049 & \textbf{0.027$\pm$0.005} & 0.263$\pm$0.049 & \textbf{0.126$\pm$0.039} \\
\textbf{SMIT (ours)}  & \textbf{0.274$\pm$0.046} & 0.020$\pm$0.006 & 0.261$\pm$0.060 & 0.123$\pm$0.029 \\
\hline
Real Data      & 0.274$\pm$0.046 & - & 0.293$\pm$0.051 & - \\
\hline
\end{tabular}
\caption{\textbf{Multimodal quantitative evaluation for edges2shoes.} We report the LPIPS score to compare the diversity (D) and partial diversity (PD) for each domain independently, in comparison with multimodal frameworks. We retrain CycleGAN, DRIT and MUNIT for these results.}
\label{table:shoes}
\end{center}
\vspace{-0.5cm}
\end{table*}

\begin{table*}[t!]
\begin{center}
\begin{tabular}{|c||c|c|c|c|}
\hline
 & \multicolumn{4}{c|}{Edges2Handbags} \\
\cline{2-5}
& \multicolumn{2}{c|}{Edges} & \multicolumn{2}{c|}{Handbags} \\              
\cline{2-5}
              & D & PD & D & PD \\
\hline
\rowcolor[HTML]{\colortable} 
CycleGAN       & 0.225$\pm $0.043 & - & \textbf{0.361$\pm$0.045} & - \\
DRIT           & 0.000$\pm$0.000 & 0.000$\pm$0.000 & 0.344$\pm$0.061 & 0.112$\pm$0.032 \\
\rowcolor[HTML]{\colortable} 
MUNIT          & 0.352$\pm$0.045 & \textbf{0.063$\pm$0.016} & 0.334$\pm$0.052 & \textbf{0.183$\pm$0.039} \\
\textbf{SMIT (ours)}  & \textbf{0.373$\pm$0.041} & 0.029$\pm$0.010 & 0.346$\pm$0.048 & 0.164$\pm$0.035 \\
\hline
Real Data      & 0.346$\pm$0.045 & - & 0.370$\pm$0.053 & - \\
\hline
\end{tabular}
\caption{\textbf{Multimodal quantitative evaluation for edges2handbags.} We report the LPIPS score to compare the diversity (D) and partial diversity (PD) for each domain independently, in comparison with multimodal frameworks. We retrain CycleGAN, DRIT and MUNIT for these results.}
\label{table:handbags}
\end{center}
\end{table*}

\begin{table*}[t!]
\begin{center}
\resizebox{\linewidth}{!}{
\begin{tabular}{|c||c|c|c|c|c|c|c|c|}
\hline
 & \multicolumn{8}{c|}{Edges2Objects} \\
\cline{2-9}
& \multicolumn{2}{c|}{Edges Shoes} & \multicolumn{2}{c|}{Shoes} & \multicolumn{2}{c|}{Edges Handbags} & \multicolumn{2}{c|}{Handbags} \\              
\cline{2-9}
              & D & PD & D & PD & D & PD & D & PD \\
\hline
\rowcolor[HTML]{\colortable} 
CycleGAN       & - & - & - & - & - & - & - & - \\
DRIT           & - & - & - & - & - & - & - & - \\
\rowcolor[HTML]{\colortable} 
MUNIT          & - & - & - & - & - & - & - & - \\
\textbf{SMIT (ours)}  & 0.130$\pm$0.104 & 0.055$\pm$0.024 & 0.286$\pm$0.07 & 0.168$\pm$0.028 & 0.279$\pm$0.045 & 0.012$\pm$0.008 & 0.304$\pm$0.052 & 0.233$\pm$0.060 \\
\hline
Real Data      & 0.274$\pm$0.046 & - & 0.293$\pm$0.051 & - & 0.346$\pm$0.045 & - & 0.370$\pm$0.053 & - \\
\hline
\end{tabular}
}
\caption{\textbf{Multimodal quantitative evaluation for edges2objects.} We report the LPIPS score to compare the diversity (D) and partial diversity (PD) for each domain independently, in comparison with multimodal frameworks. Due to the multi-label nature, SMIT is the only one that is suitable for this task.}
\label{table:objects}
\end{center}
\end{table*}

\begin{table*}[t!]
\begin{center}
\begin{tabular}{|c||c|c|c|c|}
\hline
 & \multicolumn{4}{c|}{Yosemite} \\
\cline{2-5}
& \multicolumn{2}{c|}{Summer} & \multicolumn{2}{c|}{Winter} \\              
\cline{2-5}
              & D & PD & D & PD \\
\hline
\rowcolor[HTML]{\colortable} 
CycleGAN       & \textbf{0.408$\pm$0.037} & - & 0.406$\pm$0.041 & - \\
DRIT           & 0.405$\pm$0.033 & 0.120$\pm$0.018 & 0.395$\pm$0.040 & 0.131$\pm$0.020 \\
\rowcolor[HTML]{\colortable} 
MUNIT          & 0.372$\pm$0.034 & \textbf{0.212$\pm$0.029} & 0.313$\pm$0.035 & \textbf{0.204$\pm$0.037} \\
\textbf{SMIT (ours)}  & 0.378$\pm$0.048 & 0.167$\pm$0.070 & \textbf{0.410$\pm$0.049} & 0.129$\pm$0.069 \\
\hline
Real Data      & 0.444$\pm$0.055 & - & 0.444$\pm$0.040 & - \\
\hline
\end{tabular}
\caption{\textbf{Multimodal quantitative evaluation for Yosemite.} We report the LPIPS score to compare the diversity (D) and partial diversity (PD) for each domain independently, in comparison with multimodal frameworks. We retrain CycleGAN, DRIT and MUNIT for these results.}
\label{table:yosemite}
\end{center}
\end{table*}

\begin{table*}[t!]
\begin{center}
\begin{tabular}{|c||c|c|c|c|c|c|c|c|}
\hline
& \multicolumn{8}{c|}{RafD} \\
\cline{2-9}
& \multicolumn{8}{c|}{Conditional Inception Score (CIS)} \\
\cline{2-9}
 & Neutral & Anger & Contempt & Disgust & Fear & Happy & Sad & Surprise \\
\hline
\rowcolor[HTML]{\colortable} 
StarGAN         & 1.000 & 1.000 & 1.000 & 1.000 & 1.000 & 1.000 & 1.000 & 1.000 \\
GANimation      & 1.000 & 1.000 & 1.000 & 1.000 & 1.000 & 1.000 & 1.000 & 1.000 \\
\rowcolor[HTML]{\colortable} 
\textbf{SMIT (ours)}  & \textbf{1.201} & \textbf{1.187} & \textbf{1.197} & \textbf{1.237} & \textbf{1.329} & \textbf{1.373} & \textbf{1.249} & \textbf{1.201} \\
\hline
\end{tabular}
\caption{\textbf{Multi-label quantitative evaluation for RafD.} We report the Conditional Inception Score (CIS) for each domain independently, in comparison with multi-label frameworks. We retrain StarGAN and GANimation for these results.}
\label{table:rafd_cis}
\end{center}
\end{table*}
\medskip
\begin{table*}[t!]
\begin{center}
\begin{tabular}{|c||c|c|c|c|c|c|c|c|}
\hline
& \multicolumn{8}{c|}{RafD} \\
\cline{2-9}
& \multicolumn{8}{c|}{Inception Score (IS)} \\
\cline{2-9}
 & Neutral & Anger & Contempt & Disgust & Fear & Happy & Sad & Surprise \\
\hline
\rowcolor[HTML]{\colortable} 
StarGAN               & 2.039 & 1.407 & 2.194 & 1.081 & 1.748 & 1.483 & 2.060 & 1.275 \\
GANimation            & 1.559 & 1.320 & 2.024 & 1.115 & 1.427 & 1.698 & 1.888 & 1.033 \\
\rowcolor[HTML]{\colortable} 
\textbf{SMIT (ours)}  & \textbf{3.502} & \textbf{2.246} & \textbf{3.441} & \textbf{1.598} & \textbf{2.451} & \textbf{2.327} & \textbf{3.009} & \textbf{1.527} \\
\hline
Real Data       & 1.120 & 1.439 & 1.401 & 1.001 & 1.360 & 1.001 & 1.126 & 1.007 \\
\hline
\end{tabular}
\caption{\textbf{Multi-label quantitative evaluation for RafD.} We report the Inception Score (IS) for each domain independently, in comparison with multi-label frameworks. We retrain StarGAN and GANimation for these results.}
\label{table:rafd_is}
\end{center}
\end{table*}

\begin{table*}[t!]
\begin{center}
\begin{tabular}{|c||c|c|c|c|c|c|c|c|}
\hline
& \multicolumn{8}{c|}{RafD} \\
\cline{2-9}
& \multicolumn{8}{c|}{Diversity (D)} \\
\cline{2-9}
 & Neutral & Anger & Contempt & Disgust & Fear & Happy & Sad & Surprise \\
\hline
\rowcolor[HTML]{\colortable} 
StarGAN         & 0.157 & 0.154 & 0.152 & 0.152 & 0.152 & 0.150 & 0.149 & 0.150 \\
GANimation      & 0.156 & 0.156 & 0.154 & 0.156 & 0.156 & 0.157 & 0.159 & 0.160 \\
\rowcolor[HTML]{\colortable} 
\textbf{SMIT (ours)}  & \textbf{0.164} & \textbf{0.161} & \textbf{0.162} & \textbf{0.163} & \textbf{0.163} & \textbf{0.164} & \textbf{0.165} & \textbf{0.170} \\
\hline
Real Data       & 0.167 & 0.165 & 0.166 & 0.166 & 0.166 & 0.167 & 0.167 & 0.167 \\
\hline
\end{tabular}
\caption{\textbf{Multi-label quantitative evaluation for RafD.} We report the LPIPS diversity metric (D) for each domain independently, in comparison with multi-label frameworks. We retrain StarGAN and GANimation for these results.}
\label{table:rafd_d}
\end{center}
\end{table*}

\begin{table*}[t!]
\begin{center}
\begin{tabular}{|c||c|c|c|c|c|c|c|c|}
\hline
& \multicolumn{8}{c|}{RafD} \\
\cline{2-9}
& \multicolumn{8}{c|}{Partial Diversity (PD)} \\
\cline{2-9}
 & Neutral & Anger & Contempt & Disgust & Fear & Happy & Sad & Surprise \\
\hline
\rowcolor[HTML]{\colortable} 
StarGAN         & - & - & - & - & - & - & - &  \\
GANimation      & - & - & - & - & - & - & - &  \\
\rowcolor[HTML]{\colortable} 
\textbf{SMIT (ours)}  & 0.003 & 0.004 & 0.003 & 0.004 & 0.004 & 0.004 & 0.003 & 0.005 \\
\hline
\end{tabular}
\caption{\textbf{Multi-label quantitative evaluation for RafD.} We report the LPIPS partial diversity metric (PD) for each domain independently, in comparison with multi-label frameworks. We retrain StarGAN and GANimation for these results.}
\label{table:rafd_pd}
\end{center}
\end{table*}
\end{appendices}

\end{document}